\documentclass{article}
\usepackage{PRIMEarxiv}

\usepackage[utf8]{inputenc} 
\usepackage[T1]{fontenc}    
\usepackage[pagebackref,breaklinks,colorlinks]{hyperref}       
\usepackage{url}            
\usepackage{booktabs}       
\usepackage{amsmath,amssymb} 
\usepackage{amsfonts}       
\usepackage{nicefrac}       
\usepackage{microtype}      
\usepackage{lipsum}
\usepackage{fancyhdr}       
\usepackage{graphicx}       
\usepackage{subcaption}
\usepackage{tabularx}
\usepackage[export]{adjustbox}
\usepackage{float}
\usepackage{comment}
\usepackage[shortlabels]{enumitem}
\usepackage[normalem]{ulem}
\usepackage{tikz}
\usepackage{svg}            
\usepackage{authblk}

\newcommand*\samethanks[1][\value{footnote}]{\footnotemark[#1]}
\author[1 \thanks{The authors contribute equally to this paper.}]{Shimian Zhang }
\author[1 \samethanks]{Skanda Bharadwaj}
\author[1 \samethanks]{Keaton Kraiger}
\author[2 \thanks{The work is accomplished at The Pennsylvania State University.}]{Yashasvi Asthana }
\author[3 \samethanks]{Hong Zhang}
\author[1]{Robert Collins}
\author[1]{Yanxi Liu}

\affil[1]{Pennsylvania State University, PA, US}
\affil[2]{Nimble Robotics Inc., CA, US}
\affil[3]{Meituan AI., Beijing, China}

\affil[1]{\texttt{\{svz5303, ssb248, kbk5531, rtc12, yul11\}@psu.edu}}
\affil[2]{\texttt{yashasvi@nimble.ai}}
\affil[3]{\texttt{zhanghong50@meituan.com}}

\begin{document}
\suppressfloats
\newcommand{\COMMENT}[1]{}
\definecolor{aqua}{rgb}{0.0, 1.0, 1.0}
  

\title{Novel 3D Scene Understanding Applications  From~Recurrence~in~a~Single~Image 
}

\maketitle

\vspace{-2em}
\begin{abstract}
We demonstrate the utility of recurring pattern discovery from a single image for spatial understanding of a 3D scene in terms of  
(1) vanishing point detection, 
(2) hypothesizing 3D translation symmetry and 
(3) counting the number of RP instances in the image. 
Furthermore, we illustrate the feasibility of leveraging RP discovery output to form a more precise, quantitative text description of the scene. 
Our quantitative evaluations on a new 1K+ Recurring Pattern (RP) benchmark with diverse variations show that visual perception of recurrence from one single view leads to scene understanding outcomes that are as good as or better than existing supervised methods and/or unsupervised methods that use millions of images.
\end{abstract}
\section{Introduction}
\label{sec:introduction}

\begin{figure}[] \centering
    \begin{subfigure}[t]{0.24\linewidth}
        \centering
         \includegraphics[height=\textwidth, width=\textwidth]{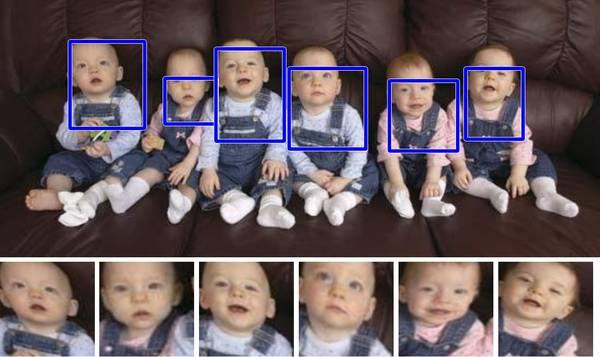}
         \vspace{-17pt}
         \caption{\small \label{fig:1-a}}
    \end{subfigure}
    \hfill
    \begin{subfigure}[t]{0.24\linewidth}
        \centering
         \includegraphics[height=\textwidth,width=\textwidth]{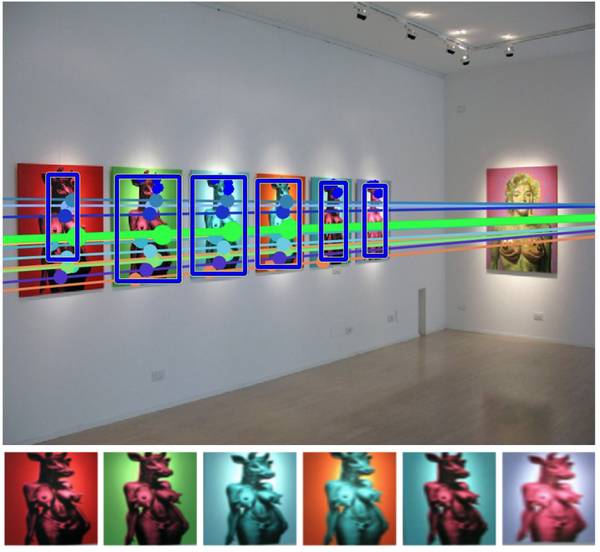}
         \vspace{-17pt}
         \caption{\small \label{fig:1-b}}
    \end{subfigure}
    \hfill
    \begin{subfigure}[t]{0.24\linewidth}
        \centering
         \includegraphics[height=\textwidth,width=\textwidth]{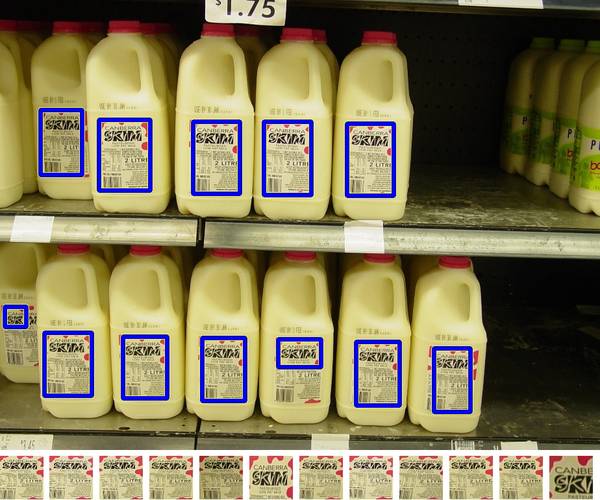}
        \vspace{-17pt}
        \caption{\small \label{fig:1-c}}
    \end{subfigure}
    \hfill
    \begin{subfigure}[t]{0.24\linewidth}
        \centering
         \includegraphics[height=\textwidth,width = \textwidth]{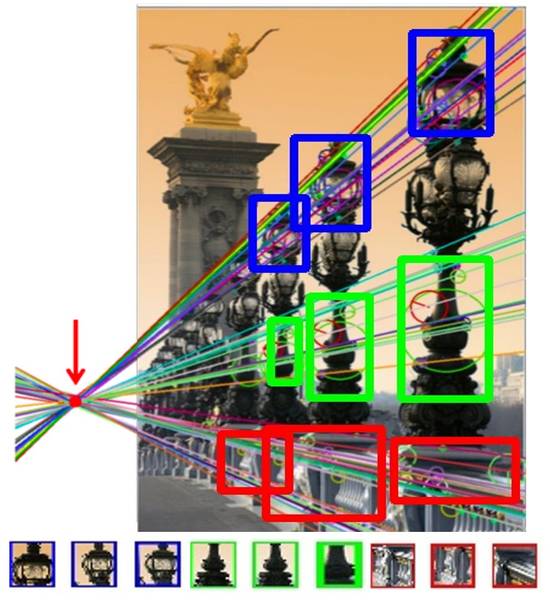}
         \vspace{-17pt}
         \caption{\small \label{fig:1-d}}
    \end{subfigure}
    \caption{\small  RP examples: \textbf{(a)} six similar baby faces, \textbf{(b)} six similar paintings, vanishing point (VP) detected (outside of the image), potential 3D translation symmetry discovered, \textbf{(c)} 12 milk labels, \textbf{(d)} three distinct RPs captured from  different regions of the lamps, with a VP found (outside of the image). These examples show that an RP instance does not necessarily correspond to an object: an RP is simply a set of similar visual entities that recur within an image.
    }
    \label{fig:rp_ex_intro}
\end{figure}

{\em Pattern} discovery from 2D images is fundamental for humans and robots to understand the 3D world \cite{li2009computational,treder2010behind,michaux2016figure,sawada2014detecting}, in particular \textbf{Recurring Patterns (RP)}, referring to the broadly  defined a set of ``{\em things that recur}" \cite{liu2013grasp}.
The non-identical (both geometrically and photometrically) yet similar elements comprising an RP are named \textbf{RP instances} (Fig.~\ref{fig:rp_ex_intro}).
Different from the term ``object'' widely used in the literature, an RP instance may or may not correspond to an object completely (Fig.~\ref{fig:1-a}, \ref{fig:1-c}, \ref{fig:1-d}), 
yet we show in this work that RP instances discovered in an unsupervised manner can lead to novel applications in 3D scene understanding. 
%
Our working hypothesis is that the recurrence of class-agnostic things (a much more relaxed notion than symmetry \cite{mitra2006partial,liu2010common}) 
may imply semantic significance in a way similar to how humans perceive/discover new environments class-agnostically \cite{mattson2014superior}.
%

{\bf U}nsupervised {\bf R}ecurring {\bf P}attern {\bf D}iscovery ({\bf URPD}) is a challenging class-agnostic, zero-shot  task. 
%
Relevant to scene understanding, recurrence can be leveraged to infer aspects of 3D scene geometry.  RP instances of similar size in 3D viewed under perspective projection appear differently-sized in 2D, allowing estimation of relative depths.  RP patterns frequently consist of colinear instances in 3D, and these project to 2D instances leading to a vanishing point, from which the 3D orientation of the line of instances can be inferred with respect to the camera view \cite{mundy1992projective,Schaffalitzky00a}. 
Furthermore, colinear RP instances are often equally spaced in 3D, and this {\it 3D translation symmetry} \cite{weyl2015symmetry} can be quantitatively evaluated and hypothesized in 2D using by projective invariants measures.
URPD from a single view can lead to a general approach for single view metrology \cite{criminisciSingleViewMetrology2000,singleviewmetrologyinthewild2020}, leveraging class-agnostic RP discovery rather than needing explicit line detection \cite{criminisciSingleViewMetrology2000} or known-class (e.g.~pedestrian) detection \cite{singleviewmetrologyinthewild2020}.

Previous work on URPD \cite{liu2013grasp} had limited success due to a lack of computational feasibility, generality and benchmarking, and missed the opportunity to infer 3D scene geometry from observations of recurrence.
%
Our main contributions in this work include:

\begin{itemize}[topsep=-2pt]
\item We introduce an RP-benchmark dataset of 1K+ images collected across diverse sources, labeled with RPs and RP-instances (Sec.~\ref{sec:collectdataset}), including a benchmark subset of 147 images with labeled ground truth vanishing points.

\item We re-implement an enhanced version of \cite{liu2013grasp} by a new two-stage {\bf RE}currence for {\bf SC}ene {\bf U}nderstanding (RESCU) architecture, with 
an unsupervised stage-I (Sec.~\ref{sec:UPRD-I}) followed by a self-supervised stage-II (Sec.~\ref{sec:UPRD-II}). Both stages outperform the baseline method \cite{liu2013grasp} on our 1K+ RP benchmark image set (Sec.~\ref{sec:evaluation}).
\item We introduce three novel downstream URPD applications for 3D scene understanding:
\begin{enumerate}
    \item Vanishing Point Detection (VPD)  without straight line requirements (Sec.~\ref{sec:VP});
    \item 3D translation symmetry prediction (Sec.~\ref{sec:translation_sym});
    \item RP instance counting (Sec.~\ref{sec:object_counting}).
\end{enumerate}
These are quantitatively evaluated and compared with existing methods.
In addition, we qualitatively illustrate how the three downstream URPD outcomes on scene understanding can enhance image captioning (Sec.~\ref{sec:captions}) to achieve a more precise geometric and quantitative scene understanding from a single image. 
\end{itemize}
\vspace{-\topsep}


\section{Related Work}

\subsection{Recurring Pattern vs. Object Discovery}
\begin{table}[b!] \centering
\vspace{-15pt}
\caption{\small  Comparison of recurring pattern discovery approaches.}

\begin{adjustbox}{width=\textwidth}
\begin{tabular}{|c|c|c|c|c|c|}
\hline
Approach & Matching Strategy & Unsupervised & Partial Matching & \#Input images & Output  \\
\hline
\hline
\cite{cho2010reweighted,rother2006cosegmentation,toshev2007image,kannala2008object} & Pairwise-Object & Yes & N/A &$=2$ & 2 Objects \\
\cite{yuan2007spatial,cho2008co,liu2010common} & Pairwise-Object& Yes & N/A & $\le2$ & 2 Objects \\
\cite{cho2009feature,cho2010reweighted,cho2010unsupervised} & Pairwise-Object& Yes & Yes & $\ge 1$ & $2-10$ Objects \\
\cite{gao2009unsupervised}& Pairwise-Visual Word& Yes  & No & $\ge1$ & $\ge 1$ RPs\\
\cite{rubinstein2013unsupervised,hong2014unsupervised,faktor2013clustering,sivic2008unsupervised,todorovic2008unsupervised,fergus2003object} & Cross-Image & Yes & Yes  &$>>2$ & N/A \\
\cite{vo2021large,vo2020toward} & Cross-Image & Yes & N/A  & $>>2$ & $\ge 1$ Objects \\
\cite{huberman2016detecting,lettry2017repeated,rodriguez2019automatic}& Pairwise-Visual Word& Yes & No  & $=1$ & $\ge 2$ RP instances  \\
\cite{geng2018fine} &  Pairwise-Visual Word\& Instance  & No & Yes & $=1$ &   $=1$ RP  \\
\cite{zhang2019objects} & Co-occurrence Features & No & N/A & $>>2$ &  $=1$ Object \\
\cite{shi2022represent} & Query Image \& Exemplars & No & Yes & $=1$ &  $\ge 2$ RP instances \\

\hline
(Ours) RESCU &  Pairwise-Visual Word \& Instance & Yes & Yes & $=1$ & $\ge 1$ RPs\\
 
\hline
\end{tabular}
\end{adjustbox}
\label{tab:compare}
\end{table}

Recurring pattern discovery (RPD) is commonly compared with co-recognition / segmentation of objects~\cite{rother2006cosegmentation,cho2008co,cho2010unsupervised}, unsupervised object discovery \cite{cho2015unsupervised,wei2019unsupervised} and object co-localization \cite{vo2021large,simeoni2021localizing}.
RPD is different from these object based approaches, which aim at localizing objects of the same known category co-occurring across multiple images.
RPD adapts a broader definition of co-occurring instances that are not necessarily complete objects, and can find recurring instances from a single image. Many co-localization methods \cite{vo2021large,simeoni2021localizing} require a collection of unlabeled or semi-labeled images seen for common object discovery, while RPD is a zero-shot method.
RPD is also different from repeated pattern detection approaches \cite{huberman2016detecting,lettry2017repeated,rodriguez2019automatic}, which aims at detecting the pattern that are periodically repeated without any distortion. RPD aims at finding recurrence in a more relaxed notion than symmetry.

Tab.~\ref{tab:compare} shows a summary and comparison of different methods in terms of
input/output and matching strategy.
From Tab.~\ref{tab:compare}, \cite{liu2013grasp} and our approach are the only methods that discover class agnostic recurring pattern from a single image in an unsupervised fashion. Thus we treat \cite{liu2013grasp} as our baseline. 
However, for RP instance counting evaluation, we also compare with \cite{geng2018fine} on the same dataset. 
It should be pointed out that \cite{geng2018fine} requires a given localization of {\em logo} regions in grocery product detection scenario (equivalent to giving an RP instance as input) before the RP discovery.

Our URPD method differs from all the existing methods in that we have no pre-assumption on the types of RPs to be found, except {\em something in the image recurs}. We experiment with both hand-crafted features (i.e., SIFT in stage-I) and deep features (i.e., CNN activations in stage-II), with a set of effective search strategies. 
Our stage-II is different from self-supervised approaches like \cite{noroozi2016unsupervised,misra2020self,pathak2016context}, which propose various pretext tasks to generate pseudo labels automatically from the pretext tasks including image puzzling \cite{noroozi2016unsupervised}, inpainting \cite{pathak2016context}, etc.
In stage-II, we obtain generated RPs from stage-I as pseudo labels, to self-train a model to search for more RP-instances.

\subsection{Recurring Pattern vs. Object/Instance Dataset}
Most existing datasets are not suitable for URPD evaluation because they either fail to group instances based on visual similarity or lack of variety.
Widely used object detection datasets \cite{lin2014microsoft,everingham2010pascal} have instance-level annotations categorized by object class, like person, cats, bottles, but do not label instances by visual similarity. 
Datasets like \cite{george2014recognizing,hsieh2017drone} are designed for object detection in the specific domains of grocery product and car counting, respectively, and thus lack variety.
The data in \cite{cordts2016cityscapes} contains both semantic and instance-wise annotations, and semantic instances belonging to the same class can be viewed as RP instances in many cases. However, the dataset is limited to city street-view images. 

We introduce a new benchmark dataset of $1K+$ images (RP-1K) collected from a wide range of sources and viewpoints. Each image is manually labeled as containing single/multiple RPs. For each RP, the multiple RP instances are labeled with bounding boxes or finer-grained contours (Sec.~\ref{sec:collectdataset}).

\subsection{Scene Understanding}
3D scene understanding, briefly summarized in Table \ref{tab:3DSU}, has made substantial progress in the recent years. But little work has been done in trying to understand a 3D scene in the context of certain fundamental aspects such as vanishing point and repeating patterns that can be used as powerful tools to understand the geometry of the 3D scene. Our method is significantly different from the aforementioned methods in that we try to infer the 3D scene aspects by merely using the RPs to detect vanishing points and predicting translation symmetry in the 3D scene from a single image.

\begin{table}[h!] \centering
\vspace{-15pt}
\caption{\small  Summary of 3D Scene Understanding Methodologies.}

\begin{adjustbox}{width=\textwidth}
\begin{tabular}{|p{0.2\linewidth}|p{0.8\linewidth}|}
\hline
Literature & Applications   \\
\hline
\hline
\cite{patil2022p3depth,li2022binsformer,yan2021channel} & Performs depth estimation in the context of scene understanding.\\
\cite{wang2022spatiality,azuma2022scanqa} & Dense captioning involving object-level 3D scene understanding.\\
\cite{xu2022groupvit,lambert2020mseg} & Performs Semantic Segmentation for visual scene understanding.\\

\hline
(Ours) RESCU &  Prediction of Vanishing point, inferring existence of  translation symmetry,counting RP-instance and enhancing scene captioning.\\
 
\hline
\end{tabular}
\end{adjustbox}
\vspace{-5pt}
\label{tab:3DSU}
\end{table}

\subsubsection{Vanishing Point Detection}
%
Several algorithms have been developed to find vanishing points (VPs) from a single image \cite{rother2002new,shi2015fast,zhou2017detecting,zhou2019neurvps}, which are heavily dependent on explicit line segments. 
\COMMENT{
Most of these algorithms, such as Zhou et al's method \cite{zhou2017detecting}, perform edge or line detection in order to detect a VP as the intersection of (infinitely extended) lines.
}
Recently, \cite{zhou2019neurvps} present a network based on a novel {\it conic convolution} operator that efficiently evaluates support across the image for hypothesized VP directions sampled on the Gaussian sphere, yet support is still based on strength of the response of learned edge filters oriented locally towards the VP. However, in real-world images,  straight line or edge evidence may not always be explicit: e.g. Fig.~\ref{fig:1-d} where lines leading to a VP are implicit.
\COMMENT{
formed by corresponding SIFT feature points along the street lights.
}
In this paper, we attempt to detect and quantitatively validate VPs by discovering these {\em implicit} lines using feature correspondences among RPs. 

\vspace{-0.3cm}
\subsubsection{Translation Symmetry Detection}
\COMMENT{
Yet another downstream application that can enhance the understanding of salience and geometry in 3D scenes is discovery of 3D translation symmetry from the 2D image.
}
RP instances that are co-linear frequently also exhibit {\it translation symmetry}, meaning adjacent pairs of instances have equal distance between them along a line in 3D. Translation symmetry has been used as a constraint for  perceptual grouping \cite{Schaffalitzky00a} and to recover 3D structure \cite{mundy1993repeated}. RPs with translation symmetry in 3D space maintain a mathematical relationship with their corresponding 2D image projections, defined by the cross-ratio \cite{mundy1992projective}. Once the are discovered, these RPs can be rectified to remove perspective foreshortening, revealing a simpler, ``frontal" appearance. 

\vspace{-0.3cm}
\subsubsection{Counting Problem}
RPD can benefit scene understanding by using number of detected  RP instances to estimate the count of recurring instances in an image (e.g., grocery product counting \cite{geng2018fine}). 
Typical supervised object counting approaches \cite{goldman2019precise,lin2017focal} may not perform well in unseen scenes. 
Class-agnostic and few-shot methods ~\cite{lei2021towards,lu2018class,ranjan2021learning,shi2022represent,hobley2022learning} often rely on references as input or are limited to single-class counting.
Our method is able to discover and count multiple unseen RP instances and does not require any additional input.

Repetition counting in video domain \cite{zhang2021repetitive,dwibedi2020counting} is typically a class-agnostic task, where the repetitive activity can be any movement happened in real world. 
Repetition counting can be viewed as RP instance counting in spatial-temporal domain.
In this paper, we focus on the application of RP discovery from 2D images and enhance a better understanding of 3D scene.


\subsection{Image Captioning}
Image captioning is a long standing task in computer vision which can provide a semantic understanding for a given image \cite{stefanini2022show}. Commonly, efforts in deep learning-based image captioning have used pre-trained object detection networks \cite{ren2015faster} to identify salient image regions to caption \cite{anderson2018bottom}. Additionally, visual question answering (VQA) is a well explored research area in computer vision and natural language processing. VQA systems strive to correctly answer natural language questions about a given image. Recent work in unified vision-language models has demonstrated state-of-the-art results in both image captioning and VQA tasks through exploiting learned features from both visual and language modalities \cite{li2020oscar,wang2022unifying,zhou2020unified}. Despite these advances, current datasets frequently used in image captioning and VQA pre-training such as COCO do not contain a large number of RP instances we are interested in. Thus, we propose utilizing our discovered RP information from images to both enhance image captions and create new image-caption pairs with existing of-the-shelf models.

\section{Methodology}
\begin{figure*}[!ht] \centering
    \vspace{-5pt}
    \includegraphics[width=0.9\linewidth]{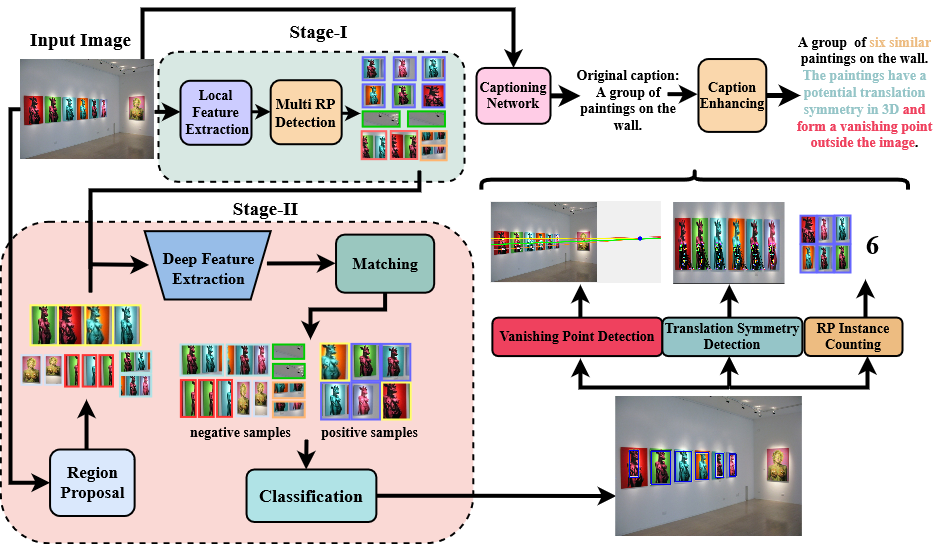}

    \caption{\small  \textbf{Overview of our proposed two-stage method.} RESCU is comprised of two primary stages. \textbf{Stage-I}. We extract features from the image to perform unsupervised RP detection and obtain a set of candidate RP(s) and their corresponding region proposals. \textbf{Stage-II}. Candidate RP image crops discovered in Stage-I are passed through a frozen pre-trained feature extractor to obtain their corresponding feature representations. Region proposals derived from the original image are sent through the same extractor to obtain a second set of representations. We apply clustering to the features obtained from the region proposals and RP crops to perform matching and ultimately derive sets of positive and negative samples to train the final RP classifier. The predicted RP(s) are then used in \textbf{three downstream tasks}: RP instance counting, vanishing point detection, and translation detection. We show the discovered information may be used to enhance image captions.}
    \vspace{-5pt}
    \label{fig:overview}
\end{figure*}

Fig.~\ref{fig:overview} illustrates our overall approach with an image example. There are two major stages of Unsupervised Recurring Pattern Discovery (URDP): Stage-I performs initial, multi-threaded detection of one or more RPs, and Stage-II uses RP instances found in Stage-I to train a CNN classifier to extend each RP by finding additional RP instances.  Final RPs output from Stage-II are then used in three downstream vision tasks: 1) vanishing point detection, 2) translation symmetry detection, 3) RP instance counting. The outputs of downstream tasks ultimately enhance image captioning for a better 3D scene understanding.

\begin{figure}[b!]
    \vspace{-5pt}
    \begin{subfigure}[t]{0.49\linewidth}
    \centering
        \includegraphics[width=\textwidth, height=0.5\linewidth]{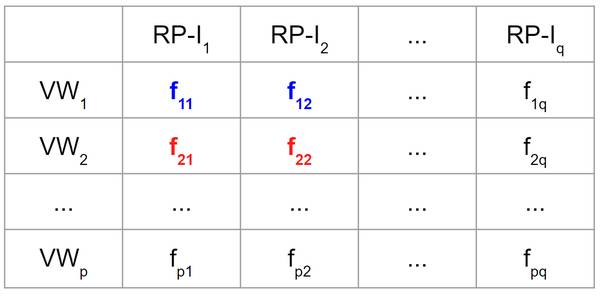}
         \vspace{-15pt}
         \caption{\small \scriptsize{an RP Matrix} \label{fig:rp_matrix}}
    \end{subfigure}
    \hfill    
    \begin{subfigure}[t]{0.49\linewidth}
    \centering
        \includegraphics[width=\textwidth, height=0.5\linewidth]{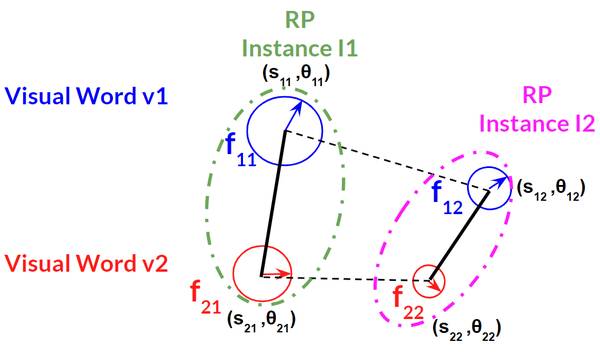}
         \vspace{-15pt}
         \caption{\small \scriptsize{ Unit Recurring Pattern (URP)} \label{fig:basic_rp_structure}}
    \end{subfigure}
    \vspace{-5pt}
    \caption{\small 
    \textbf{(a)} an RP matrix, rows are features belonging to the same visual word, and columns are features belonging to the same RP instance.
    \textbf{(b)} a Unit Recurring Pattern (URP) is formed from the RP matrix by choosing two visual words (\textit{rows}) and two RP instances (\textit{columns}) to select four visual primitives, e.g.~local SIFT features, $(f_{11},f_{12},f_{21},f_{22})$, each of which has a scale $s$ and angle $\theta$. 
    }
    \vspace{-5pt}
\end{figure}

\subsection{RESCU Stage-I: Multiple RP Discovery}
\label{sec:UPRD-I}

\subsubsection{Recurring Pattern Discovery}
\label{sec:rp_representation}

Similar to \cite{liu2013grasp}, we denote a {\em recurring pattern} mathematically as a $M\times N$ matrix $\mathbf{\Omega}$, where each row corresponds to a distinct {\em visual word} and each column corresponds to a distinct {\em RP instance} (Fig.~\ref{fig:rp_matrix}).
We define each 2x2 sub-matrix $\mathbf{\Omega}_{2,2}$, formed by choosing 2 rows $m_1, m_2$ and 2 columns $n_1, n_2$ of $\mathbf{\Omega}$ as a {\em Unit Recurring Pattern} (URP).  A URP is the smallest recurring pattern (Fig.~\ref{fig:basic_rp_structure}). 
The affinity score $u$ of a URP $\mathbf{\Omega}_{2,2}$ is measured by
\begin{align}
    u(m_1,m_2,n_1,n_2|\mathbf{\Omega}) = \exp (-\Delta_s^2 - \Delta_{\theta}^2)
    \label{eq:affinity_score}
\end{align}
where $\Delta_{s}$ and $\Delta_{\theta}$ measure normalized scale and angular differences, defined as follows:

Following \cite{liu2013grasp}, we first define URP instance size ratio $r = \frac{d(f_{11} - f_{21})}{d(f_{12} - f_{22})}$, where $d(\cdot)$ is the Euclidean distance between two features of an URP instance.

We then define $D_s(f_i, f_j)$ as the normalized scale difference by RPI size ratio $r$ of two SIFT features $f_i$ and $f_j$, as follows:
\begin{align}
    D_s(f_i, f_j) &= \frac{s_i - r\cdot s_j}{s_i + r\cdot s_j}
\end{align}
where $s_k$ is the scale of feature $f_k$.

Similarly, we define $D_\theta(f_i, f_j)$ as the normalized angle difference by RPI size ratio $r$ of two SIFT features $f_i$ and $f_j$, as follows:
\begin{align}
    D_\theta(f_i, f_j) &= \frac{{\theta}_1 - r\cdot {\theta}_2}{{\theta}_1 + r\cdot {\theta}_2}
\end{align}

Finally, we define $\Delta_{s}$ as the largest normalized scale difference:
\begin{align}
    \Delta_{s} = \max(D_s(f_{11}, f_{12}), D_s(f_{21}, f_{22}))
\end{align}

We define $\Delta_{\theta}$ as the largest normalized angle difference:
\begin{align}
    \Delta_{\theta} = \max(D_{\theta}(f_{11}, f_{12}), D_{\theta}(f_{21}, f_{22}))
\end{align}

Unsupervised recurring pattern discovery (URPD) becomes an optimization problem for $\mathbf{\Omega}^* = \arg\max_{\mathbf{\Omega},M,N}\{U(\mathbf{\Omega}_{M,N})\}$ where

\begin{eqnarray}
U(\mathbf{\Omega}_{M,N}) = \frac{1}{M\cdot N}\sum_{\substack{m_1,m_2=1..M\\n_1,n_2=1..N}}u(m_1,m_2,n_1,n_2|\mathbf{\Omega}).
\label{eq:optG}
\end{eqnarray}

\subsubsection{URP-based Search}
\label{sec:minimum_rp_extraction}
To optimize Eq.~\ref{eq:optG}, the baseline approach~\cite{liu2013grasp} adopts the GRASP~\cite{feo1995greedy} optimization with randomly selected 2 visual words as initial. 
For robustness on various types of images, more initials can benefit the approach on discovering RPs. 
However, \cite{liu2013grasp} is lack of scalability to multiple initials due to computation expense which leads to a prolonged searching time.

We propose a URP-based approach to improve the robustness and scalability of \cite{liu2013grasp} as follows:
In each detection iteration, our approach maintains an RP matrix for each initial as Fig.~\ref{fig:rp_matrix} shows. The algorithm expend the URP (as a $2\times2$ initial matrix) with two movement directions: (1) add/remove a column, (2) add/remove a row.

Without losing generalizability, consider the current RP as $\Omega_{p,q}$ with $p$ visual words, $q$ RP instances as Fig~\ref{fig:rp_matrix} shows.
To add a column for one more instance is to add a series of features $\{f_{1,q+1}, \allowbreak f_{2,q+1}, \cdots, f_{p,q+1}\}$ into column $(q+1)$, to form RP $\Omega_{p,q+1}$.
To find the candidate feature $f_{i,q+1}$ to be put to visual word $i$, we define the visual word affinity sum-up score of $f_{i,q+1}$ respect to $\Omega_{p,q+1}$, as follows: 
\begin{align}
    \mathbf{V}(f_{i,q+1}|\Omega_{p,q+1}) &= \sum_{j\in [1,q]} V(f_{i,q+1}, f_{i,j})\\
    V(f_{i,q+1}, f_{i,j}) &= \sum_{\substack{k\in [1,p],\\k\neq i}} u(f_{i,q+1}, f_{i,j}, f_{k,q+1}, f_{k,j})\\
\end{align}
hence,
\begin{align}
    \mathbf{V}(f_{i,q+1}|\Omega_{p,q+1}) = \sum_{j\in [1,q]} \sum_{\substack{k\in [1,p],\\k\neq i}} u(f_{i,q+1}, f_{i,j}, f_{k,q+1}, f_{k,j})
\end{align}
$\mathbf{V}(f_{i,q+1}|\Omega_{p,q+1})$ measures the overall affinity of URPs containing $f_{i,q+1}$ in $\Omega_{p,q+1}$. A candidate feature $f_{i,q+1}$ that maximize $\mathbf{V}(f_{i,q+1} |\Omega_{p,q+1})$ will be added to row $i$, column $q+1$ of RP $\Omega_{p,q+1}$. 
Similarly column-wise.

To reduce computation cost for each initial, we compute pair-wise affinity (Eq.~\ref{eq:affinity_score}) of all feature candidates prior to the optimization searching. 
The quantitative comparison shows that our method advanced the baseline in all metrics (Tab.~\ref{tab:rp_eval}).

\vspace{-1.3em}
\subsubsection{Adaptive Parameter}
\label{subsec:para scanning}
To improve stage-I performance on various type of images from \cite{liu2013grasp}, we introduce three hyper-parameters $P_d, P_s, P_{\theta}$ as adaptive candidate constraints in URP-based search.
The three hyper-parameters control the maximum feature distance, maximum feature size ratio, and maximum feature angle difference.
$P_d$ sets the maximum feature distance, which controls the granularity of RP detection.
$P_s$ sets the maximum size difference among RP-instances.
And $P_{\theta}$ sets the maximum orientation difference among objects. 
The hyper-parameters are adapted for each input image, by a grid search to find the most suitable parameter set that leads to the maximal $\mathbf{\Omega}^*$ in Eq. \ref{eq:optG}.
By adaptive parameter, the stage-I performance can be improved. 
See Sec.~\ref{sec:ablation_study} for ablation study.

\subsection{RESCU Stage-II: RP Instance Extension}
\label{sec:UPRD-II}
Stage-I sometimes suffers from missing RP instances. Hence, we propose a self-supervised stage-II for additional RP instance discovery as Fig.~\ref{fig:overview} shows. 

In stage-II, we first use an off-the-shelf region proposal approach \cite{zitnick2014edge} to obtain a large set of proposed regions from the single image. The region proposal is designed to extend the pseudo positive \& negative samples for self-learning purpose. 
We use a feature extractor backbone \cite{huang2017densely} to extract deep features of each RP instance from multiple RPs detected by stage-I, together with the proposed regions.
In deep feature space, the proposed regions are matched with RP instances based on a clustering method DBSCAN \cite{ester1996density}, for its robustness against outliers.
The proposed regions with feature closer to the cluster centers of RP instances, together with RP instances detected by stage-I, will be considered as positive samples. 
The proposed regions with feature further to those cluster centers will be considered as negative samples.

\begin{figure}[tb!]
    \centering
    \begin{subfigure}[t]{0.22\linewidth}
    \caption{\small Original Image}
    \includegraphics[width = \textwidth,height=0.8\textwidth, valign=t]{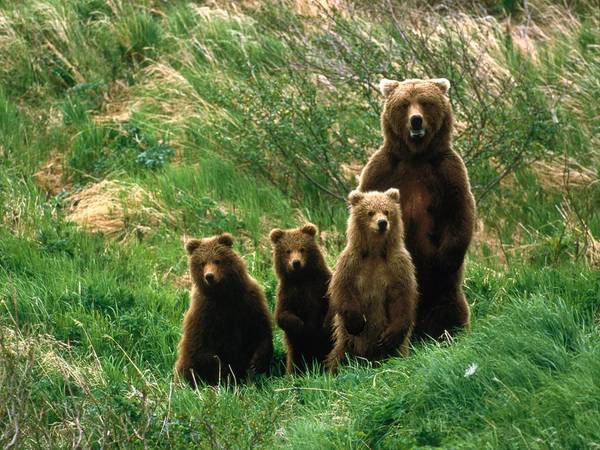}
    \end{subfigure}
    \hfill
    \begin{subfigure}[t]{0.22\linewidth}
    \caption{\small RESCU-I}
    \includegraphics[width = \textwidth,height=0.8\textwidth, valign=t]{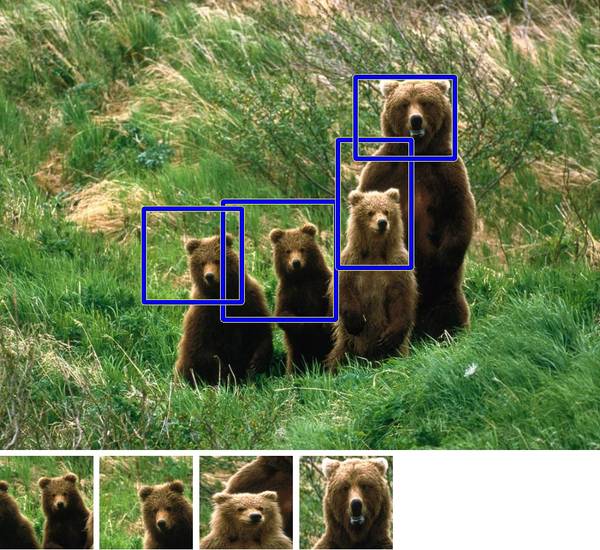}
    \end{subfigure}
    \hfill
    \begin{subfigure}[t]{0.22\linewidth}
    \caption{\small Positive}
    \includegraphics[width = \textwidth,height=0.8\textwidth, valign=t]{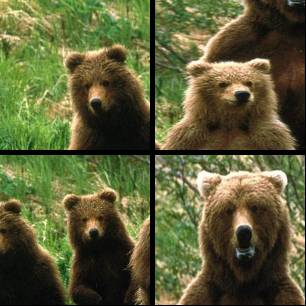}
    \end{subfigure}
    \hfill
    \begin{subfigure}[t]{0.22\linewidth}
    \caption{\small Negative}
    \includegraphics[width = \textwidth,height=0.8\textwidth, valign=t]{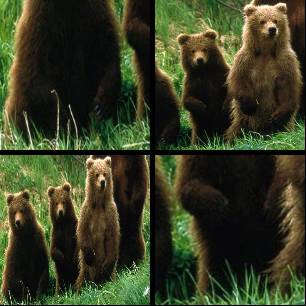}
    \end{subfigure}
    \vfill
    \begin{subfigure}[t]{0.22\linewidth}
    \includegraphics[width = \textwidth,height=0.8\textwidth ,valign=t]{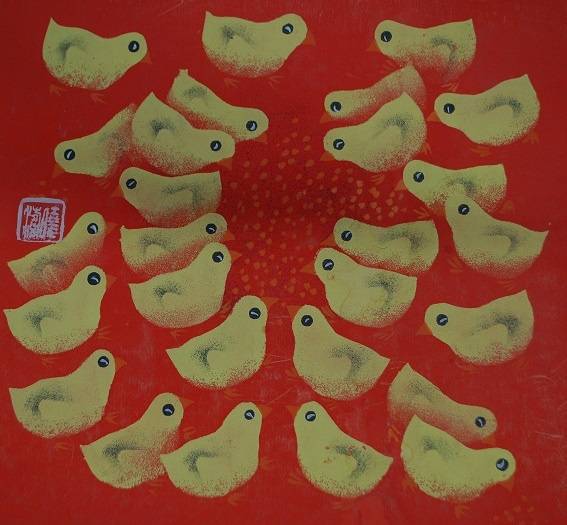}
    \end{subfigure}
    \hfill
    \begin{subfigure}[t]{0.22\linewidth}
    \includegraphics[width = \textwidth, height=0.8\textwidth,valign=t]{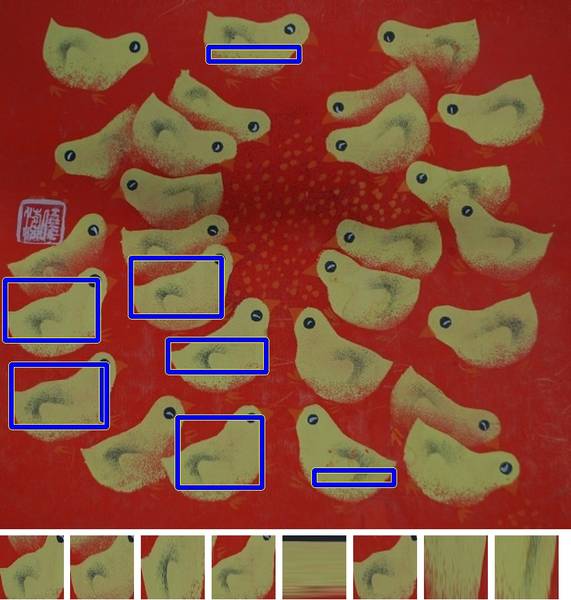}
    \end{subfigure}
    \hfill
    \begin{subfigure}[t]{0.22\linewidth}
    \includegraphics[width = \textwidth,height=0.8\textwidth, valign=t]{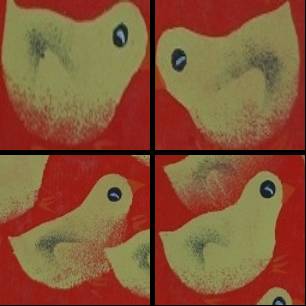}
    \end{subfigure}
    \hfill
    \begin{subfigure}[t]{0.22\linewidth}
    \includegraphics[width = \textwidth, height=0.8\textwidth, valign=t]{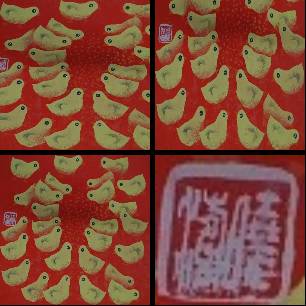}
    \end{subfigure}
    
    \caption{\small RESCU-II positive \& negative samples of some images. 
    \textbf{(a)} Original Image.
    \textbf{(b)} RESCU-I output.
    \textbf{(c)} Positive samples generated by RESCU-II.
    \textbf{(d)} Negative samples generated by RESCU-II.
    }
    \label{fig:pos_neg}
\end{figure}

In practice, we select the top $k=N\times P$ nearest proposals to the cluster centers of RP instances, where $N$ is the number of cluster centers and $P$ is an empirical parameter, set to $40$. 
By this procedure, we maintain $<10\%$ of the total proposals generated by the off-the-shelf \cite{zitnick2014edge}. 
Fig.~\ref{fig:pos_neg} shows some example of generated positive \& negative samples.

\begin{figure}[tb!]
    \centering
    \includegraphics[width=0.5\textwidth]{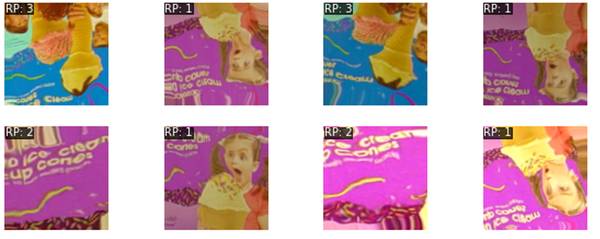}
    \caption{\small RESCU-II local Deformation Showcase. Each sample is applied with various deformation factors.}
    \label{fig:local_deformation}
\end{figure}

We apply data augmentations to enlarge positive samples for a better training. 
Besides the general global transformations applied for augmentation, we also apply local appearance \& geometric deformation based on corresponding features of RP instances, as Fig.~\ref{fig:local_deformation} shows.
Finally, we apply DenseNet \cite{huang2017densely}, pretrained on Image-Net \cite{deng2009imagenet}, as the backbone of classifier and append 3 dense layers. We supply implementation details in the Appendix.
For each image, it is self-supervised with the obtained positive \& negative samples.

\section{Experiments \& Results}
\subsection{RP-1K Dataset \label{sec:collectdataset}}

\begin{figure}[b!] \centering
\vspace{-5pt}
\foreach \id\picname in {1/new682, 2/new685, 3/new677, 4/new036, 5/new531, 6/new680, 7/new025, 8/new026}
{ %
\begin{subfigure}[t]{0.22\linewidth}
\includegraphics[height=\textwidth, width = \textwidth,valign=t]{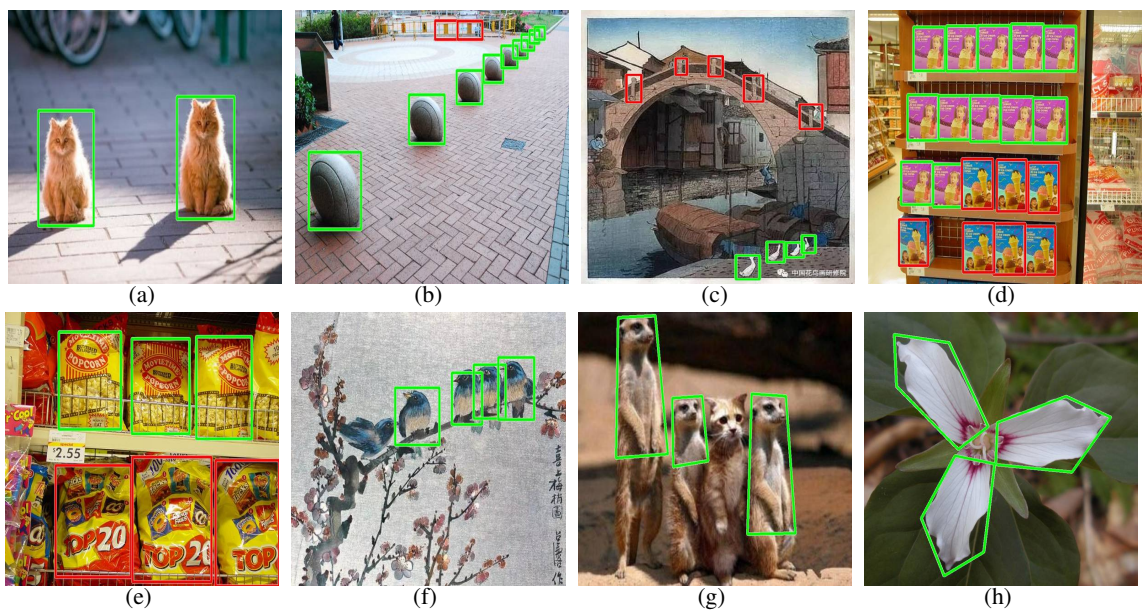}
\vspace{-7pt}
\caption{\small }
\end{subfigure} 
}
\caption{\small  View of the Scene Examples: \textbf{(a)}Frontal, \textbf{(b)}Projective, and \textbf{(c)}Unknown. RP type Examples: \textbf{(d)}Man-made rigid, \textbf{(e)}Man-made deformable, \textbf{(f)}Painting, \textbf{(g)}Animal/human, and \textbf{(h)}Others. 
The colored contours are human labeled Ground Truth. Zoom in for a more detailed viewing.}
\vspace{-5pt}
\label{fig:dataset_ex}
\end{figure}

The statistics on the Recurring Pattern dataset are in Tab.~\ref{tab:RP_dataset}.
The RP-1K dataset is collected from several resources: 
83 images from \cite{liu2013grasp}, 
150 images containing perspective deformation from \cite{zhou2017detecting}, 769 images from \cite{zhang2009handling} and the rest of the images are collected by the authors. 
Each image is manually labeled using a customized online labeling tool. See sample labeled RPs in Fig.~\ref{fig:dataset_ex}.

\begin{table*}[b!] \centering
\vspace{-15pt}
\caption{\small  Statistic of \textbf{RP-1K} (with all subsets) and \textbf{Grozi-3.2K}}
\label{tab:RP_dataset}
\resizebox{\linewidth}{!}{%
\begin{tabular}{c|c|c|c|c|c|c|c}
    \hline
    &  &  &  & \multicolumn{2}{c|}{\#RP per Image} &  \multicolumn{2}{c}{\#Instance per RP} \\ \cline{5-8}
    \textbf{Camera View / RP Category} &  \# Images &  \# RPs &  \# Instances & Max / Min & Mean $\pm$ STD & Max / Min & Mean $\pm$ STD\\ \hline \hline 

    \textbf{Front view         } &       789 &   4279 &        16353 &     24 / 1 &  5.42 $\pm$ 4.04 &           96 / 1 &       3.82 $\pm$ 3.78 \\
    \textbf{Projective view    } &       215 &    476 &         2675 &     17 / 1 &  2.21 $\pm$ 2.50 &           75 / 2 &       5.62 $\pm$ 6.24 \\
    \textbf{Unknown            } &        20 &     22 &          174 &      2 / 1 &  1.10 $\pm$ 0.30 &           28 / 2 &       7.91 $\pm$ 5.59 \\ \hline
    
    \textbf{Man-made rigid     } &       813 &   3920 &        16029 &     24 / 1 &  4.82 $\pm$ 4.09 &           96 / 1 &       4.09 $\pm$ 4.37 \\
    \textbf{Man-made deformable} &       156 &    797 &         2741 &     16 / 1 &  5.11 $\pm$ 3.43 &           28 / 1 &       3.44 $\pm$ 2.24 \\
    \textbf{Painting           } &        22 &     26 &          201 &      3 / 1 &  1.18 $\pm$ 0.49 &           28 / 2 &       7.73 $\pm$ 5.60 \\
    \textbf{Animal/human       } &        27 &     28 &          153 &      2 / 1 &  1.04 $\pm$ 0.19 &           31 / 2 &       5.46 $\pm$ 6.06 \\
    \textbf{Others             } &         6 &      6 &           78 &      1 / 1 &  1.00 $\pm$ 0.00 &           25 / 3 &      13.00 $\pm$ 8.66 \\ \hline  \hline

    \textbf{Total of \textbf{RP-1K} } &  \textbf{1024} &   \textbf{4777} &        \textbf{19202} &     \textbf{24 / 1} &  \textbf{4.67 $\pm$ 3.98} &           \textbf{96 / 1} &      \textbf{ 4.02 $\pm$ 4.15} \\ \hline \hline
    \textbf{Grozi-3.2K \cite{george2014recognizing}} &       677 &   2302 &         8265 &     10 / 1 &  3.40 $\pm$ 1.62 &           26 / 2 &       3.59 $\pm$ 2.22 \\\hline
    
\end{tabular}
 }
\end{table*}

\subsection{Evaluating Unsupervised Recurring Pattern Discovery}
\label{sec:evaluation}

\subsubsection{Evaluation Metric}
We propose two evaluation measurements at Recurring Pattern (RP) and Recurring Pattern Instance (RPI) level respectively : \\
\noindent{\bf 1) RP instance level:}
%
We propose an {\bf intersection-over-detection (IOD)} metric as follows.
If $RPI_i$ is a detected RPI, and $RPI_{GT_j}$ a groundtruth RPI.
$RPI_j$ is considered acceptable if and only if 
$(RPI_i\cap RPI_{GT_j})/RPI_i > h$, where $h$ is a numerical threshold. 
Given a detected RP with a set of RPIs $\mathbf{RPI_D}$, and some ground truth RP with a set of ground truth RPIs $\mathbf{RPI_{GT}}$, and the set of acceptable RPIs denoted as $\mathbf{RPI_A}$, \textbf{RP Instance level} precision $P_I$ and recall $R_I$ rates are defined as:
$P_I = |\mathbf{RPI_A}|/|\mathbf{RPI_D}|$,\ $R_I = |\mathbf{RPI_A}|/|\mathbf{RPI_{GT}}|$

\begin{figure}[tb!] \centering
\begin{subfigure}[t]{0.23\linewidth}\centering
    \includegraphics[width=\textwidth, height=\textwidth]{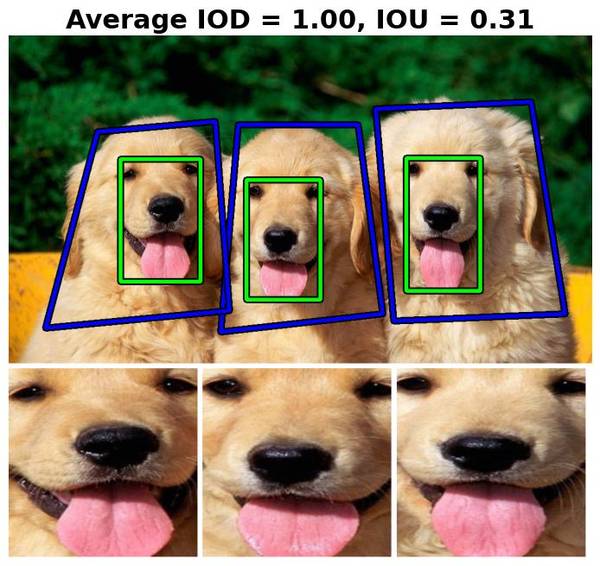}
    \caption{\small \label{fig:iod_high_ex1}}
\end{subfigure}
\begin{subfigure}[t]{0.23\linewidth}\centering
    \includegraphics[width=\textwidth, height=\textwidth]{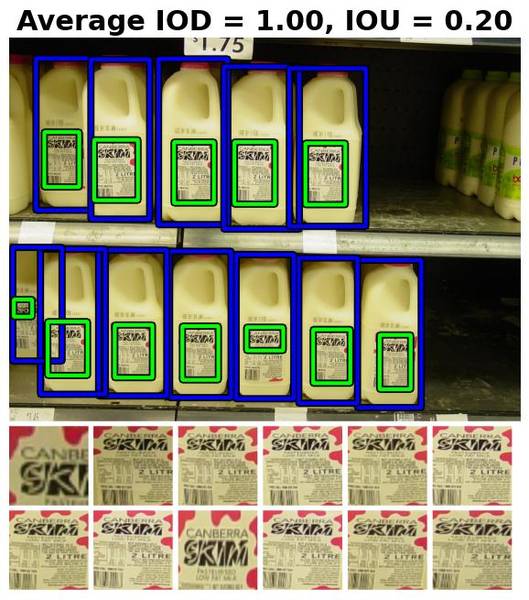}
    \caption{\small \label{fig:iod_high_ex2}}
\end{subfigure}
\begin{subfigure}[t]{0.23\linewidth}\centering
    \includegraphics[width=\textwidth, height=\textwidth]{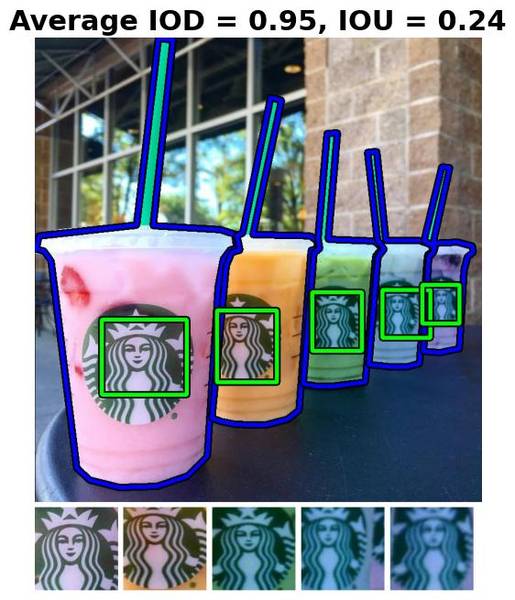}
    \caption{\small \label{fig:iod_high_ex3}}
\end{subfigure}
\begin{subfigure}[t]{0.23\linewidth}\centering
    \includegraphics[width=\textwidth, height=\textwidth]{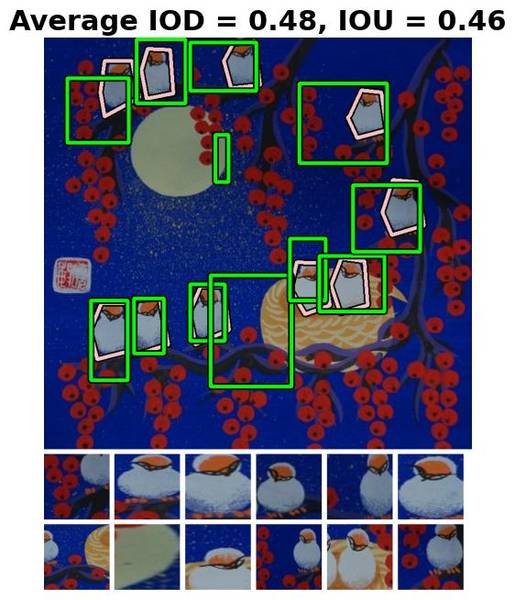}
    \caption{\small \label{fig:iod_low_ex1}}
\end{subfigure}

\caption{\small  \textbf{(a,b,c)}: RPD results on real image examples where both IOD and IOU scores are shown. It is clear that IOD captures the success of RP detection on each image correctly. The partial overlaps detected by RPD are usually the most interesting/complex sub-parts of the objects that recur. (d) demonstrates that large bounding boxes do not hype up IOD scores either.
\textcolor{green}{Green}: detected RPs. \textcolor{blue}{Blue} (except (d) \textcolor{pink}{pink}): GT RPs.}
\vspace{-5pt}
\label{fig:IOD_vs_IOU}
\end{figure}

Precise segmentation of an object is not the goal of RPD, though human labels (GT RPs) tend to maximize the boundary of each object (Fig.~\ref{fig:IOD_vs_IOU}) as an RPI.
To evaluate RPD output properly, a metric should reward consistently detected overlaps with GT RPs even if the overlaps are partial. The IOD metric captures this correctly while IOU does not (Fig.~\ref{fig:iod_high_ex1}, \ref{fig:iod_high_ex2}, \ref{fig:iod_high_ex3}).

%
We report the evaluation results with threshold $h=0.5$ in Sec.~\ref{sec:quantiative_validation},\ref{sec:ablation_study}. See Appendix Fig.~\ref{fig:eval_curves} for a detailed study on altering $h$.

\noindent{\bf 2) RP level:}
%
For a detected RP $RP_D$ 
%
and all ground truth RPs $\mathbf{RP_{GT}}$, the one $RP_{GT}$ with the 
{\em highest RP instance level precision $P_I$} 
is assigned to $RP_D$.
%
%
$RP_D$ is considered as an \textit{accepted} RP, denoted as $RP_A$, if it is assigned to a $RP_{GT}$.  
Given multiple RP detections of a single image $\mathbf{RP_D}$, the ground truth RPs of the same image $\mathbf{RP_{GT}}$, and the set of \textit{accepted} RP detections $\mathbf{RP_A}$.
\textbf{RP level} precision $P_{RP}$ and recall rates $R_{RP}$ are defined as:\\
$P_{RP}= |\mathbf{RP_A}|/|\mathbf{RP_D}|$, \ $R_{RP}= |\mathbf{RP_A}|/|\mathbf{RP_{GT}}|$.

\begin{figure}[b!] \centering
\vspace{-5pt}
\foreach \id/\picname in {1/new687, 2/new015, 3/new020, 4/new703}
{ %
    \begin{subfigure}[t]{0.46\linewidth}
    \includegraphics[height=0.5\textwidth, width = \textwidth,valign=t]{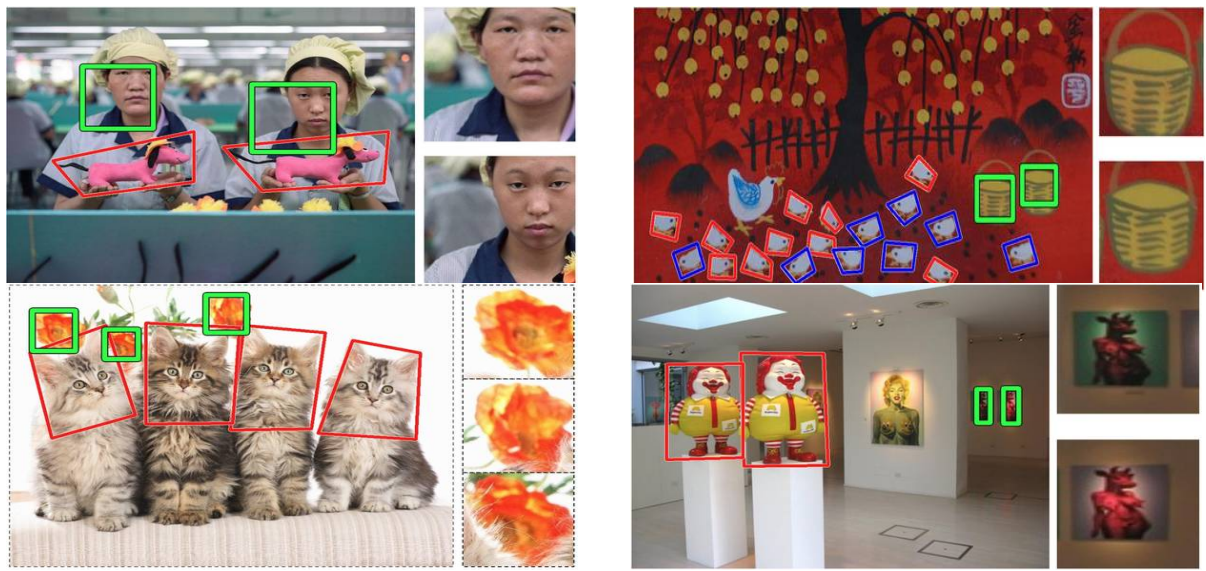} 
    \end{subfigure} 
    \hfill
    \vspace{-2pt}
} 
\caption{\small  Four sample outputs of our method demonstrate the algorithm's ability to discover RPs beyond human labels.
\textcolor{green}{Green} thicker boxes are the detected RP instances by our method, which are not labeled by human raters. The \textcolor{red}{red} \& \textcolor{blue}{blue} boxes are human labeled Ground Truth RPs.}
\vspace{-5pt}
\label{fig:unlabel_case}
\end{figure}

\subsubsection{Quantitative Validation}
\label{sec:quantiative_validation}
\begin{figure}[!ht]
\centering
\foreach \id/\picname in {1/new295, 2/new764, 3/new997, 4/new498, 5/new014, 6/new017}
{ %
    \foreach \x/\method in {1/GT,2/Baseline,3/URPD-I,4/URPD-II}
    { %
    \begin{subfigure}[t]{0.19\linewidth}
        \if \id 1
            \if \x 1 \caption{\small Ground Truth} \fi
            \if \x 2 \caption{\small Baseline \cite{liu2013grasp}} \fi
            \if \x 3 \caption{\small RESCU-I} \fi
            \if \x 4 \caption{\small RESCU-II} \fi
        \fi
        \includegraphics[width = \textwidth,valign=t]{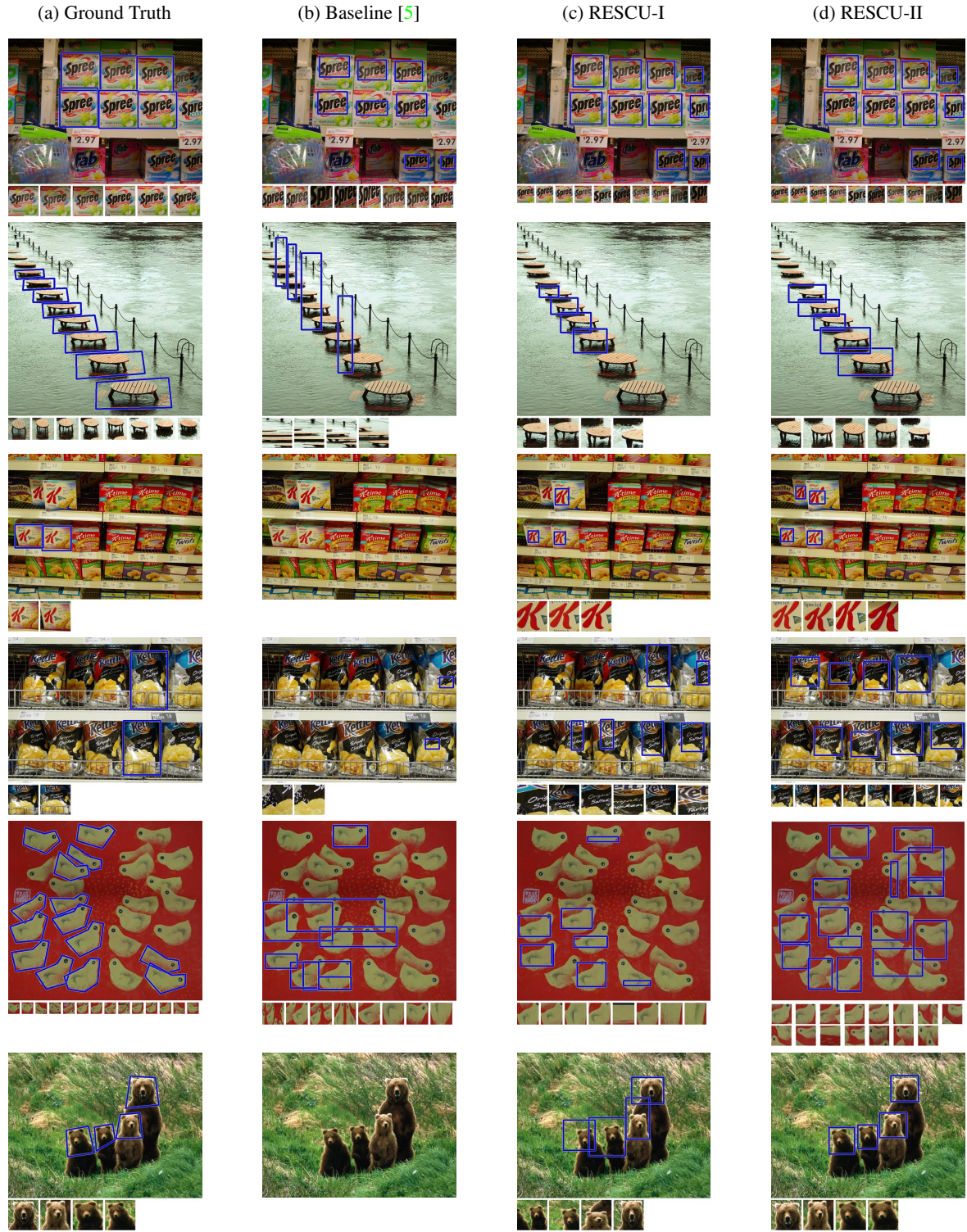} 
    \end{subfigure} 
    \hfill
    }
    \vspace{2pt}
    
} \vspace{-5pt}
\caption{\small Examples of RP detection results from each camera view/RP category.
\textbf{Top to Down:} Results broken out by subset categories: Front View, Projective View, Man-Made Rigid, Man-Made Deformable, Painting, Animal/Human subsets.}
\label{fig:enhanced_ex}
\end{figure}

%
Tab.~\ref{tab:rp_eval} shows the evaluation on the full \textbf{RP-1K} dataset as well as broken down into its various viewpoint and category subsets. 
Some RP detection examples from each subset category are shown in Fig.~\ref{fig:enhanced_ex}. All examples shown have \textbf{RESCU-II} performs better or as good as \textbf{RESCU-I}, and better than the baseline method \cite{liu2013grasp}.

Quantitative results of RP detections from Table~\ref{tab:rp_eval} show that our RESCU outperforms the baseline in both RP and RP Instance level, and especially on the Recall rate. 
RESCU with stage-II leads to a 2\% increase in RP Instance Recall from the baseline.
Moreover, our method is able to detect recurring patterns that are missed by human raters. Fig.~\ref{fig:unlabel_case} shows such examples. This observation indicates a limitation on manually labeled\textbf{ RP-1K} dataset, while also a promise to enhance human RP labels via computational tools such as RESCU.

However, the results from Table~\ref{tab:rp_eval} also reveal that our RESCU does not perform as well on the Projective view and Man-made deformable subsets. Images from these subsets are usually more challenging for detecting RPs due to severe geometric distortion, non-uniform lighting, and blurring. See Sec.~\ref{sec:limitations}.

\begin{table}[tb!]
\caption{\small  RP Discovery Evaluation on \textbf{RP-1K}.
The values in \textit{italic} are not statistically significant with range in the same column section.
The values in \textbf{bold} are the best mean/std with range in the same column section.
}
\label{tab:rp_eval}
\centering

\begin{tabular}{l|cc|cc} 
\toprule
 & \multicolumn{2}{c|}{RP Level} & \multicolumn{2}{c}{RP Instance Level} \\
Method &     Precision &        Recall &   Precision &      Recall \\
\midrule
\multicolumn{5}{c}{\textbf{The Whole RP-1K}}\\
\hline
\textbf{Baseline\cite{liu2013grasp}} &           0.40 $\pm$\textbf{ 0.37} &           0.32 $\pm$ 0.35 &           0.67 $\pm$ 0.41 &           0.47 $\pm$ 0.38 \\
\textbf{RESCU-I} & 0.44 $\pm$ \textbf{0.32} & \textit{0.50 $\pm$ 0.38} &  0.68 $\pm$ 0.33 &   0.62 $\pm$ 0.34\\
\textbf{RESCU-II} &  \textbf{0.45} $\pm$ \textbf{0.32} &  \textbf{0.52} $\pm$ 0.38 &  \textbf{0.71} $\pm$ \textbf{0.32} &  \textbf{0.64} $\pm$ \textbf{0.33} \\

\hline
\midrule
\multicolumn{5}{c}{\textbf{Front View Subset}}\\
\hline
\textbf{Baseline\cite{liu2013grasp}} &           0.45 $\pm$ 0.36 &           0.34 $\pm$ 0.34 &  \textbf{0.73} $\pm$ 0.37 &           0.51 $\pm$ 0.36 \\
\textbf{RESCU-I  } &   0.47 $\pm$ \textbf{0.30} & 0.51 $\pm$ \textbf{0.35} &  0.70 $\pm$ 0.31 &  \textbf{ 0.65} $\pm$ \textbf{0.32} \\
\textbf{RESCU-II } &  \textbf{0.48} $\pm$ 0.31 &  \textbf{0.52} $\pm$ \textbf{0.35} &           \textbf{0.73} $\pm$ \textbf{0.30} & \textbf{0.65} $\pm$ \textbf{0.32} \\

\hline
\midrule
\multicolumn{5}{c}{\textbf{Projective View Subset}}\\
\hline
\textbf{Baseline\cite{liu2013grasp}} &           0.23 $\pm$ 0.36 &           0.22 $\pm$ \textbf{0.37} &           0.36 $\pm$ 0.45 &           0.27 $\pm$ 0.38 \\
\textbf{RESCU-I  } &           0.32 $\pm$ \textbf{0.34} &           \textbf{0.47} $\pm$ 0.45 &           0.58 $\pm$ 0.40 &           0.44 $\pm$ 0.37 \\
\textbf{RESCU-II } &  \textbf{0.33} $\pm$ 0.35 &  \textbf{0.47} $\pm$ 0.45 &  \textbf{0.63 $\pm$ 0.38} &  \textbf{0.49 $\pm$ 0.37} \\

\hline
\midrule
\multicolumn{5}{c}{\textbf{Man-Made Rigid Subset}}\\
\hline
\textbf{Baseline\cite{liu2013grasp}} &           0.42 $\pm$ 0.37 &           0.32 $\pm$ \textbf{0.35} &           0.69 $\pm$ 0.40 &           0.49 $\pm$ 0.37 \\
\textbf{RESCU-I  } &           0.44 $\pm$ \textbf{0.32} &          \textbf{ 0.50} $\pm$ 0.37 &           0.69 $\pm$ 0.33 &           0.62 $\pm$ \textbf{0.34} \\
\textbf{RESCU-II } &  \textbf{0.45} $\pm$ \textbf{0.32} &  \textbf{0.51} $\pm$ 0.37 &  \textbf{0.72} $\pm$ \textbf{0.32} &  \textbf{0.63} $\pm$ \textbf{0.34} \\

\hline
\midrule
\multicolumn{5}{c}{\textbf{Man-Made Deformable Subset}}\\
\hline
\textbf{Baseline\cite{liu2013grasp}} &           0.33 $\pm$ 0.35 &           0.21 $\pm$ \textbf{0.26} &           0.58 $\pm$ 0.43 &           0.43 $\pm$ 0.37 \\
\textbf{RESCU-I  } &           0.38 $\pm$ \textbf{0.28} &           0.41 $\pm$ 0.33 &           0.63 $\pm$ \textbf{0.31} &           \textbf{0.63} $\pm$ \textbf{0.30} \\
\textbf{RESCU-II } &  \textbf{0.40} $\pm$ 0.29 &  \textbf{0.44} $\pm$ 0.33 &  \textbf{0.66} $\pm$ \textbf{0.31} &  \textbf{0.63} $\pm$ \textbf{0.30} \\

\hline
\midrule
\multicolumn{5}{c}{\textbf{Painting Subset}}\\
\hline
\textbf{Baseline\cite{liu2013grasp}} &           0.45 $\pm$ 0.40 &           0.61 $\pm$ 0.48 &           0.64 $\pm$ 0.46 &           0.34 $\pm$ 0.34 \\
\textbf{RESCU-I  } &           0.73 $\pm$ \textbf{0.30} &  \textit{\textbf{0.89} $\pm$ \textbf{0.29}} &           \textbf{0.82} $\pm$ \textbf{0.24} &  \textbf{0.60} $\pm$ \textbf{0.31} \\
\textbf{RESCU-II } &  \textbf{0.74} $\pm$ \textbf{0.30} &           \textit{\textbf{0.89} $\pm$ \textbf{0.29}} &  \textbf{0.82} $\pm$ 0.25 &           \textbf{0.60} $\pm$ \textbf{0.31} \\

\hline
\midrule
\multicolumn{5}{c}{\textbf{Animal/Human Subset}}\\
\hline
\textbf{Baseline\cite{liu2013grasp}} &           0.45 $\pm$ 0.47 &           0.52 $\pm$ 0.50 &           0.39 $\pm$ 0.48 &           0.25 $\pm$ 0.35 \\
\textbf{RESCU-I  } &           0.57 $\pm$ \textbf{0.37} &           0.78 $\pm$ 0.42 &           0.76 $\pm$ 0.36 &           0.59 $\pm$ 0.35 \\
\textbf{RESCU-II } &  \textbf{0.62} $\pm$ \textbf{0.37} &  \textbf{0.81} $\pm$ \textbf{0.39} &  \textbf{0.83} $\pm$ \textbf{0.30} &  \textbf{0.67} $\pm$ \textbf{0.32} \\

\bottomrule
\end{tabular}

\end{table}

\subsubsection{Ablation Study}
\label{sec:ablation_study}
As mentioned in Sec.~\ref{sec:UPRD-I}, here we study the impact of each adaptive parameter.
\begin{itemize}
    \item $P_d$ sets the maximum feature distance, which controls the granularity of RP detection.
    \item $P_s$ sets the maximum size difference among RP instances.
    \item $P_{\theta}$ sets the maximum orientation difference among RP instances.
\end{itemize}

To study the impact of these adaptive parameters, we design experiments to separately optimize each parameter (and each pair of two parameters), with fixed other parameters. Tab.~\ref{tab:para_settings} shows the fixed value and optimization values of each parameter.

\begin{table}[tb!]

\caption{\small  A Summary of Adaptive Parameters used in RESCU Stage-I.}
\label{tab:para_settings}
\centering

\resizebox{\linewidth}{!}
{
\begin{tabular}{l|c|c|c} 
\toprule
Adaptive Parameter & Description &  Fixed Value & Optimized Values\\
\midrule

$P_d$ & the maximum feature distance & 0.2 & $[0.1, 0.15, 0.2]$\\
$P_s$ & the maximum size difference among RP instances & 0.5 & $[0.1, 0.2, 0.3, 0.4, 0.5]$\\
$P_\theta$ & the maximum orientation difference among RP instances & 30 & $[30, 90, 180]$\\
\bottomrule
\end{tabular}
}
\end{table}

\begin{figure}[]
    \centering
    \includegraphics[width=0.8\linewidth]{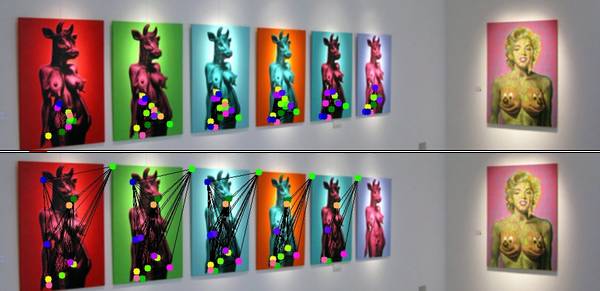}
    \caption{\small An example of using different $P_d$ in RESCU. 
    \textbf{Up}: with $P_d=0.05$ to detect smaller RP instance, \textbf{Down}: with $P_d=0.25$ to detect larger RP instance.}
    \label{fig:p_d_ex}
\end{figure}

Tab.~\ref{tab:stage_1_ablation} shows the ablation study of adaptive parameters. From the study we can see that $P_d$ parameter influences the performance most.
Fig.~\ref{fig:p_d_ex} shows an example of how changing $P_d$ can control the size of an RP-instance.

\begin{table}[b]
\caption{\small  Ablation Study of RESCU-I IOD Threshold $h=0.5$.
$P_d$ (maximum RPI feature distance),
$P_s$ (maximum RPI size difference),
$P_{\theta}$ (maximum RPI orientation difference),
See Tab.~\ref{tab:para_settings} for parameter details.
The values in \textit{italic} are not statistically significant with range in the same column.
The values in \textbf{bold} are the best mean/std with range in the same column.
}
\label{tab:stage_1_ablation}
\centering

\begin{tabularx}{\textwidth}{c|XXX|cc|cc} 
\toprule
 &  \multicolumn{3}{c|}{Parameter} & \multicolumn{2}{c|}{RP Level} & \multicolumn{2}{c|}{RP Instance Level} \\
Method &  $P_d$ & $P_s$ & $P_\theta$ &   Precision &        Recall &   Precision &      Recall \\
\midrule

\textbf{Baseline\cite{liu2013grasp}} &  & & &         0.40 $\pm$ 0.37 &           0.32 $\pm$ \textbf{0.35} &           0.67 $\pm$ 0.41 &           0.47 $\pm$ 0.38 \\

\textbf{RESCU-I} &  & & &  0.37 $\pm$ \textbf{0.31} &  0.44 $\pm$ 0.38 &  0.70 $\pm$ 0.30 &  0.64 $\pm$ \textbf{0.32} \\

\textbf{RESCU-I} &  \checkmark & & & \textit{\textbf{0.48} $\pm$ 0.33} &  \textit{\textbf{0.53} $\pm$ 0.38} &  0.73 $\pm$ 0.29 & \textit{\textbf{0.67} $\pm$ \textbf{0.32}} \\
\textbf{RESCU-I} & & \checkmark & &  0.43 $\pm$ 0.33 &  0.46 $\pm$ 0.38 &  0.72 $\pm$ 0.30 &  0.65 $\pm$ \textbf{0.32} \\
\textbf{RESCU-I} & & & \checkmark &  0.41 $\pm$ 0.33 &  0.44 $\pm$ 0.38 &  0.67 $\pm$ 0.30 &  0.63 $\pm$ \textbf{0.32} \\

\textbf{RESCU-I} & \checkmark & \checkmark &  & \textit{0.47 $\pm$ 0.33} &  \textit{\textbf{0.53} $\pm$ 0.38} & \textbf{ 0.75} $\pm$ \textbf{0.28} &  \textit{\textbf{0.67} $\pm$ \textbf{0.32}} \\
\textbf{RESCU-I} & \checkmark &  & \checkmark &  \textit{\textbf{0.48} $\pm$ 0.33} &  \textit{\textbf{0.53} $\pm$ 0.38} &  0.72 $\pm$ 0.29 & \textit{ 0.66 $\pm$ \textbf{0.32}} \\
\textbf{RESCU-I} &  & \checkmark & \checkmark & 0.43 $\pm$ 0.33 &  0.45 $\pm$ 0.38 &  0.69 $\pm$ 0.30 &  0.64 $\pm$ \textbf{0.32} \\

\textbf{RESCU-I} & \checkmark & \checkmark & \checkmark &  \textit{0.47 $\pm$ 0.34} &  \textit{0.52 $\pm$ 0.38} &  0.72 $\pm$ 0.29 &  \textit{0.66 $\pm$ \textbf{0.32} }\\
\bottomrule
\end{tabularx}
\end{table}

\section{RPD Scene Understanding}

\begin{figure}[] \centering
\begin{subfigure}[t]{0.24\linewidth}\centering
    \includegraphics[height = 0.85\textwidth, width =1\textwidth]{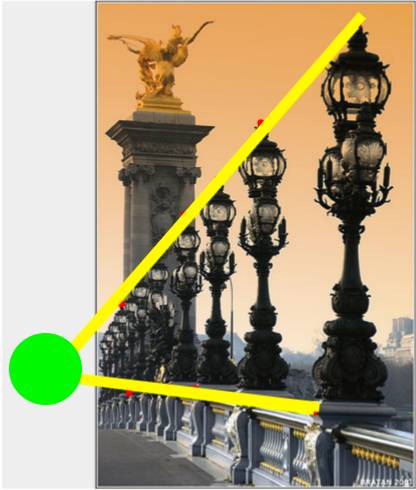}
    \includegraphics[height = 0.85\textwidth, width =1\textwidth]{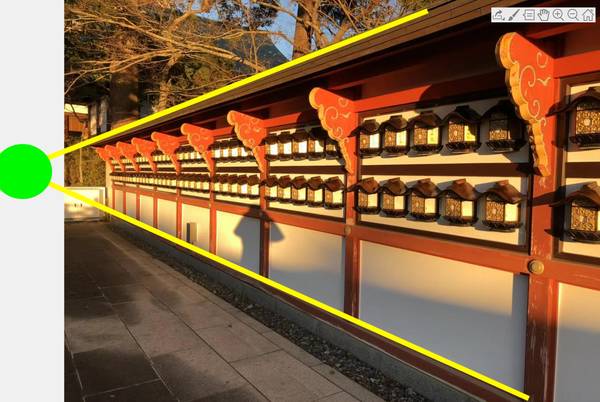}
    \vspace{-15pt}
    \caption{\small GT}
\end{subfigure}
\begin{subfigure}[t]{0.24\linewidth}\centering
    \includegraphics[height = 0.85\textwidth, width =1\textwidth]{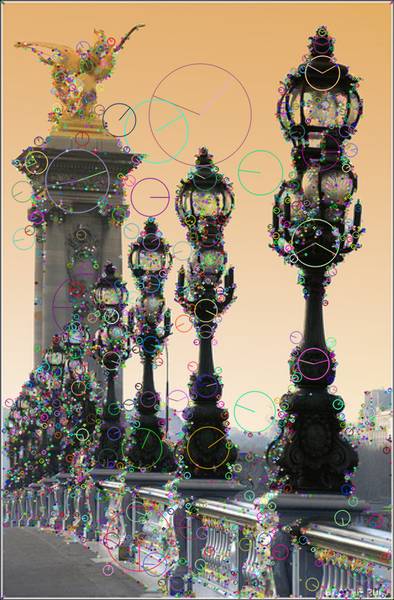}
    \includegraphics[height = 0.85\textwidth, width =1\textwidth]{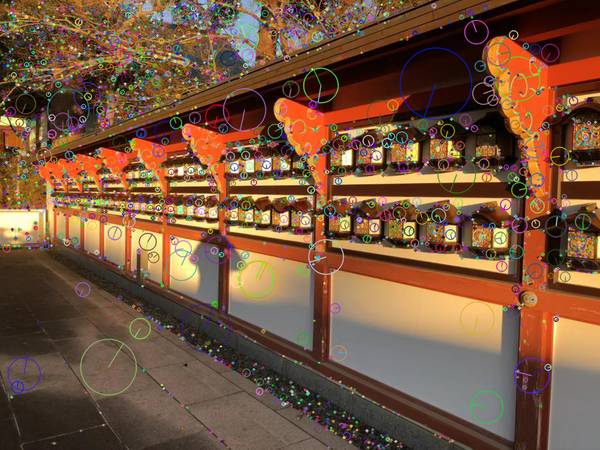}
    \vspace{-15pt}
    \caption{\small SIFT}
\end{subfigure}
\begin{subfigure}[t]{0.24\linewidth}\centering
    \includegraphics[height = 0.85\textwidth, width =1\textwidth]{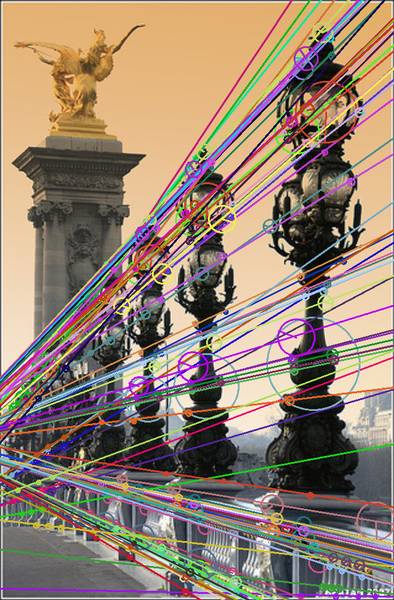}
    \includegraphics[height = 0.85\textwidth, width =1\textwidth]{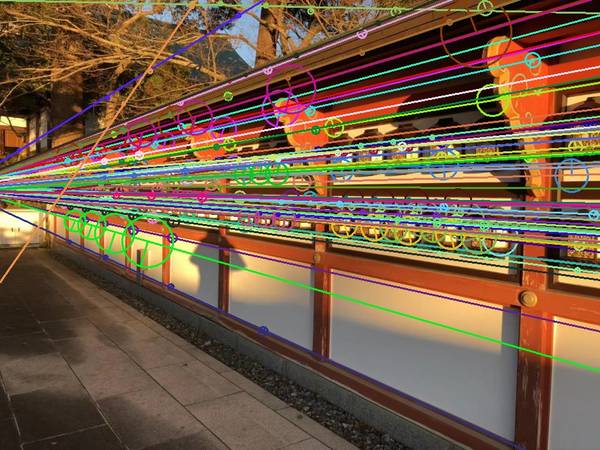}
    \vspace{-15pt}
    \caption{\small  Line Fit}
\end{subfigure}
\begin{subfigure}[t]{0.24\linewidth}\centering
    \includegraphics[height = 0.85\textwidth, width =1\textwidth]{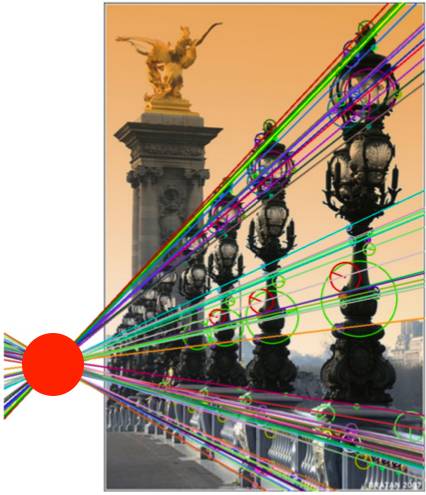}
    \includegraphics[height = 0.85\textwidth, width =1\textwidth]{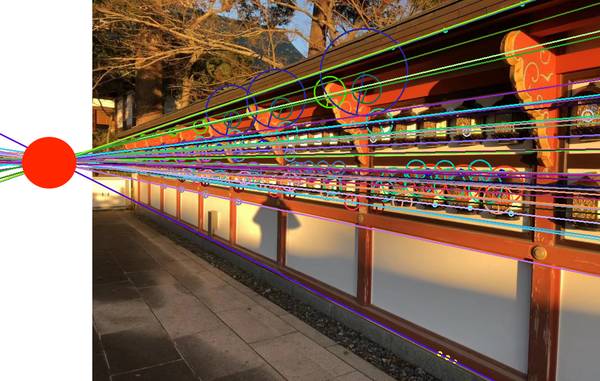}
    \vspace{-15pt}
    \caption{\small  Detected}
\end{subfigure}
\vspace{-10pt}
\caption{\small  Sample results for VPD. \textbf{(a)} Input image with ground truth, 
\textbf{(b)} SIFT features extracted, 
\textbf{(c)} lines fitted to SIFT feature groups, 
\textbf{(d)} vanishing point detected from our method. More results in supplementary Sec~2.3.}
\vspace{-5pt}
\label{fig:vp_ex}
\end{figure}
\subsection{Vanishing Point Detection from A Single View}
\label{sec:VP}
``Under perspective projection, parallel lines in 3D do not remain parallel but instead meet at the point called {\it vanishing point} (VP)'' \cite{mundy1992projective}. We also emphasize that line segments are not always explicitly visible in real-world images (Fig.~\ref{fig:1-d}). We take advantage of groups of corresponding feature points on detected recurring patterns (Figure \ref{fig:vp_ex}) to construct and validate the co-linearity of implicit lines and line intersections to find VP through a robust RANSAC procedure.

\subsubsection{Line Fitting and RANSAC in Vanishing Point Detection}
\begin{figure*}[]
    \centering
    \includegraphics[width=0.65\linewidth, height=0.3\textwidth]{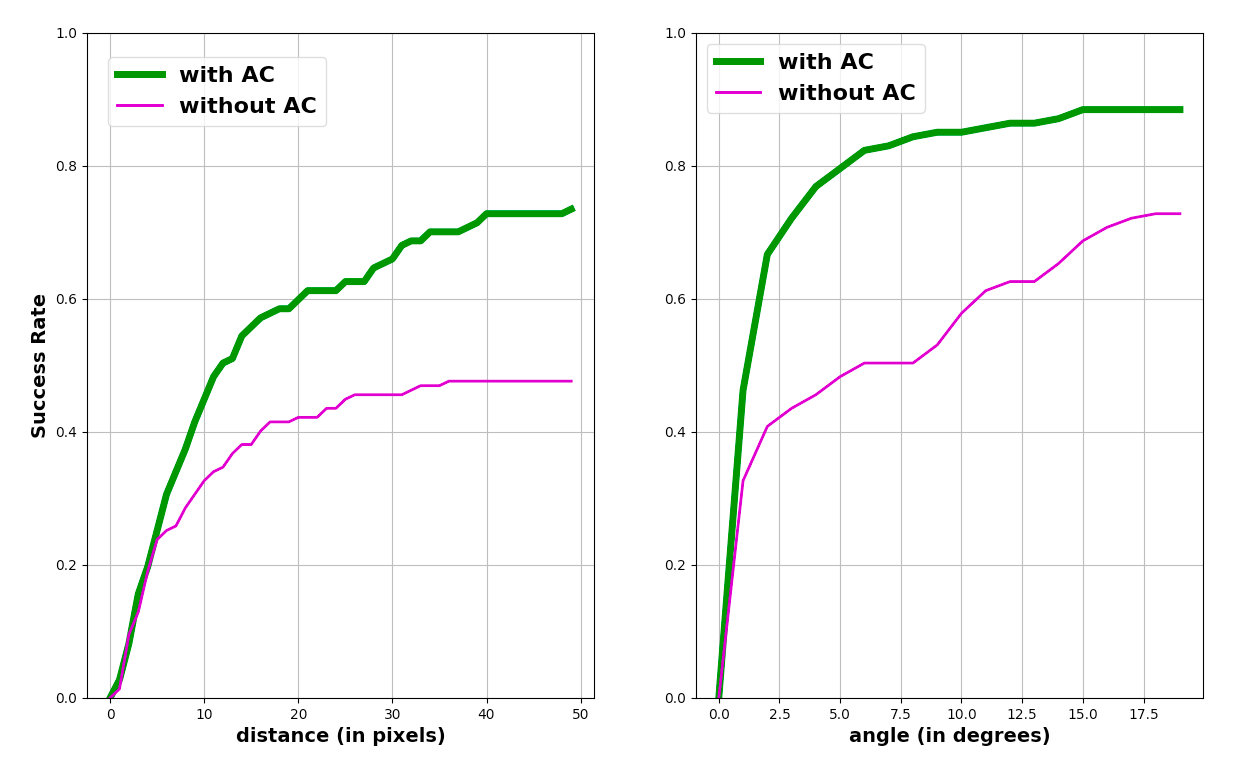}
    \caption{\small Success rate of vanishing point detection (VPD) with respect to distance (left) and angle (right) difference between predicted and groundtruth VPs. With the introduction of the angular constraint (AC) we see a significant improvement in the performance. The performance of the proposed approach is statistically different with {\it p-value} = 1.13e-06 with the angular constraint.}
    \label{fig:angular_constraint}
\end{figure*}

The intersection of near-parallel lines often results in faulty estimation of the vanishing point. To overcome this ill-conditioning, we introduce an angular constraint (AC) into our algorithm by comparing the angle between lines before selecting them to initialize the RANSAC algorithm. If the angle is smaller (near-parallel lines) than a threshold, we do not consider their contribution towards vanishing point calculation. Using this angular constraint produces a better estimate of the vanishing point. Fig.~\ref{fig:angular_constraint} shows the success rate of our method with and without the Angular Constraint.

\vspace{-0.3cm}
\subsubsection{Vanishing Point Detection Evaluation}

\begin{figure*}[!ht]
    \centering
    \includegraphics[width=0.65\linewidth, height=0.3\textwidth]{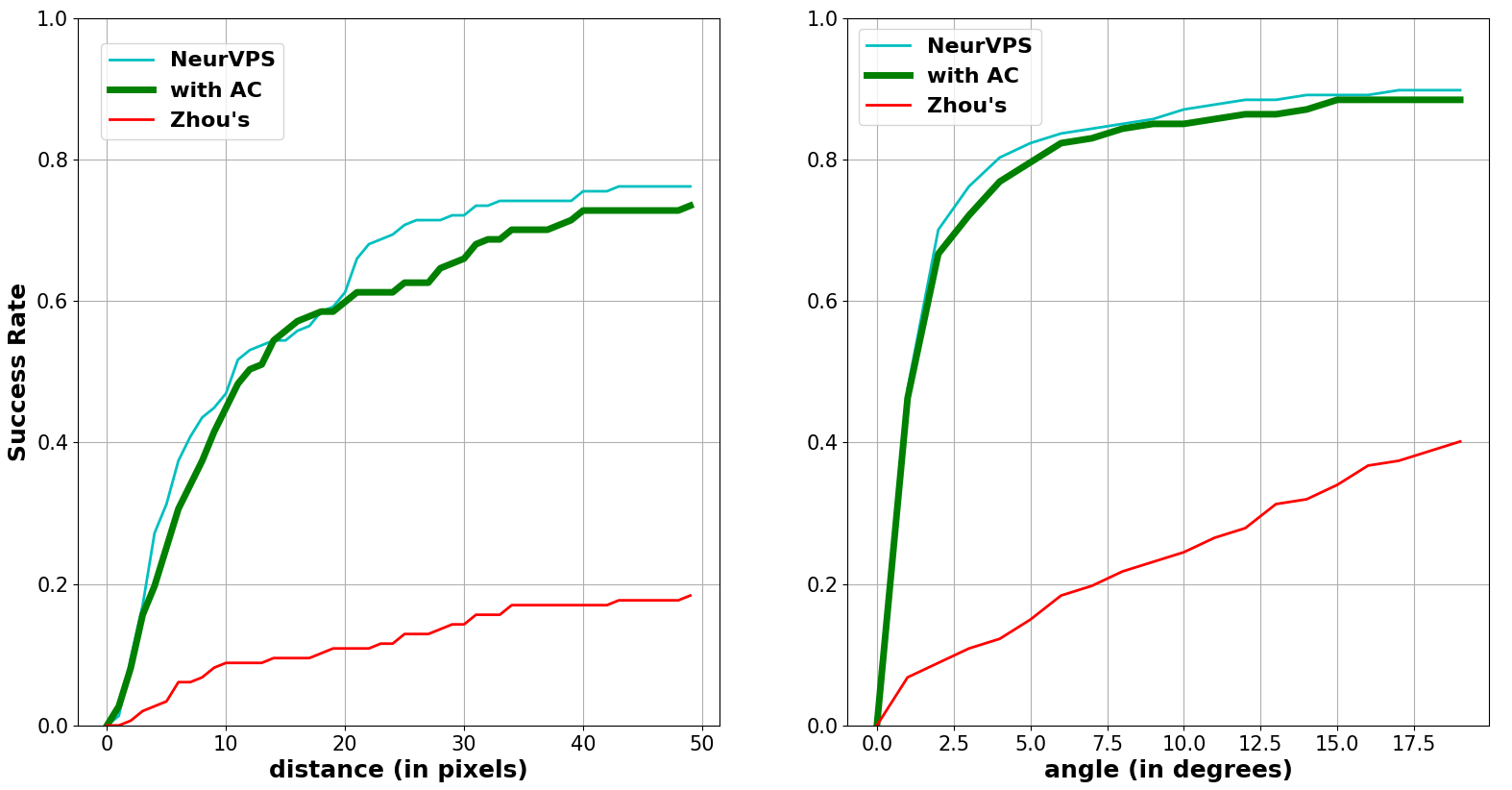}
    \caption{\small Qualitative analysis of VPD evaluated using distance (left) and angle (right) difference to groundtruth (GTs) respectively. On a test image set of $147$ images, we compare our unsupervised recurring pattern based VPD method (Sec.~\ref{sec:VP}) with a state-of-the-art deep learning vanishing point detector \textbf{NeurVPS} \cite{zhou2019neurvps}, a supervised learning method trained on $270,000$ training images. Our statistical analysis of the results show that these two methods have no statistically difference ({\em p-values} are 0.349 for distance and 0.723 for angle difference).}
    \label{fig:vp_sr}
\end{figure*}

~We compare our VP detection method with \cite{zhou2017detecting}
and a deep learning method for VPD, NeurVPS \cite{zhou2019neurvps}, using public available projective view subset of 147 images with labeled VP ground truth  (Fig.~\ref{fig:vp_sr}). 

The success rate (SR) of vanishing point detection is computed as the ratio of number of acceptable vanishing points against ground truth over all the detected vanishing points. We define two methods for comparing a detected VP and the ground truth: a point-based method and a vector-based method. 
The point-based method compares Euclidean distance in pixels between the 2D VP location $(\mbox{vp}_x, \mbox{vp}_y)$ and the ground truth labeled pixel location, with detections considered accepted if this distance falls within a threshold. In the vector-based method, the VP is represented as a unit vector in the direction $(\mbox{vp}_x - x_0, \mbox{vp}_y - y_0, f)$, where $(x_0,y_0)$ is the image center, and $f$ is a nominal value that would be the camera focal length if known, but otherwise is chosen to be (image width + image height)/4. Distance in this case is the angle between the detected and ground truth VP unit vectors.

Fig.~\ref{fig:vp_sr} plots the success rate of our method for Euclidean distance (left) and for vector angle (right). We compare our results with NeurVPS \cite{zhou2019neurvps} and Zhou's method \cite{zhou2017detecting} on a set of $147$ images. The baseline method \cite{zhou2017detecting} performs poorly on this dataset because it relies on explicit straight line segments in the image to estimate vanishing points.
See supplementary Fig.10~12 for more examples with few visually explicit line segments.
Our method performs nearly as well as the supervised SOTA method NeurVPS\cite{zhou2019neurvps}, as shown quantitatively in Fig.~\ref{fig:vp_sr}. 
Our statistical analysis of the results shows that there is no statistically difference on VPD between our method and NeurVPS with {\it{p-values}} of $0.349$ for distance and $0.723$ for angle difference.

\begin{figure*}\vspace{-1pt}
\centering
\begin{subfigure}[t]{0.2375\linewidth}

\caption{\small \tiny{$A^{\circ}<1^{\circ}$ \& $D<10p$}} 
    \includegraphics[width = \linewidth, height=0.75\linewidth]{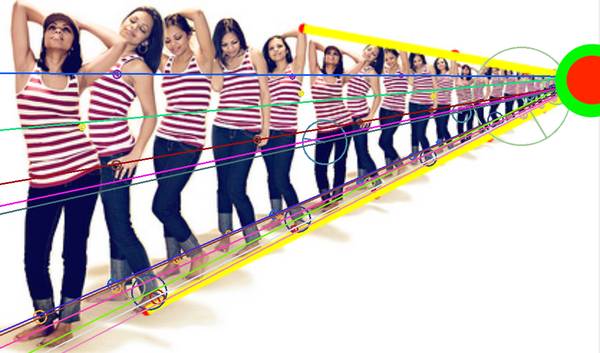}
    \includegraphics[width = \linewidth, height=0.75\linewidth]{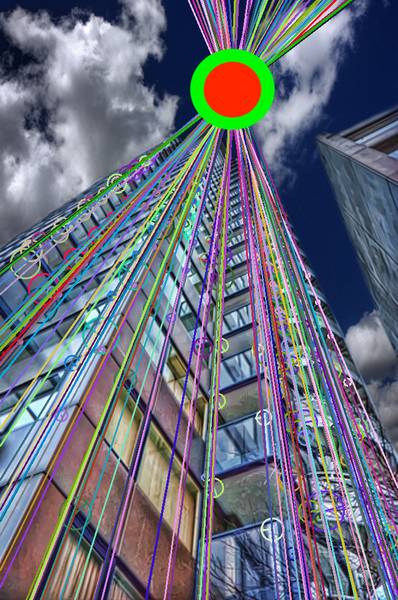}
    \includegraphics[width = \linewidth, height=0.75\linewidth]{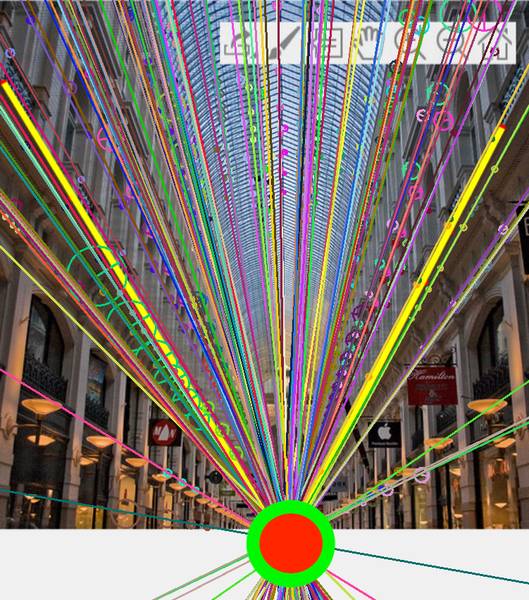}
\end{subfigure}\hfill
\begin{subfigure}[t]{0.2375\linewidth}
    \caption{\small \tiny{$A^{\circ}<5^{\circ}$ \& $D < 20p$}}
    \includegraphics[width = \linewidth, height=0.75\linewidth]{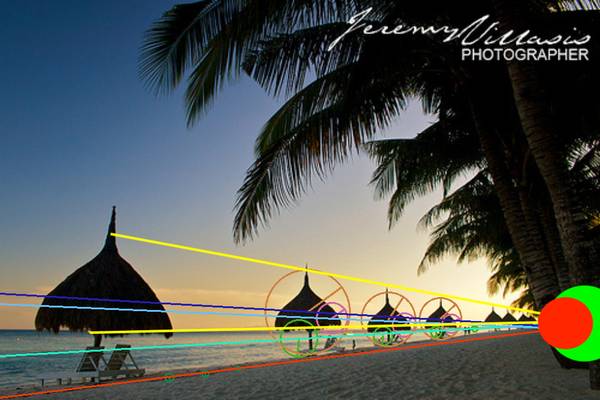}
    \includegraphics[width = \linewidth, height=0.75\linewidth]{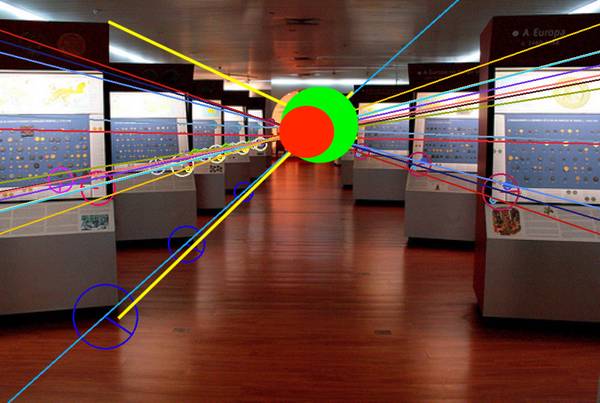}
    \includegraphics[width = \linewidth, height=0.75\linewidth]{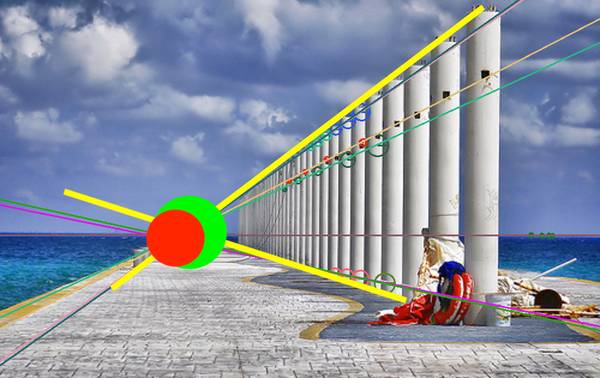}
\end{subfigure}\hfill
\begin{subfigure}[t]{0.2375\linewidth}
    \caption{\small \tiny{$A^{\circ}>5^{\circ}$ or $D<20p$}}
    \includegraphics[width = \linewidth, height=0.75\linewidth]{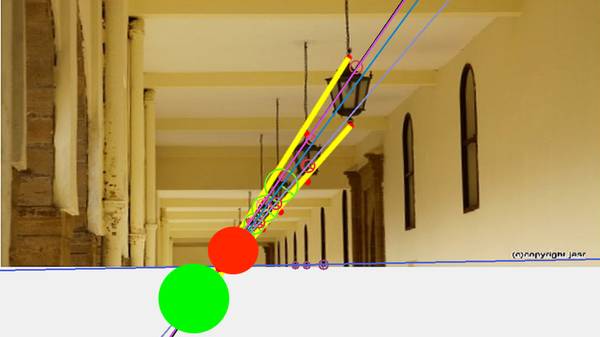}
    \includegraphics[width = \linewidth, height=0.75\linewidth]{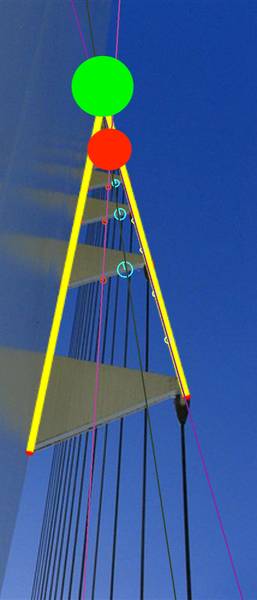}
    \includegraphics[width = \linewidth, height=0.75\linewidth]{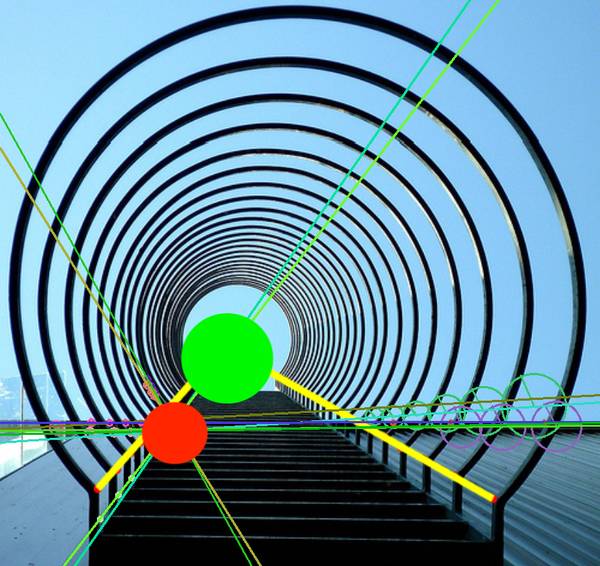}
\end{subfigure}\hfill 
\begin{subfigure}[t]{0.2375\linewidth}
    \caption{\small \tiny{$A^{\circ}>20^{\circ}$ or $D > 30p$}}
    \includegraphics[width = \linewidth, height=0.75\linewidth]{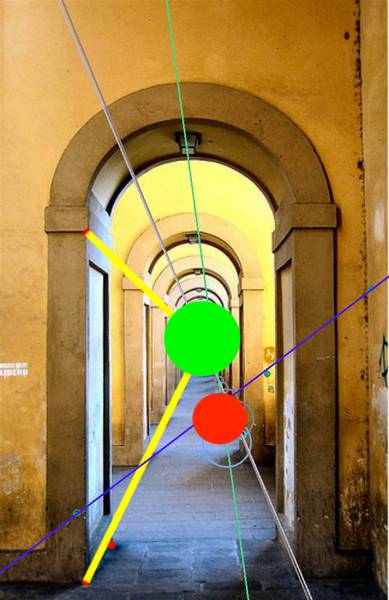}
    \includegraphics[height = \textwidth, width = \linewidth, height=0.75\linewidth]{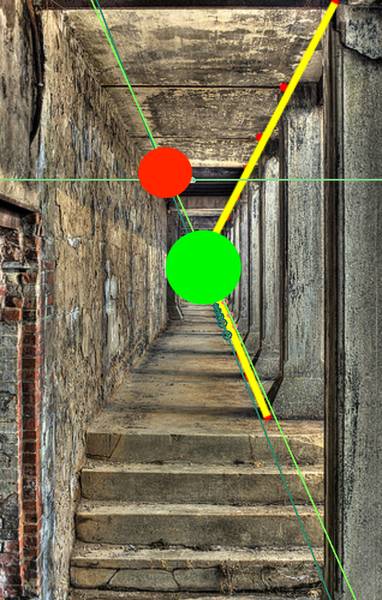}
    \includegraphics[width = \linewidth, height=0.75\linewidth]{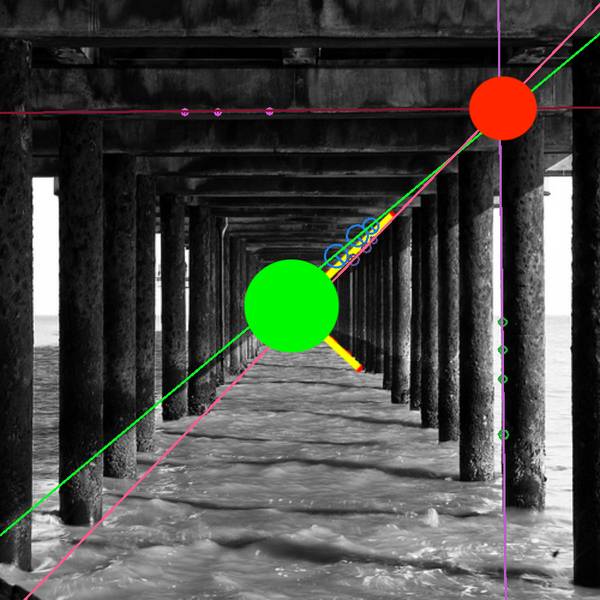}
\end{subfigure}
\caption{\small Example vanishing point prediction by our method. $A^{\circ}$ denotes the vector angle between predicted and ground truth VP vectors in degrees. $D$ represents the distance between predicted and ground truth VP image locations in pixels ($p$). \textcolor{red}{Red} dot: our prediction. \colorbox{yellow}{Yellow} lines and \textcolor{green}{Green} dot: two supporting lines intersected at the ground truth vanishing point.  
}
\label{fig:vp_best_worst}
\end{figure*}

In addition, we provide the complete results for Vanishing Point (VP) prediction. Fig. \ref{fig:vp_best_worst} shows prediction results from our method separated in columns based on the prediction quality compared with the ground truth. We use angle and distance with respect to the ground truth to assess the prediction quality.

\begin{figure*}
\centering
\begin{subfigure}[t]{0.198\linewidth}
    \caption{\small GT}
    \includegraphics[height = 0.80\textwidth, width = \textwidth]{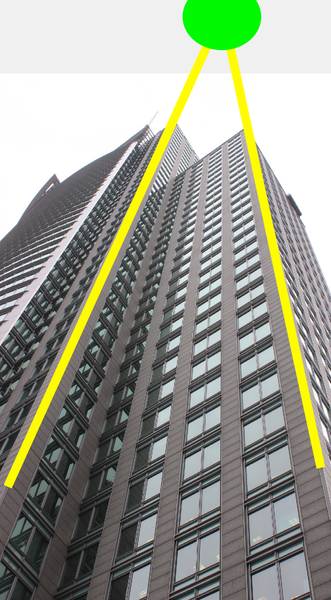}
    \includegraphics[height = 0.80\textwidth, width = \textwidth]{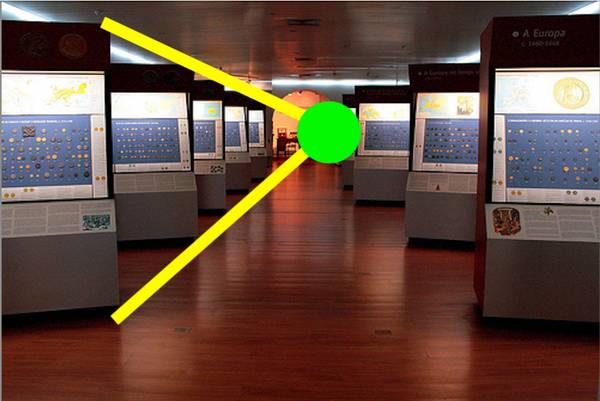}
    \includegraphics[height = 0.80\textwidth, width = \textwidth]{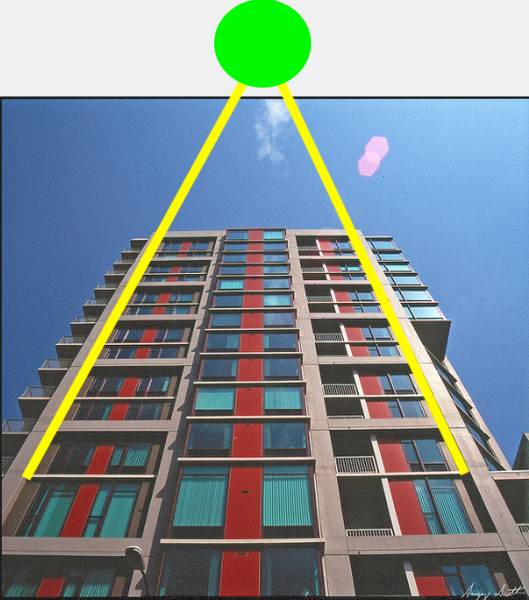}
\end{subfigure}\hfill
\begin{subfigure}[t]{0.198\linewidth}
    \caption{\small Our's}
    \includegraphics[height = 0.80\textwidth, width = \textwidth]{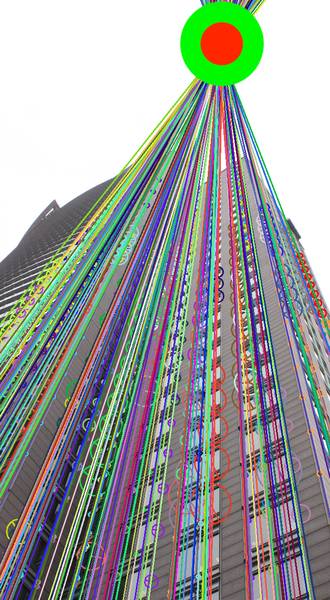}
    \includegraphics[height = 0.80\textwidth, width = \textwidth]{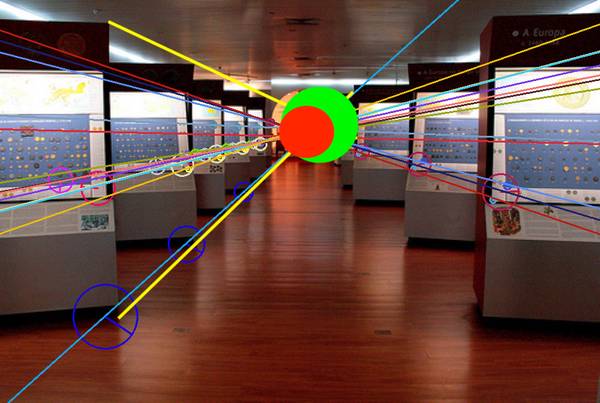}
    \includegraphics[height = 0.80\textwidth, width = \textwidth]{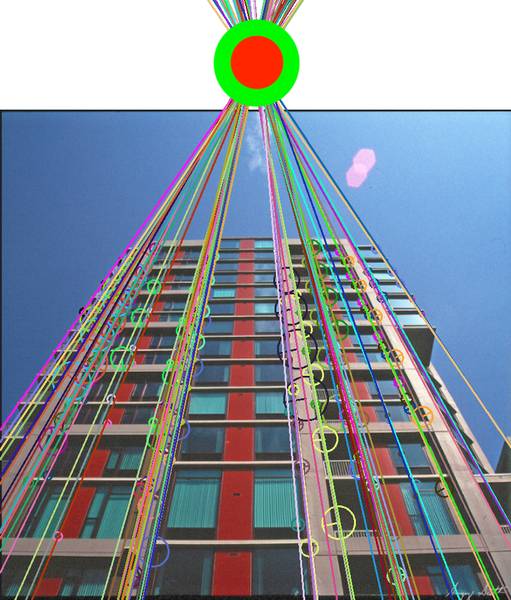}
\end{subfigure}\hfill
\begin{subfigure}[t]{0.198\linewidth}
    \caption{\small NeurVPS\label{fig:zhou_vp}}
    \includegraphics[height = 0.80\textwidth, width = \textwidth]{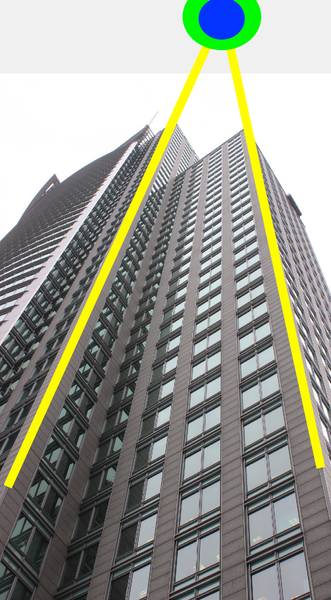}
    \includegraphics[height = 0.80\textwidth, width = \textwidth]{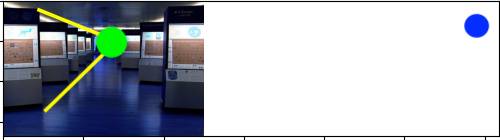}
    \includegraphics[height = 0.80\textwidth, width = \textwidth]{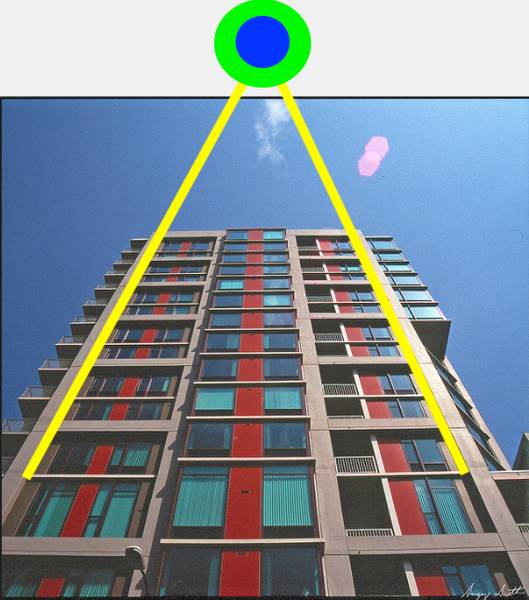}
\end{subfigure}\hfill
\begin{subfigure}[t]{0.198\linewidth}
    \caption{\small Zhou's\label{fig:zhou_vp_cont2}}
    \includegraphics[height = 0.80\textwidth, width = \textwidth]{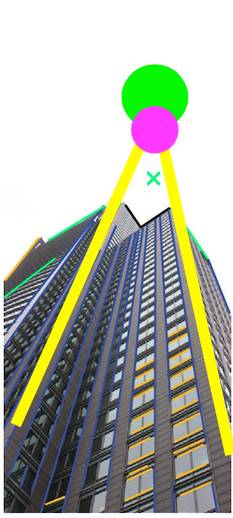}
    \includegraphics[height = 0.80\textwidth, width = \textwidth]{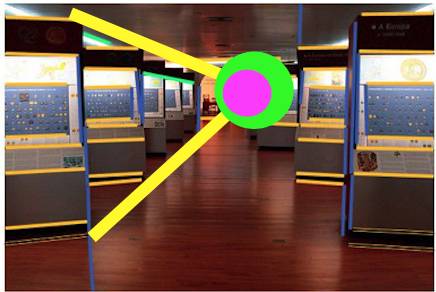}
    \includegraphics[height = 0.80\textwidth, width = \textwidth]{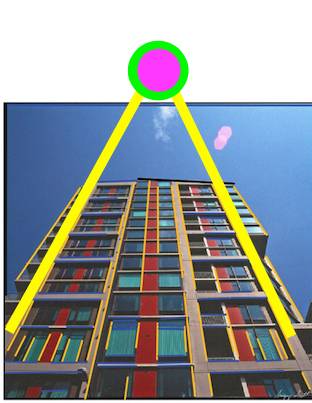}
\end{subfigure}\hfill
\caption{\small 
Sample results for vanishing point (VP) detection using our method and a comparison with \cite{zhou2019neurvps} and \cite{zhou2017detecting}. a) Input image with ground truth, 
b) VP prediction using our method, 
c) VP prediction using NeurVPS\cite{zhou2019neurvps}, 
d) VP prediction using Zhou et.al.'s method \cite{zhou2017detecting}. \textcolor{red}{Red} dot: our prediction. \colorbox{yellow}{Yellow} lines and \textcolor{green}{Green} dot: two supporting lines intersected at the ground truth vanishing point. \textcolor{blue}{Blue} is the output of NeurVPS\cite{zhou2019neurvps} and \textcolor{magenta}{Magenta} is the output of Zhou et.al.'s method\cite{zhou2017detecting}.
}
\label{fig:vp_ex_1}
\end{figure*}

We evaluate our unsupervised recurring pattern based VPD method on a projective view subset of 147 images with labeled VP ground truth and compare it with  a state-of-the-art supervised deep learning vanishing point detector \textbf{NeurVPS} \cite{zhou2019neurvps} and another baseline method \cite{zhou2017detecting}. Fig.~\ref{fig:vp_ex_1} shows the output of the three methods along with the ground truth. 

\subsection{Translation Symmetry Detection}
\label{sec:translation_sym}

\begin{figure}[!h]
    \centering
    \begin{subfigure}{0.3\linewidth}
    \includegraphics[width = \textwidth, height=0.65\textwidth]{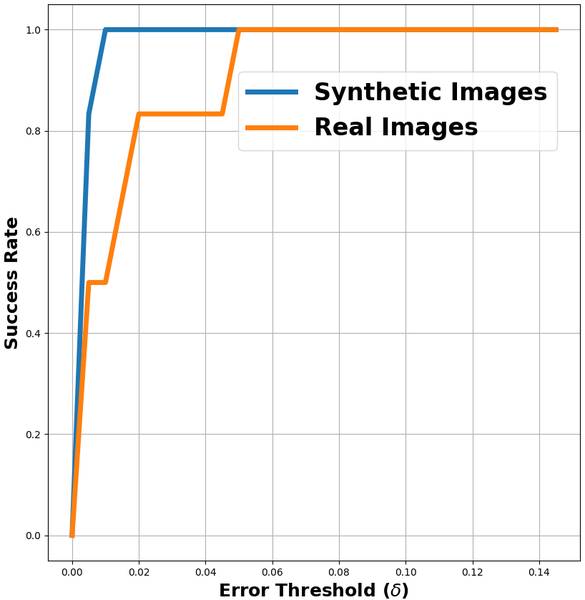}
    \caption{\small TS\_GT Dataset (12 images)}
    \end{subfigure} \hfill
    \begin{subfigure}{0.3\linewidth}
    \includegraphics[width = \textwidth, height=0.65\textwidth]{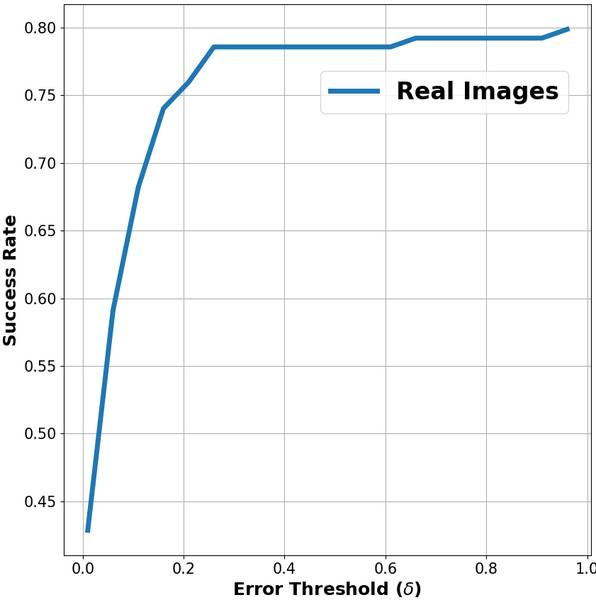}
    \caption{\small VPD Dataset (147 images)}
    \end{subfigure} \hfill
    \begin{subfigure}{0.30\linewidth}
    \includegraphics[width = \textwidth, height=0.65\textwidth]{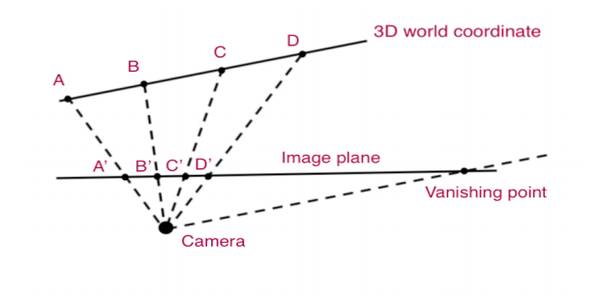}
    \caption{\small Cross-ratio Model}
    \end{subfigure} \hfill
    \caption{\small Success rate of our translation symmetry detection for varying thresholds. (a) represents the success rate on a translation symmetry ground truth (TS\_GT) dataset consisting of 6 synthetic images and 6 real images for which the translation symmetry is known to exist (example images are shown in Fig.~\ref{fig:ts_examples}). (b) represents the success rate on the VPD dataset with 147 images. (c) represents the cross-ratio model that depicts the relation between the colinear points in the 3D space and their corresponding projections in the 2D image.}
    \label{fig:ts_roc}
\end{figure}

To further understand and quantify a 3D scene, we investigate the possibility of an RP having 3D translation symmetry. Using projective invariance we are able to detect potential 3D translation symmetry of RP instances in a 2D image. Consider four points $A, B, C$ and $D$ having translation symmetry in 3D space their corresponding projections $A', B', C'$ and $D'$ on the 2D image as shown in Fig.~\ref{fig:ts_roc}\textcolor{red}{c}. By virtue of translation symmetry, $AB=BC=CD= d$ for some 3D translation vector length $||t||=d$.
The cross ratio, defined as \cite{mundy1992projective}:
\begin{equation} \label{eq:cross_ratio}
    cr (A, B, C, D) = \frac{AC*BD}{BC*AD}=
    cr (A', B', C', D')
\end{equation}
is an invariant value shared by four co-linear 3D points and their 2D image projections regardless of camera position and orientation. For the case of 3D translation symmetry, this computed invariant value will be
$$  \frac{AC*BD}{BC*AD}  = 
\frac{(d+d)(d+d)}{d(d+d+d)} = \frac{4}{3}. $$

\begin{figure}[!h]
\centering
    \begin{tabular}{cccc}
    \includegraphics[width = 0.25\textwidth, valign=t]{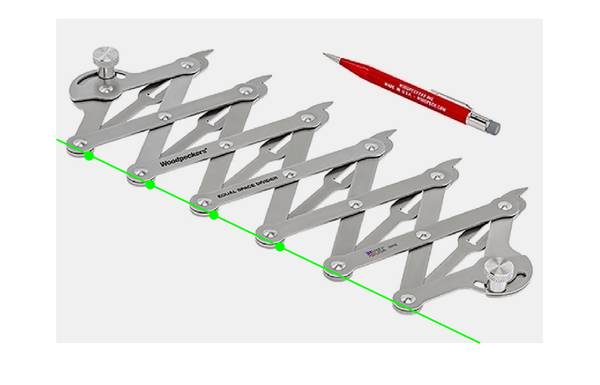} &
    \includegraphics[height = 0.15\textwidth, valign=t]{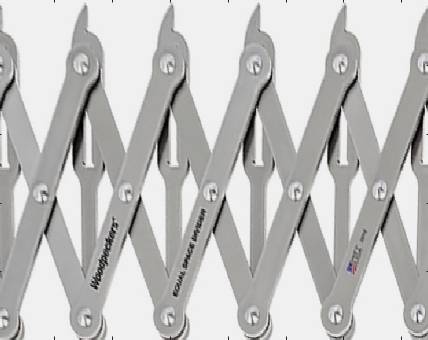} &
    \includegraphics[height = 0.2\textwidth, valign=t]{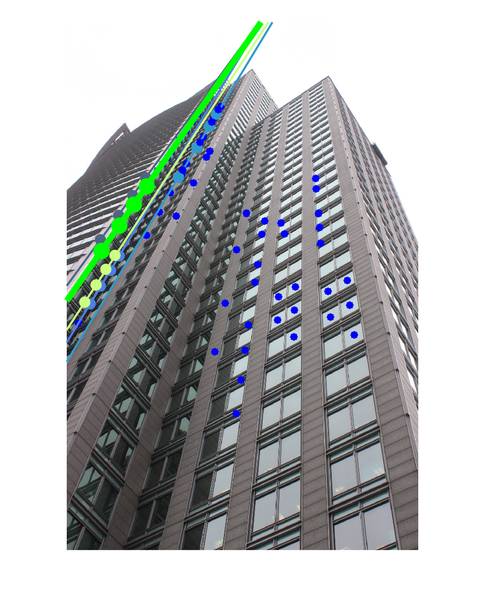} &
    \includegraphics[height = 0.15\textwidth, valign=t]{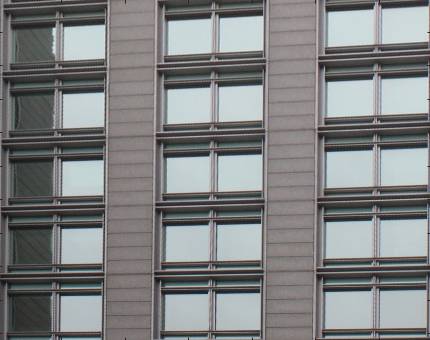}  \\
    \hline\\
    \includegraphics[width = 0.25\textwidth, valign=t]{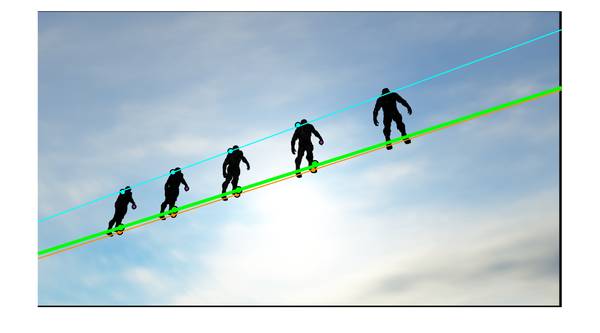} &
    \includegraphics[width = 0.25\textwidth, height = 0.1\textwidth, valign=t]{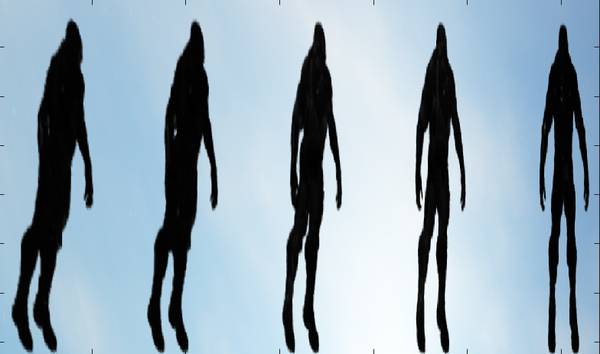} &
    \includegraphics[width = 0.25\textwidth, valign=t]{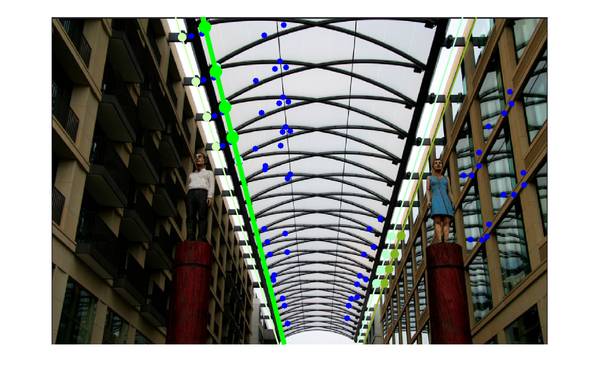} &
    \includegraphics[height = 0.15\textwidth, valign=t]{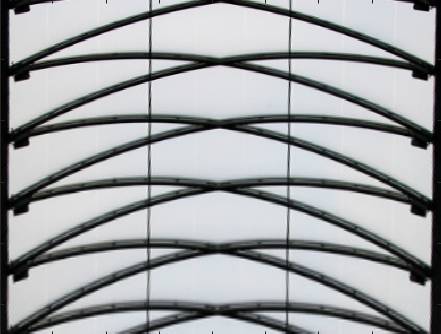}  \\
    \hline\\
    \includegraphics[width = 0.25\textwidth, valign=t]{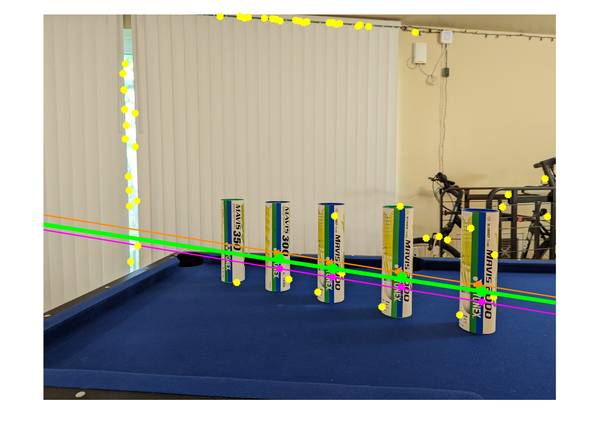} &
    \includegraphics[width = 0.25\textwidth, valign=t]{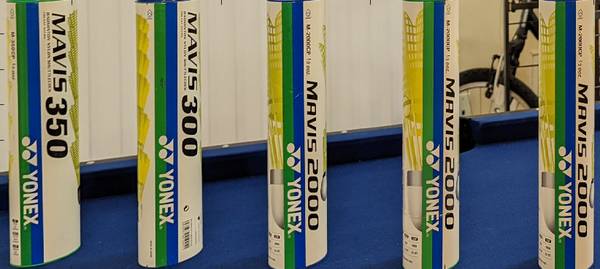} &
    \includegraphics[width = 0.25\textwidth, valign=t]{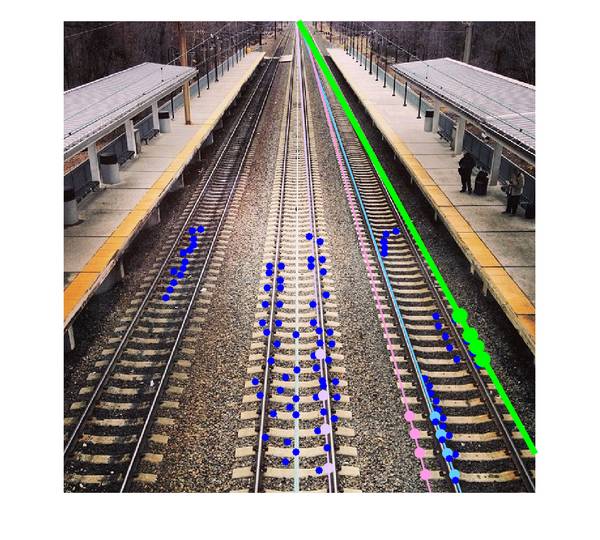} &
    \includegraphics[height = 0.15\textwidth, valign=t]{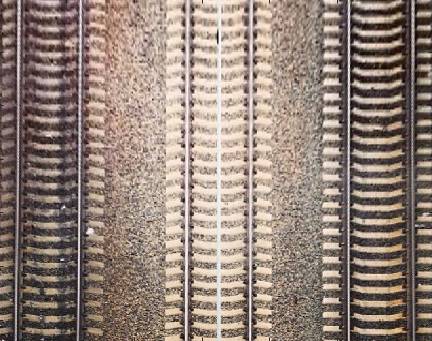}  \\
    
    (a) TS\_GT & (b) Rectified & (c) VPD  & (D) Rectified\\
    
    \end{tabular}
    \vspace{-5pt}
    \caption{\small Sample results for translation symmetry detection. (a) represents the original images of the translation symmetry ground truth (TS\_GT) dataset for which translation symmetry was known to exist. (b) represents rectified outputs of TS\_GT images. (c) represents translation symmetry detected in sample images from VPD dataset, and (d) represents rectified outputs of VPD images.} 
    \vspace{-5pt}
    \label{fig:ts_examples}
\end{figure}

Fig.~\ref{fig:ts_roc} evaluates the success rate of translation symmetry for varying thresholds on two different dataset. First, a small translation symmetry ground truth (TS\_GT) dataset was created consisting of 12 images (6 synthetic, 6 real images) shown in Fig.~\ref{fig:ts_examples}, for which translation symmetry is known to exist. Second, success rate of translation symmetry was calculated on the VPD dataset. We define success rate for translation symmetry ($SR_{ts}$) as 
$$ SR_{ts} = \dfrac{\# \text{ of images with translation symmetry}}{\# \text{ of images in the dataset}}.$$ The success rate was calculated for varying thresholds in the range $t = (0, 0.15)$ for TS\_GT dataset and $t =(0, 1)$ for VPD dataset. It can be seen that success rate reaches a maximum for TS\_GT images at the very low threshold of $t = 0.06$ 

Fig.~\ref{fig:ts_examples} shows different examples in which translation symmetry was detected from both the aforementioned datasets. Column (a) represents images from TS\_GT dataset and column (c) represents images from VPD datasets. Columns to their right represent the rectified outputs of the RPs from their corresponding images. Since URPD is completely unsupervised, we make no assumptions regarding any image. Consequently, we do not have any information regarding the camera parameters with which the image was captured. We maintain this assumption even for the synthetically generated images. Thus the rectified outputs are affine views of the original RPs \cite{Schaffalitzky00a} and therefore, aspect ratios of the rectified outputs do not always match the aspect ratio of the RPs in the real 3D scene (example: first synthetic image of Fig.~\ref{fig:ts_examples} (row 1, columns 1 and 2)). This is also the case in Fig.~{\color{red}9} of the main paper. It should also be noted that all images in our VPD dataset have translation symmetry of the RPs. But the ground truth for the same is unknown. 

\subsection{RP Instance Counting}
\label{sec:object_counting}

RP instance counting differs from object counting that doesn't aim at counting all objects in an image, but focuses on counting the instances that reoccurred.
A typical application for RP instance counting is to count number of product belonging to various types, since each type of products can be viewed as an RP for visual similarity, under various lighting conditions, partial occlusion and non-rigid distortion. 

\begin{table}[h]
\caption{\small  RP Discovery Evaluation on \textbf{Grozi-3.2K}, IOD Threshold $h=0.5$.
\cite{geng2018fine} code is not available thus we cannot compare its performance on \textbf{RP-1K}.
Instance Recall of \cite{geng2018fine} on \textbf{Grozi-3.2K} obtained from Tab.~1 in \cite{geng2018fine}.
The values in \textit{italic} are not statistically significant with each other in the same column.
}
\label{tab:rp_grozi}
\centering
\begin{tabular}{l|cc|cc} 
\toprule
  & \multicolumn{2}{c|}{RP Level} & \multicolumn{2}{c}{RP Instance Level} \\
 Method &     Precision &        Recall &   Precision &      Recall \\
\midrule

\textbf{Baseline\cite{liu2013grasp}} &           0.57 $\pm$ 0.41 &           0.37 $\pm$ 0.32 &           0.67 $\pm$ 0.38 &           0.52 $\pm$ 0.36 \\
\textbf{RESCU-I} &  \textit{0.72 $\pm$ 0.27} &          \textit{ \textbf{0.73 $\pm$ 0.30}} &           0.77 $\pm$ 0.27 &           \textit{0.71 $\pm$ 0.29} \\
\textbf{RESCU-II} &  \textit{\textbf{0.74 $\pm$ 0.27}}  &  \textit{\textbf{0.73 $\pm$ 0.30}} &  \textbf{0.81 $\pm$ 0.26} &  \textit{0.71 $\pm$ 0.29} \\

\textbf{Product Detection \cite{geng2018fine}} & --- & --- & --- & \textbf{0.72}  \\

\bottomrule
\end{tabular}
\end{table}

\textbf{Grozi-3.2K}\cite{george2014recognizing} with 680 test images is used for evaluation. We apply the fine-grained annotation per product type provided by \cite{osokin2020os2d}. We exclude the annotation of certain product types with no RPs. 
Tab.~\ref{tab:RP_dataset} also shows the RP statistic comparison between \textbf{Grozi-3.2K} and our \textbf{RP-1K} dataset. 
%

Tab.~\ref{tab:rp_grozi} shows that our method can achieve a recall rate of over 70\% for both RP and RP instance level recall, which indicates the capability of a recurring pattern detection method on counting re-occurred products in a grocery. Comparing with \cite{geng2018fine} which requires \textit{``logo''} regions as additional input, RESCU can achieve almost the same RP instance level recall rate.
%


\subsection{Enhanced Image Caption from Detected RPs}
\label{sec:captions}

Using an off-the-shelf image captioning model, we utilize detected RP information from RESCU to augment generated captions. To achieve this, we employ OFA, a unified multimodal pre-trained network \cite{wang2022unifying}. OFA is a Transformer based, modal-agnostic pre-trained network that has achieved state-of-the-art results in multiple multimodal benchmarks including image captioning. We use the \textit{Large} size OFA model and weights pre-trained on the MS COCO Caption dataset \cite{chen2015microsoft} for generating captions.

\begin{figure}[]
\centering
\begin{subfigure}[t]{0.3\linewidth}
    \centering
    \includegraphics[width=1\linewidth]{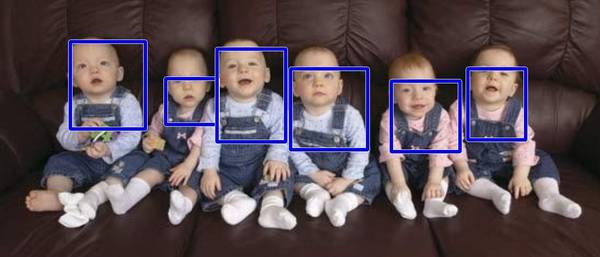}
    \caption*{\sout{A~group~of} \textcolor{red}{six similar} babies sitting on the couch.}
\end{subfigure}
\unskip\ \vrule\
\begin{subfigure}[t]{0.3\linewidth}
    \centering
    \includegraphics[width=1\linewidth]{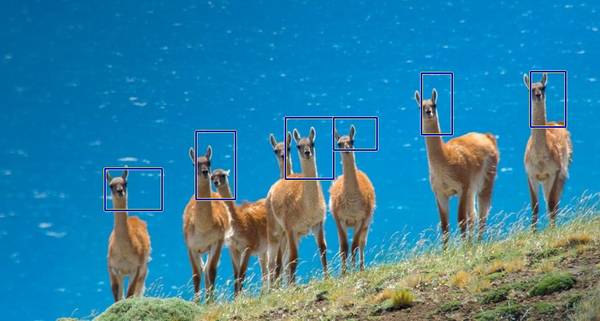}
    \caption*{\sout{A~heard~of} \textcolor{red}{six~similar} llamas standing on top of a hill.}
\end{subfigure}
\unskip\ \vrule\
\begin{subfigure}[t]{0.3\linewidth}
    \centering
    \includegraphics[width=1\linewidth]{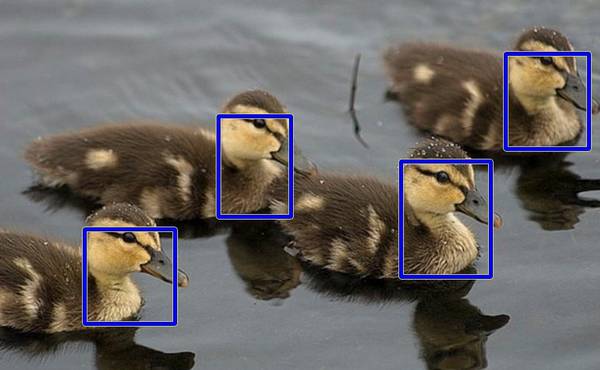}
    \caption*{\sout{A~group~of} \textcolor{red}{four~similar} ducks swimming in the water.}
\end{subfigure}
\\
\vfill
\par\noindent\rule{\textwidth}{0.4pt}
\begin{subfigure}[t]{0.3\linewidth}
    \centering
    \includegraphics[width=1\linewidth]{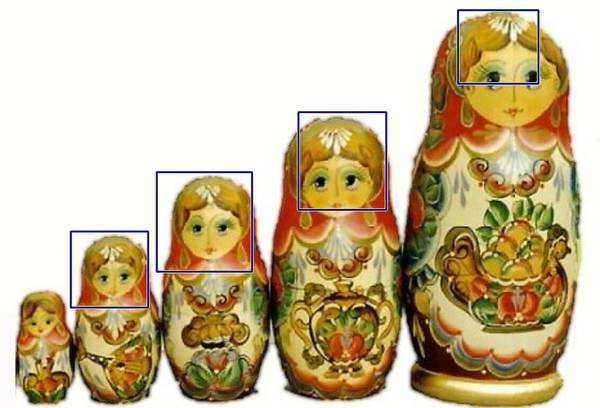}
    \caption*{\sout{A~group~of} \textcolor{red}{four~similar} Russian nesting dolls on a white background.}
\end{subfigure}
\unskip\ \vrule\
\begin{subfigure}[t]{0.3\linewidth}
    \centering
    \includegraphics[width=1\linewidth]{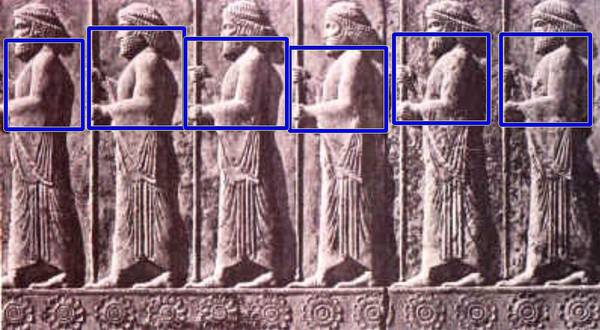}
    \caption*{An old picture of \textcolor{red}{six~similar} stone statues on a wall.}
\end{subfigure}
\unskip\ \vrule\
\begin{subfigure}[t]{0.3\linewidth}
    \centering
    \includegraphics[width=1\linewidth]{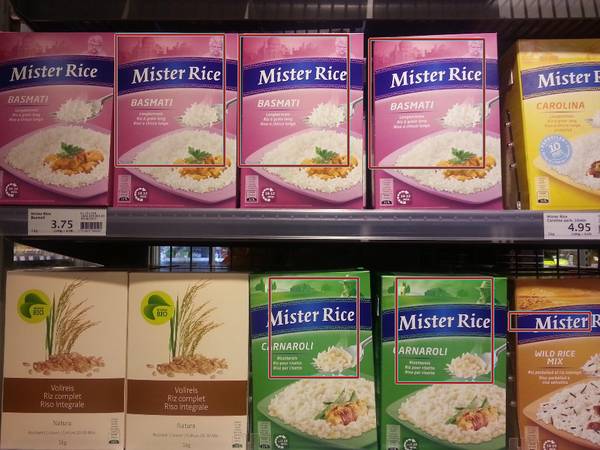}
    \caption*{\sout{A~row~of} \textcolor{red}{six~similar} mister rice boxes on a supermarket shelf.}
\end{subfigure}
\\
\vfill
\par\noindent\rule{\textwidth}{0.4pt}
\begin{subfigure}[t]{0.3\linewidth}
    \centering
    \includegraphics[width=1\linewidth]{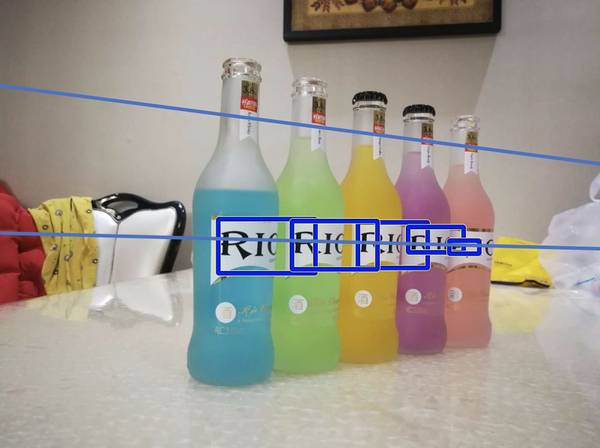}
    \caption*{\sout{A~group~of} \textcolor{red}{five~similar} bottles with different colored liquid in them on a table. \textcolor{blue}{The~bottles~have~a~potential~translation~symmetry~in~3D} and \textcolor{green}{form~a~vanishing~point~outside~of~the~image}.}
\end{subfigure}
\unskip\ \vrule\
\begin{subfigure}[t]{0.3\linewidth}
    \centering
    \includegraphics[width=1\linewidth]{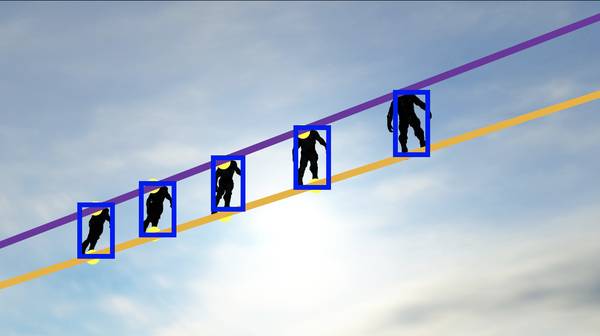}
    \caption*{\sout{A~group~of} \textcolor{red}{five~similar} men jumping in the sky. \textcolor{blue}{The~men~have~a~potential~translation~symmetry~in~3D} and \textcolor{green}{form~a~vanishing~point~outside~of~the~image}.}
\end{subfigure}
\unskip\ \vrule\
\begin{subfigure}[t]{0.3\linewidth}
    \centering
    \includegraphics[width=1\linewidth]{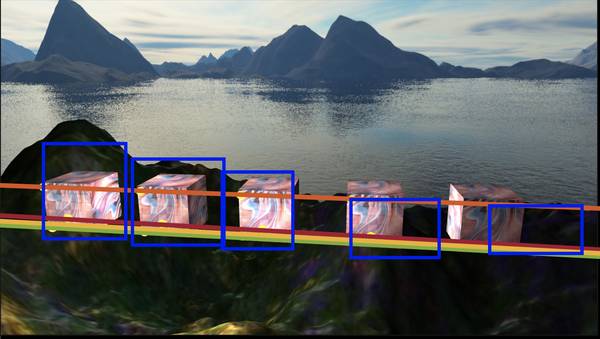}
    \caption*{\sout{A~series~of} \textcolor{red}{five~similar} canvases in front of a lake and mountains. \textcolor{blue}{The canvases~have~a~potential~translation~symmetry~in~3D} and  \textcolor{green}{form~a~vanishing~point~outside~of~the~image}.}
\end{subfigure}
\caption{\small Example image-caption pairs we enhance using detected RP and VP information. The captions below each image contain the original text generated by OFA \cite{wang2022unifying}. We add the dominant RP's \textcolor{red}{count} in red following collective nouns. If detected, additional \textcolor{blue}{translation symmetry} and \textcolor{green}{VP} information is added to the caption.}
\label{fig:captions}
\end{figure}

\begin{figure*}[]
    \centering
    \includegraphics[width=\linewidth]{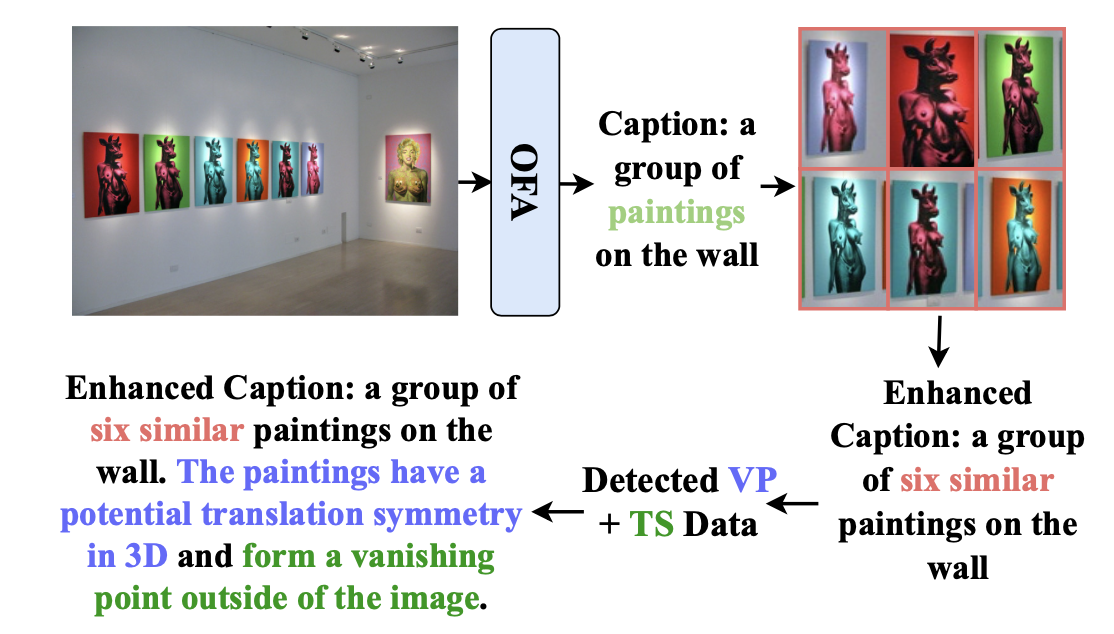}
    \caption{\small  Proposed image caption enhancement pipeline. We obtain image captions using OFA \cite{wang2022unifying}. We parse the sentence for collective nouns or a subject noun and use the previously discovered dominant RP count for that corresponding image to add the detected noun's count. We further add discovered VP and TS information to the final caption. }
    \label{fig:cap_ex}
\end{figure*}

To enhance the captions obtained from OFA, we assume RP, VP, and TS information has been previously detected for the image. We begin by generating a caption for the image using OFA. We then parse the image's generated caption for instances of collective nouns such as ``group''. If no collective nouns are found, we search the string for noun instances proceeded by the word ``of''. Finally, we follow a simple grammar to replace collective nouns and words dependent on them with the dominant RP's count discovered by RESCU. If no collective nouns are detected, we place the RP's count after ``of'' and before the detected noun. In addition to enhancing the image caption with RP counts, we add further context by noting potential translation symmetry and whether a detected vanishing point lies inside or outside the image borders. Figure \ref{fig:captions} shows example image-caption pairs augmented by our detected RP and VP information. 
The proposed caption enhancing pipeline is shown in Figure \ref{fig:cap_ex}.

\begin{figure*}[h]
    \centering
    \includegraphics[width=\linewidth, height=0.45\textwidth]{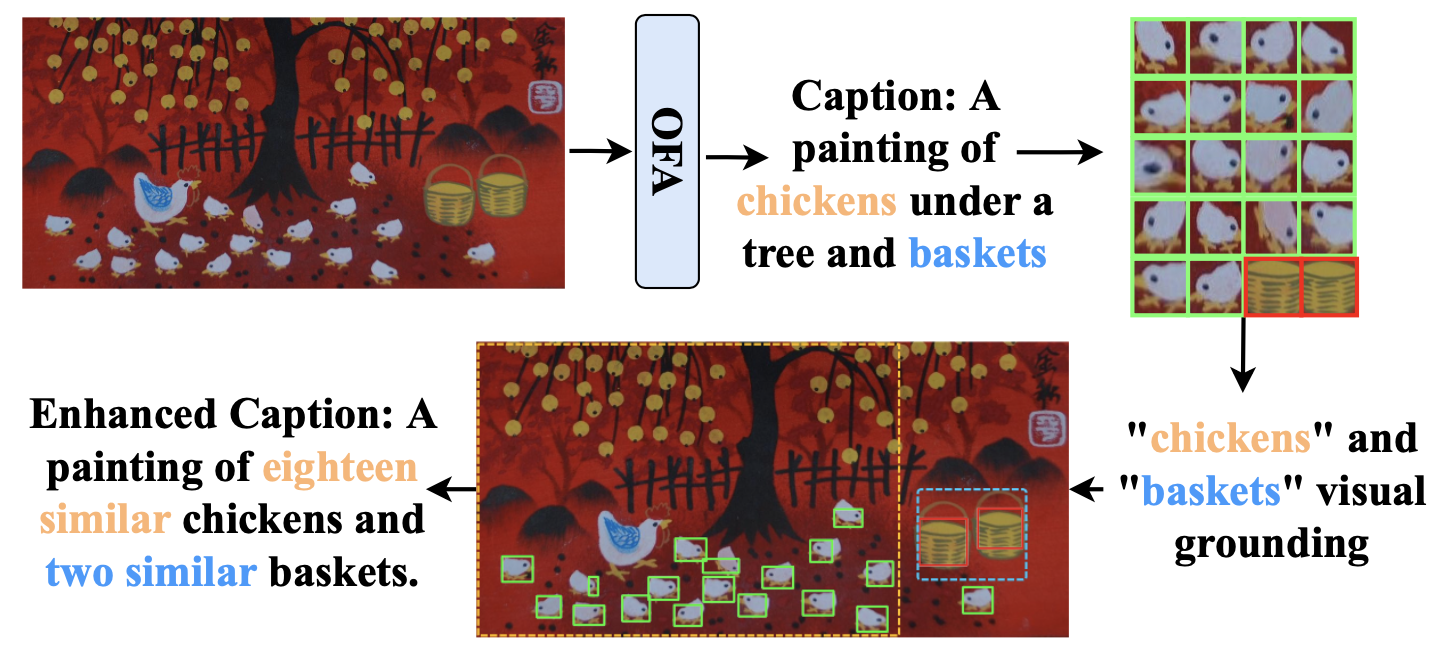}
    \caption{\small  Caption enhancement using two detected RPs. We utilize OFA's visual grounding for the prompts ``\textcolor{orange}{chickens}'' and ``\textcolor{cyan}{baskets}'' to determine which noun subject the RP is most likely to corresponds with.}
    \label{fig:two_rp}
\end{figure*}

We further utilize OFA to handle multiple potential RPs and their corresponding nouns in a caption. If the image's caption contains multiple subject nouns, we use OFA's visual grounding to ground each detected subject noun in the caption. This gives OFA's bounding box for the potential region of the text input. We then check the detected RPs overlap with OFA's proposed region. We assign the corresponding RP to the region and its corresponding noun if the RP's total area has at least 0.90 overlap with OFA's proposed region. Figure \ref{fig:two_rp} demonstrates an example image-caption pair enhanced with multiple detected RPs and OFA's proposed regions for two detected nouns in the image's caption.

\section{Limitations}
\label{sec:limitations}
\begin{figure*}[!h] 
\centering
\foreach \id/\picname in {1/new200, 2/new758,  3/new797}
{ %
    \begin{subfigure}[t]{0.3\linewidth} \centering
        \includegraphics[height = \textwidth, width = \textwidth,valign=t]{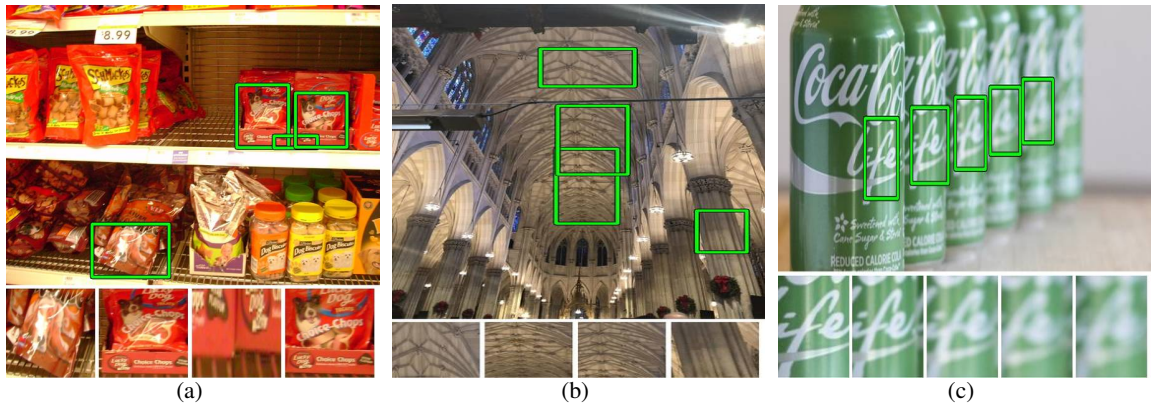}
        \vspace{-6pt}
        \caption{\small \label{fig:failure_\picname}}
    \end{subfigure} 
} 
\caption{\small Limitations of RESCU. \textbf{(a)} Wrongly involves dissimilar RP instance due to deformation and non-uniform lighting. \textbf{(b)} Wrongly involves dissimilar RP instance due to the interference of texture. \textbf{(c)} Misses the furthest RP instance due to blurring. }
\label{fig:failure_ex}
\vspace{-10pt}
\end{figure*}

Fig.~\ref{fig:failure_ex} shows some limitations of our approach. The method may wrongly detect RP on images with large distortion, non-uniform lighting, and blurring. To address the limitations under these scenarios, our future work approach will be to include additional varying types of extracted initial features to aid and complement the current use of SIFT features. Stage-II will also be improved to enhance the concept of learning similarity under different scenarios.


\section{Conclusion}
We have illustrated our path from recurrence discovery and representation in a single image as RPs and RP instances to scene understanding in terms of vanishing point, potential 3D translation symmetry and number of recurrences of an unknown ``object". By leveraging some off-the-shelf image caption tools, we demonstrated by examples the feasibility of using knowledge learned from RPs on a single image to produce enhanced, quantitative and detailed captions as .
Our 2-stage RESCU differs from all existing unsupervised learning systems in design andwhich advances the state of the art in URPD. Furthermore, we offer a new benchmark for future URPD algorithm evaluation, and a new RP-based Vanishing Point image set for unsupervised Vanishing Point Discovery (VPD).
Our demo video link: \href{https://drive.google.com/file/d/1YUeJiuY4IIpbNTvBacMyXcgbbLikCqQo/view?usp=sharing}{a Demo Video}
%

\COMMENT{
especially for supermarket goods counting and VP detection tasks. 
Novel research directions to explore may include, 
for example, 
the discrepancy and interplay between human and machine perception of recurring patterns, to ultimately advance our understanding of human-machine visual perception in a diverse and dynamically changing real world, in an unsupervised manner. 
}

\section{Acknowledgement}
This work is supported in part by an NSF grant \# 1909315.
The authors would like to thank Jesse Scott and Chris Funk for their help during the paper preparation.

\COMMENT{
Fig.~\ref{fig:unsupervised_improved_failure_case} shows some failure detections by the improved unsupervised method.

\begin{figure}[ht!]
    \centering
    \includegraphics[width=0.48\linewidth]{figures/select_results/unsupervised_improved/failure/new012_0.jpg}
    \includegraphics[width=0.48\linewidth]{figures/select_results/unsupervised_improved/failure/new018_5.jpg}
    
    \includegraphics[width=0.48\linewidth]{figures/select_results/unsupervised_improved/failure/new044_0.jpg}
    \includegraphics[width=0.48\linewidth]{figures/select_results/unsupervised_improved/failure/new752_1.jpg}
    \caption{\small Four examples of Negative detections. These failure detections are caused by: tolerant to missing features, limited local view, improper feature extraction.}
    \label{fig:unsupervised_improved_failure_case}
\end{figure}
}
\clearpage


\bibliographystyle{splncs}
\bibliography{egbib}

\begin{thebibliography}{10}

\bibitem{li2009computational}
Li, Y., Pizlo, Z., Steinman, R.M.:
\newblock A computational model that recovers the 3d shape of an object from a
  single 2d retinal representation.
\newblock Vision research \textbf{49} (2009)  979--991

\bibitem{treder2010behind}
Treder, M.S.:
\newblock Behind the looking-glass: A review on human symmetry perception.
\newblock Symmetry \textbf{2} (2010)  1510--1543

\bibitem{michaux2016figure}
Michaux, A., Jayadevan, V., Delp~III, E.J., Pizlo, Z.:
\newblock Figure-ground organization based on three-dimensional symmetry.
\newblock Journal of Electronic Imaging \textbf{25} (2016)  1--11

\bibitem{sawada2014detecting}
Sawada, T., Li, Y., Pizlo, Z.:
\newblock Detecting 3-d mirror symmetry in a 2-d camera image for 3-d shape
  recovery.
\newblock Proceedings of the IEEE \textbf{102} (2014)  1588--1606

\bibitem{liu2013grasp}
Liu, J., Liu, Y.:
\newblock Grasp recurring patterns from a single view.
\newblock In: 2013 IEEE Conference on Computer Vision and Pattern Recognition.
  (2013)  2003--2010

\bibitem{mitra2006partial}
Mitra, N.J., Guibas, L.J., Pauly, M.:
\newblock Partial and approximate symmetry detection for 3d geometry.
\newblock ACM Transactions on Graphics (TOG) \textbf{25} (2006)  560--568

\bibitem{liu2010common}
Liu, H., Yan, S.:
\newblock Common visual pattern discovery via spatially coherent
  correspondences.
\newblock In: 2010 IEEE Computer Society Conference on Computer Vision and
  Pattern Recognition, IEEE (2010)  1609--1616

\bibitem{mattson2014superior}
Mattson, M.P.:
\newblock Superior pattern processing is the essence of the evolved human
  brain.
\newblock Frontiers in neuroscience \textbf{8} (2014)  1--17

\bibitem{mundy1992projective}
Mundy, J.L., Zisserman, A., eds.:
\newblock Geometric Invariance in Computer Vision.
\newblock MIT Press, Cambridge, MA, USA (1992)

\bibitem{Schaffalitzky00a}
Schaffalitzky, F., Zisserman, A.:
\newblock Planar grouping for automatic detection of vanishing lines and
  points.
\newblock Image and Vision Computing \textbf{8} (2000)  647--658

\bibitem{weyl2015symmetry}
Weyl, H.:
\newblock Symmetry. Volume~47.
\newblock Princeton University Press (2015)

\bibitem{criminisciSingleViewMetrology2000}
Criminisi, A., Reid, I., Zisserman, A.:
\newblock Single view metrology.
\newblock International Journal of Computer Vision \textbf{40} (2000)  123--148

\bibitem{singleviewmetrologyinthewild2020}
Zhu, R., Yang, X., Hold-Geoffroy, Y., Perazzi, F., Eisenmann, J., Sunkavalli,
  K., Chandraker, M.:
\newblock Single view metrology in the wild.
\newblock In Vedaldi, A., Bischof, H., Brox, T., Frahm, J.M., eds.: Computer
  Vision -- ECCV 2020, Cham, Springer International Publishing (2020)  316--333

\bibitem{cho2010reweighted}
Cho, M., Lee, J., Lee, K.M.:
\newblock Reweighted random walks for graph matching.
\newblock In: European conference on Computer vision, Springer (2010)  492--505

\bibitem{rother2006cosegmentation}
Rother, C., Minka, T., Blake, A., Kolmogorov, V.:
\newblock Cosegmentation of image pairs by histogram matching-incorporating a
  global constraint into mrfs.
\newblock In: 2006 IEEE Computer Society Conference on Computer Vision and
  Pattern Recognition (CVPR'06). Volume~1., IEEE (2006)  993--1000

\bibitem{toshev2007image}
Toshev, A., Shi, J., Daniilidis, K.:
\newblock Image matching via saliency region correspondences.
\newblock In: 2007 IEEE Conference on Computer Vision and Pattern Recognition,
  IEEE (2007)  1--8

\bibitem{kannala2008object}
Kannala, J., Rahtu, E., Brandt, S.S., Heikkila, J.:
\newblock Object recognition and segmentation by non-rigid quasi-dense
  matching.
\newblock In: 2008 IEEE Conference on Computer Vision and Pattern Recognition,
  IEEE (2008)  1--8

\bibitem{yuan2007spatial}
Yuan, J., Wu, Y.:
\newblock Spatial random partition for common visual pattern discovery.
\newblock In: 2007 IEEE 11th International Conference on Computer Vision, IEEE
  (2007)  1--8

\bibitem{cho2008co}
Cho, M., Shin, Y.M., Lee, K.M.:
\newblock Co-recognition of image pairs by data-driven monte carlo image
  exploration.
\newblock In: European conference on computer vision, Springer (2008)  144--157

\bibitem{cho2009feature}
Cho, M., Lee, J., Lee, K.M.:
\newblock Feature correspondence and deformable object matching via
  agglomerative correspondence clustering.
\newblock In: 2009 IEEE 12th International Conference on Computer Vision, IEEE
  (2009)  1280--1287

\bibitem{cho2010unsupervised}
Cho, M., Shin, Y.M., Lee, K.M.:
\newblock Unsupervised detection and segmentation of identical objects.
\newblock In: 2010 IEEE Computer Society Conference on Computer Vision and
  Pattern Recognition, IEEE (2010)  1617--1624

\bibitem{gao2009unsupervised}
Gao, J., Hu, Y., Liu, J., Yang, R.:
\newblock Unsupervised learning of high-order structural semantics from images.
\newblock In: 2009 IEEE 12th International Conference on Computer Vision, IEEE
  (2009)  2122--2129

\bibitem{rubinstein2013unsupervised}
Rubinstein, M., Joulin, A., Kopf, J., Liu, C.:
\newblock Unsupervised joint object discovery and segmentation in internet
  images.
\newblock In: Proceedings of the IEEE conference on computer vision and pattern
  recognition. (2013)  1939--1946

\bibitem{hong2014unsupervised}
Hong, Y., Si, Z., Hu, W., Zhu, S.C., Wu, Y.N.:
\newblock Unsupervised learning of compositional sparse code for natural image
  representation.
\newblock Quarterly of Applied Mathematics (2014)  373--406

\bibitem{faktor2013clustering}
Faktor, A., Irani, M.:
\newblock “clustering by composition”—unsupervised discovery of image
  categories.
\newblock IEEE transactions on pattern analysis and machine intelligence
  \textbf{36} (2013)  1092--1106

\bibitem{sivic2008unsupervised}
Sivic, J., Russell, B.C., Zisserman, A., Freeman, W.T., Efros, A.A.:
\newblock Unsupervised discovery of visual object class hierarchies.
\newblock In: 2008 IEEE Conference on Computer Vision and Pattern Recognition,
  IEEE (2008)  1--8

\bibitem{todorovic2008unsupervised}
Todorovic, S., Ahuja, N.:
\newblock Unsupervised category modeling, recognition, and segmentation in
  images.
\newblock IEEE Transactions on Pattern Analysis and Machine Intelligence
  \textbf{30} (2008)  2158--2174

\bibitem{fergus2003object}
Fergus, R., Perona, P., Zisserman, A.:
\newblock Object class recognition by unsupervised scale-invariant learning.
\newblock In: 2003 IEEE Computer Society Conference on Computer Vision and
  Pattern Recognition, 2003. Proceedings. Volume~2., IEEE (2003)  1--8

\bibitem{vo2021large}
Vo, V.H., Sizikova, E., Schmid, C., P{\'e}rez, P., Ponce, J.:
\newblock Large-scale unsupervised object discovery.
\newblock Advances in Neural Information Processing Systems \textbf{34} (2021)
  16764--16778

\bibitem{vo2020toward}
Vo, H.V., P{\'e}rez, P., Ponce, J.:
\newblock Toward unsupervised, multi-object discovery in large-scale image
  collections.
\newblock In: European Conference on Computer Vision, Springer (2020)  779--795

\bibitem{huberman2016detecting}
Huberman, I., Fattal, R.:
\newblock Detecting repeating objects using patch correlation analysis.
\newblock In: 2016 IEEE Conference on Computer Vision and Pattern Recognition
  (CVPR), Los Alamitos, CA, USA, IEEE Computer Society (2016)  2903--2911

\bibitem{lettry2017repeated}
Lettry, L., Perdoch, M., Vanhoey, K., Van~Gool, L.:
\newblock Repeated pattern detection using cnn activations.
\newblock In: 2017 IEEE Winter Conference on Applications of Computer Vision
  (WACV), IEEE (2017)  47--55

\bibitem{rodriguez2019automatic}
Rodriguez-Pardo, C., Suja, S., Pascual, D., Lopez-Moreno, J., Garces, E.:
\newblock Automatic extraction and synthesis of regular repeatable patterns.
\newblock Computers \& Graphics \textbf{83} (2019)  33--41

\bibitem{geng2018fine}
Geng, W., Han, F., Lin, J., Zhu, L., Bai, J., Wang, S., He, L., Xiao, Q., Lai,
  Z.:
\newblock Fine-grained grocery product recognition by one-shot learning.
\newblock In: Proceedings of the 26th ACM international conference on
  Multimedia. (2018)  1706--1714

\bibitem{zhang2019objects}
Zhang, Z., Jing, T., Tian, C., Cui, P., Li, X., Gao, M.:
\newblock Objects discovery based on co-occurrence word model with anchor-box
  polishing.
\newblock IEEE Transactions on Circuits and Systems for Video Technology
  \textbf{30} (2019)  632--645

\bibitem{shi2022represent}
Shi, M., Lu, H., Feng, C., Liu, C., Cao, Z.:
\newblock Represent, compare, and learn: A similarity-aware framework for
  class-agnostic counting.
\newblock In: Proceedings of the IEEE/CVF Conference on Computer Vision and
  Pattern Recognition. (2022)  9529--9538

\bibitem{cho2015unsupervised}
Cho, M., Kwak, S., Schmid, C., Ponce, J.:
\newblock Unsupervised object discovery and localization in the wild:
  Part-based matching with bottom-up region proposals.
\newblock In: Proceedings of the IEEE conference on computer vision and pattern
  recognition. (2015)  1201--1210

\bibitem{wei2019unsupervised}
Wei, X.S., Zhang, C.L., Wu, J., Shen, C., Zhou, Z.H.:
\newblock Unsupervised object discovery and co-localization by deep descriptor
  transformation.
\newblock Pattern Recognition \textbf{88} (2019)  113--126

\bibitem{simeoni2021localizing}
Sim{\'e}oni, O., Puy, G., Vo, H.V., Roburin, S., Gidaris, S., Bursuc, A.,
  P{\'e}rez, P., Marlet, R., Ponce, J.:
\newblock Localizing objects with self-supervised transformers and no labels.
\newblock In: BMVC. (2021)

\bibitem{noroozi2016unsupervised}
Noroozi, M., Favaro, P.:
\newblock Unsupervised learning of visual representations by solving jigsaw
  puzzles.
\newblock In: European conference on computer vision, Springer (2016)  69--84

\bibitem{misra2020self}
Misra, I., Maaten, L.v.d.:
\newblock Self-supervised learning of pretext-invariant representations.
\newblock In: Proceedings of the IEEE/CVF Conference on Computer Vision and
  Pattern Recognition. (2020)  6707--6717

\bibitem{pathak2016context}
Pathak, D., Krahenbuhl, P., Donahue, J., Darrell, T., Efros, A.A.:
\newblock Context encoders: Feature learning by inpainting.
\newblock In: Proceedings of the IEEE conference on computer vision and pattern
  recognition. (2016)  2536--2544

\bibitem{lin2014microsoft}
Lin, T.Y., Maire, M., Belongie, S., Hays, J., Perona, P., Ramanan, D., Dollár,
  P., Zitnick, C.L.:
\newblock Microsoft coco: Common objects in context.
\newblock Computer Vision – ECCV 2014 \textbf{8693} (2014)  740–--755

\bibitem{everingham2010pascal}
Everingham, M., Van~Gool, L., Williams, C.K., Winn, J., Zisserman, A.:
\newblock The pascal visual object classes (voc) challenge.
\newblock International journal of computer vision \textbf{88} (2010)  303--338

\bibitem{george2014recognizing}
George, M., Floerkemeier, C.:
\newblock Recognizing products: A per-exemplar multi-label image classification
  approach.
\newblock In: European Conference on Computer Vision, Springer (2014)  440--455

\bibitem{hsieh2017drone}
Hsieh, M.R., Lin, Y.L., Hsu, W.H.:
\newblock Drone-based object counting by spatially regularized regional
  proposal network.
\newblock In: Proceedings of the IEEE international conference on computer
  vision. (2017)  4145--4153

\bibitem{cordts2016cityscapes}
Cordts, M., Omran, M., Ramos, S., Rehfeld, T., Enzweiler, M., Benenson, R.,
  Franke, U., Roth, S., Schiele, B.:
\newblock The cityscapes dataset for semantic urban scene understanding.
\newblock In: Proceedings of the IEEE conference on computer vision and pattern
  recognition. (2016)  3213--3223

\bibitem{patil2022p3depth}
Patil, V., Sakaridis, C., Liniger, A., Van~Gool, L.:
\newblock P3depth: Monocular depth estimation with a piecewise planarity prior.
\newblock In: Proceedings of the IEEE/CVF Conference on Computer Vision and
  Pattern Recognition. (2022)  1610--1621

\bibitem{li2022binsformer}
Li, Z., Wang, X., Liu, X., Jiang, J.:
\newblock Binsformer: Revisiting adaptive bins for monocular depth estimation.
\newblock arXiv preprint arXiv:2204.00987 (2022)

\bibitem{yan2021channel}
Yan, J., Zhao, H., Bu, P., Jin, Y.:
\newblock Channel-wise attention-based network for self-supervised monocular
  depth estimation.
\newblock In: 2021 International Conference on 3D Vision (3DV), IEEE (2021)
  464--473

\bibitem{wang2022spatiality}
Wang, H., Zhang, C., Yu, J., Cai, W.:
\newblock Spatiality-guided transformer for 3d dense captioning on point
  clouds.
\newblock arXiv preprint arXiv:2204.10688 (2022)

\bibitem{azuma2022scanqa}
Azuma, D., Miyanishi, T., Kurita, S., Kawanabe, M.:
\newblock Scanqa: 3d question answering for spatial scene understanding.
\newblock In: Proceedings of the IEEE/CVF Conference on Computer Vision and
  Pattern Recognition. (2022)  19129--19139

\bibitem{xu2022groupvit}
Xu, J., De~Mello, S., Liu, S., Byeon, W., Breuel, T., Kautz, J., Wang, X.:
\newblock Groupvit: Semantic segmentation emerges from text supervision.
\newblock In: Proceedings of the IEEE/CVF Conference on Computer Vision and
  Pattern Recognition. (2022)  18134--18144

\bibitem{lambert2020mseg}
Lambert, J., Liu, Z., Sener, O., Hays, J., Koltun, V.:
\newblock Mseg: A composite dataset for multi-domain semantic segmentation.
\newblock In: Proceedings of the IEEE/CVF conference on computer vision and
  pattern recognition. (2020)  2879--2888

\bibitem{rother2002new}
Rother, C.:
\newblock A new approach to vanishing point detection in architectural
  environments.
\newblock Image and Vision Computing \textbf{20} (2002)  647--655

\bibitem{shi2015fast}
Shi, J., Wang, J., Fu, F.:
\newblock Fast and robust vanishing point detection for unstructured road
  following.
\newblock IEEE Transactions on Intelligent Transportation Systems \textbf{17}
  (2015)  970--979

\bibitem{zhou2017detecting}
Zhou, Z., Farhat, F., Wang, J.Z.:
\newblock Detecting dominant vanishing points in natural scenes with
  application to composition-sensitive image retrieval.
\newblock IEEE Transactions on Multimedia \textbf{19} (2017)  2651--2665

\bibitem{zhou2019neurvps}
Zhou, Y., Qi, H., Huang, J., Ma, Y.
\newblock In: NeurVPS: Neural Vanishing Point Scanning via Conic Convolution.
  Volume~32. Curran Associates Inc., Red Hook, NY, USA (2019)

\bibitem{mundy1993repeated}
Mundy, J.L., Zisserman, A.:
\newblock Repeated structures: Image correspondence constraints and 3d
  structure recovery.
\newblock In: Joint European-US Workshop on Applications of Invariance in
  Computer Vision, Springer (1993)  89--106

\bibitem{goldman2019precise}
Goldman, E., Herzig, R., Eisenschtat, A., Goldberger, J., Hassner, T.:
\newblock Precise detection in densely packed scenes.
\newblock In: Proceedings of the IEEE/CVF Conference on Computer Vision and
  Pattern Recognition. (2019)  5227--5236

\bibitem{lin2017focal}
Lin, T.Y., Goyal, P., Girshick, R., He, K., Doll{\'a}r, P.:
\newblock Focal loss for dense object detection.
\newblock In: Proceedings of the IEEE international conference on computer
  vision. (2017)  2980--2988

\bibitem{lei2021towards}
Lei, Y., Liu, Y., Zhang, P., Liu, L.:
\newblock Towards using count-level weak supervision for crowd counting.
\newblock Pattern Recognition \textbf{109} (2021)  107616

\bibitem{lu2018class}
Lu, E., Xie, W., Zisserman, A.:
\newblock Class-agnostic counting.
\newblock In: Asian conference on computer vision, Springer (2018)  669--684

\bibitem{ranjan2021learning}
Ranjan, V., Sharma, U., Nguyen, T., Hoai, M.:
\newblock Learning to count everything.
\newblock In: Proceedings of the IEEE/CVF Conference on Computer Vision and
  Pattern Recognition. (2021)  3394--3403

\bibitem{hobley2022learning}
Hobley, M., Prisacariu, V.:
\newblock Learning to count anything: Reference-less class-agnostic counting
  with weak supervision.
\newblock arXiv preprint arXiv:2205.10203 (2022)

\bibitem{zhang2021repetitive}
Zhang, Y., Shao, L., Snoek, C.G.:
\newblock Repetitive activity counting by sight and sound.
\newblock In: Proceedings of the IEEE/CVF Conference on Computer Vision and
  Pattern Recognition. (2021)  14070--14079

\bibitem{dwibedi2020counting}
Dwibedi, D., Aytar, Y., Tompson, J., Sermanet, P., Zisserman, A.:
\newblock Counting out time: Class agnostic video repetition counting in the
  wild.
\newblock In: Proceedings of the IEEE/CVF conference on computer vision and
  pattern recognition. (2020)  10387--10396

\bibitem{stefanini2022show}
Stefanini, M., Cornia, M., Baraldi, L., Cascianelli, S., Fiameni, G.,
  Cucchiara, R.:
\newblock From show to tell: A survey on deep learning-based image captioning.
\newblock IEEE Transactions on Pattern Analysis and Machine Intelligence (2022)
   1--20

\bibitem{ren2015faster}
Ren, S., He, K., Girshick, R., Sun, J.:
\newblock Faster r-cnn: Towards real-time object detection with region proposal
  networks.
\newblock In Cortes, C., Lawrence, N., Lee, D., Sugiyama, M., Garnett, R.,
  eds.: Advances in Neural Information Processing Systems. Volume~28., Curran
  Associates, Inc. (2015)

\bibitem{anderson2018bottom}
Anderson, P., He, X., Buehler, C., Teney, D., Johnson, M., Gould, S., Zhang,
  L.:
\newblock Bottom-up and top-down attention for image captioning and visual
  question answering.
\newblock In: Proceedings of the IEEE conference on computer vision and pattern
  recognition. (2018)  6077--6086

\bibitem{li2020oscar}
Li, X., Yin, X., Li, C., Zhang, P., Hu, X., Zhang, L., Wang, L., Hu, H., Dong,
  L., Wei, F.,  et~al.:
\newblock Oscar: Object-semantics aligned pre-training for vision-language
  tasks.
\newblock In: European Conference on Computer Vision, Springer (2020)  121--137

\bibitem{wang2022unifying}
Wang, P., Yang, A., Men, R., Lin, J., Bai, S., Li, Z., Ma, J., Zhou, C., Zhou,
  J., Yang, H.:
\newblock Unifying modalities, tasks, and architectures through a simple
  sequence-to-sequence learning framework.
\newblock In: Proceedings of the 39th International Conference on Machine
  Learning, {ICML} 2022. Proceedings of Machine Learning Research, {PMLR}
  (2022)

\bibitem{zhou2020unified}
Zhou, L., Palangi, H., Zhang, L., Hu, H., Corso, J., Gao, J.:
\newblock Unified vision-language pre-training for image captioning and vqa.
\newblock Proceedings of the AAAI Conference on Artificial Intelligence
  \textbf{34} (2020)  13041--13049

\bibitem{feo1995greedy}
Feo, T.A., Resende, M.G.:
\newblock Greedy randomized adaptive search procedures.
\newblock Journal of global optimization \textbf{6} (1995)  109--133

\bibitem{zitnick2014edge}
Zitnick, C.L., Doll{\'a}r, P.:
\newblock Edge boxes: Locating object proposals from edges.
\newblock In Fleet, D., Pajdla, T., Schiele, B., Tuytelaars, T., eds.: European
  Conference on Computer Vision, Cham, Springer International Publishing (2014)
   391--405

\bibitem{huang2017densely}
Huang, G., Liu, Z., Van Der~Maaten, L., Weinberger, K.Q.:
\newblock Densely connected convolutional networks.
\newblock In: Proceedings of the IEEE conference on computer vision and pattern
  recognition. (2017)  4700--4708

\bibitem{ester1996density}
Ester, M., Kriegel, H.P., Sander, J., Xu, X.:
\newblock A density-based algorithm for discovering clusters in large spatial
  databases with noise.
\newblock In: Proceedings of the Second International Conference on Knowledge
  Discovery and Data Mining. KDD'96, AAAI Press (1996)  226–--231

\bibitem{deng2009imagenet}
Deng, J., Dong, W., Socher, R., Li, L.J., Li, K., Fei-Fei, L.:
\newblock Imagenet: A large-scale hierarchical image database.
\newblock In: 2009 IEEE conference on computer vision and pattern recognition,
  Ieee (2009)  248--255

\bibitem{zhang2009handling}
Zhang, Y., Wang, L., Hartley, R., Li, H.:
\newblock Handling significant scale difference for object retrieval in a
  supermarket.
\newblock In: 2009 Digital Image Computing: Techniques and Applications, IEEE
  (2009)  468--475

\bibitem{osokin2020os2d}
Osokin, A., Sumin, D., Lomakin, V.:
\newblock Os2d: One-stage one-shot object detection by matching anchor
  features.
\newblock In: European Conference on Computer Vision, Springer (2020)  635--652

\bibitem{chen2015microsoft}
Chen, X., Fang, H., Lin, T.Y., Vedantam, R., Gupta, S., Doll{\'a}r, P.,
  Zitnick, C.L.:
\newblock Microsoft coco captions: Data collection and evaluation server.
\newblock arXiv preprint arXiv:1504.00325 (2015)

\bibitem{kingma2014adam}
Kingma, D.P., Ba, J.:
\newblock Adam: A method for stochastic optimization.
\newblock arXiv preprint arXiv:1412.6980 (2014)

\end{thebibliography}

\newpage
\section{Appendix}
\subsection{Varying IOD Threshold}
\begin{figure}[tb!]
\centering
\foreach \id/\subset in {1/All, 2/Front View, 3/Projective View, 4/Man-Made Rigid, 5/Man-Made Deformable, 6/Painting, 7/Animal Human}
{ %
    \foreach \x/\metric in {1/RP_Prec, 2/RP_Rec, 3/Inst_Prec, 4/Inst_Rec}
    { %
    \begin{subfigure}[t]{0.23\linewidth}
        \if \id 1
            \if \x 1 \caption{\small \scriptsize{RP Precision}} \fi
            \if \x 2 \caption{\small \scriptsize{RP Recall}} \fi
            \if \x 3 \caption{\small \scriptsize{Instance Precision}} \fi
            \if \x 4 \caption{\small \scriptsize{Instance Recall}} \fi
        \fi
        \includegraphics[width = \textwidth,valign=t]{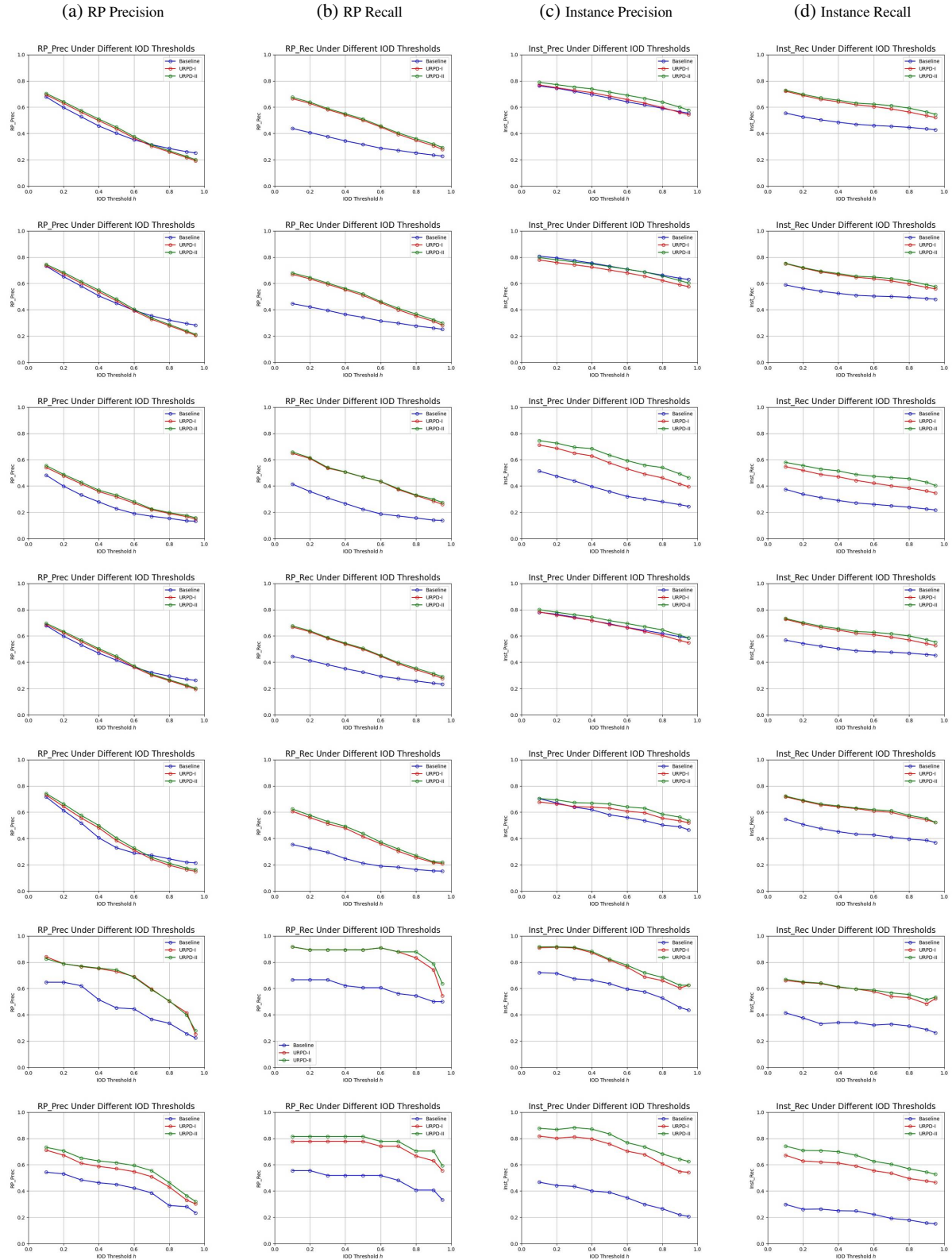} 
    \end{subfigure} 
    \hfill
    }
    \vspace{2pt}
    
} \vspace{-5pt}
\caption{\small IOD Curves showing each evaluation metric at different IOD thresholds $h$.
\textcolor{blue}{Blue}: the \textbf{baseline} method\cite{liu2013grasp}. \textcolor{red}{Red}: ours \textbf{RESCU-I}. \textcolor{green}{Green}: ours \textbf{RESCU-II}.
\textbf{Top Row}: Full \textbf{RP-1K} dataset.
\textbf{Second Row to Bottom Row:} Results broken out by subset categories: Front View, Projective View, Man-Made Rigid, Man-Made Deformable, Painting, Animal/Human subsets.
\textbf{Row 3, 6 (b)}: the baseline \cite{liu2013grasp} cannot detect the same RP as RESCU does.
}
\label{fig:eval_curves}
\end{figure}

To study the effect of different values of IOD threshold $h$, Figure \ref{fig:eval_curves} shows the performance metrics plotted with respect to increasing values of IOD threshold $h$. We observe that the trends of both RP level and RP-Instance level metrics  monotonically decrease with increasing $h$. Moreover, the RP recall rate and Instance precision \& recall rates are mostly flat when $h$ changes. This indicates that most metrics are not losing correctness when altering $h$ value. These results further support the conclusion that our \textbf{RESCU} method performs quantitatively better than the \textbf{baseline}.

\subsection{Implementation Details}
During training we utilize the Adam \cite{kingma2014adam} optimizer with Cross-Entropy loss. We use $1e^{-4}$ as our learning rate with a $0.5$ decay rate. We train for a maximum of 20 epochs using a batch size of 32 and apply early stopping if validation loss does not decrease by $5e^{-4}$ over three iterations.

\subsection{Demo Video}
\begin{figure}
    \centering
    \begin{subfigure}[t]{0.18\linewidth}
    \includegraphics[width = \textwidth,valign=t]{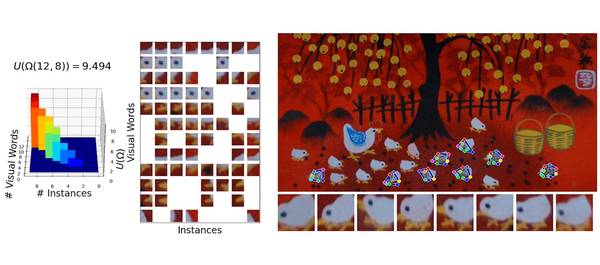}
    \caption{\small  Example 1, RESCU-I, RP 1}
    \end{subfigure}
    \hfill
    \begin{subfigure}[t]{0.18\linewidth}
    \includegraphics[width = \textwidth,valign=t]{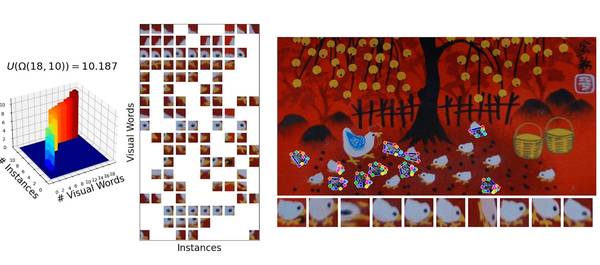}
    \caption{\small  Example 1, RESCU-I, RP 2}
    \end{subfigure}
    \hfill
    \begin{subfigure}[t]{0.18\linewidth}
    \includegraphics[width = \textwidth,valign=t]{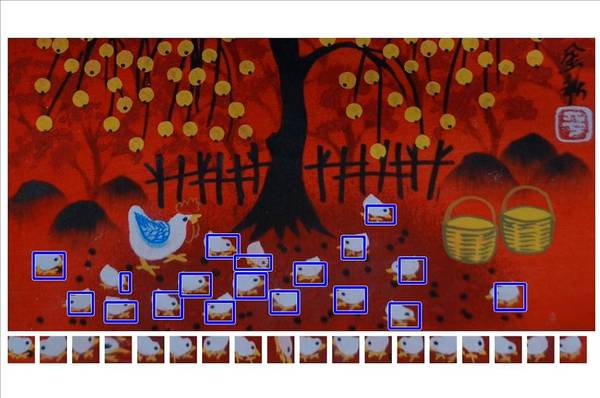}
    \caption{\small  Example 1, RESCU-II}
    \end{subfigure}
    \hfill
    \begin{subfigure}[t]{0.18\linewidth}
    \includegraphics[width = \textwidth,valign=t]{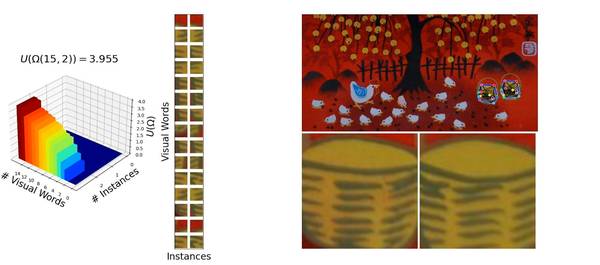}
    \caption{\small  Example 1, RESCU-I, RP 3}
    \end{subfigure}
    \hfill
    \begin{subfigure}[t]{0.18\linewidth}
    \includegraphics[width = \textwidth,valign=t]{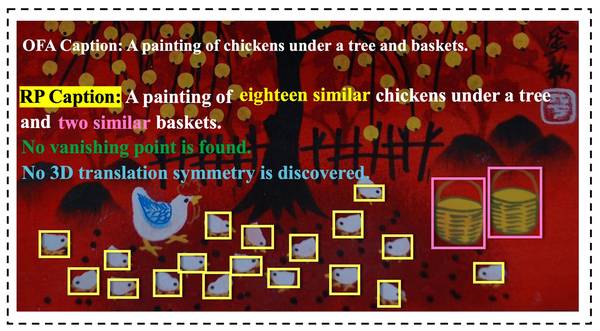}
    \caption{\small  Example 1, Enhanced Caption}
    \end{subfigure}
    \vfill
    \begin{subfigure}[t]{0.18\linewidth}
    \includegraphics[width = \textwidth,valign=t]{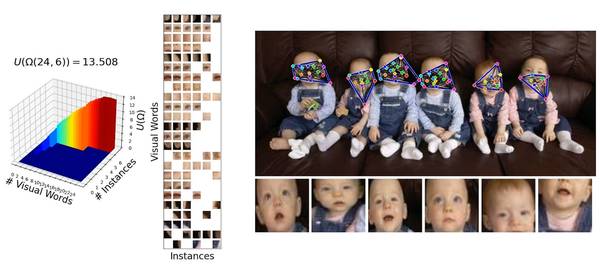}
    \caption{\small  Example 2, RESCU}
    \end{subfigure}
    \hfill
    \begin{subfigure}[t]{0.18\linewidth}
    \includegraphics[width = \textwidth,valign=t]{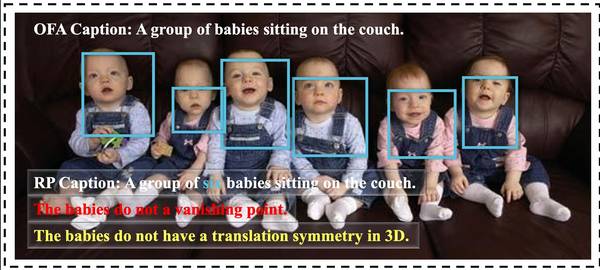}
    \caption{\small  Example 2, Enhanced Caption}
    \end{subfigure}
    \hfill
    \begin{subfigure}[t]{0.18\linewidth}
    \includegraphics[width = \textwidth,valign=t]{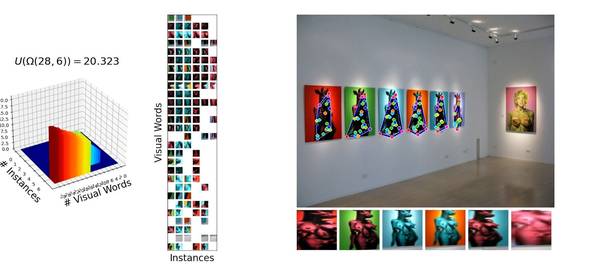}
    \caption{\small  Example 3, RESCU}
    \end{subfigure}
    \hfill
    \begin{subfigure}[t]{0.18\linewidth}
    \includegraphics[width = \textwidth,valign=t]{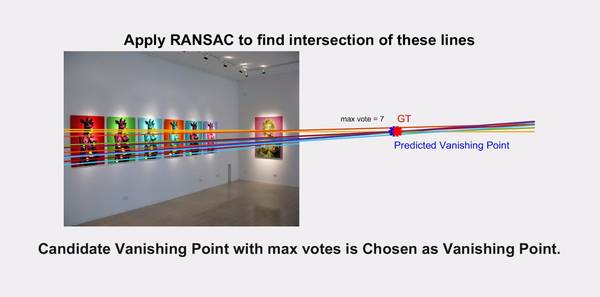}
    \caption{\small  Example 3, Vanishing Point Detection}
    \end{subfigure}
    \hfill
    \begin{subfigure}[t]{0.18\linewidth}
    \includegraphics[width = \textwidth,valign=t]{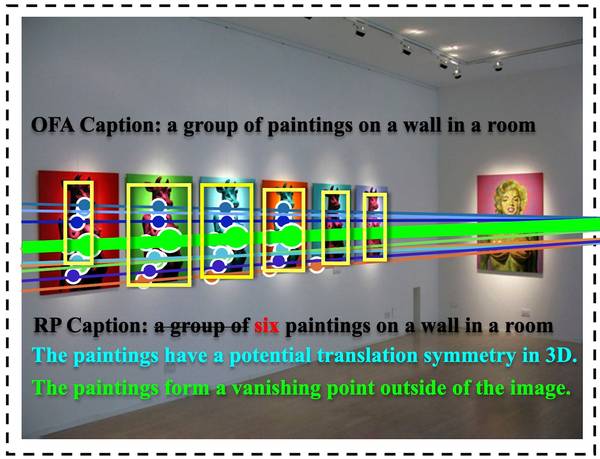}
    \caption{\small  Example 3, Enhanced Caption}
    \end{subfigure}
    \vfill
    \begin{subfigure}[t]{0.18\linewidth}
    \includegraphics[width = \textwidth,valign=t]{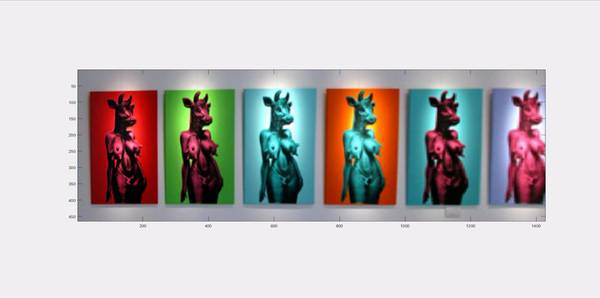}
    \caption{\small  Example 3, Rectification}
    \end{subfigure}
    \hfill
    \begin{subfigure}[t]{0.18\linewidth}
    \includegraphics[width = \textwidth,valign=t]{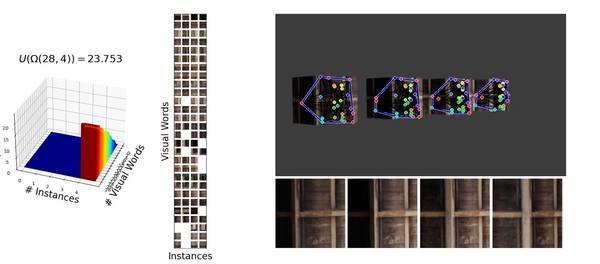}
    \caption{\small  Example 4, RESCU}
    \end{subfigure}
    \hfill
    \begin{subfigure}[t]{0.18\linewidth}
    \includegraphics[width = \textwidth,valign=t]{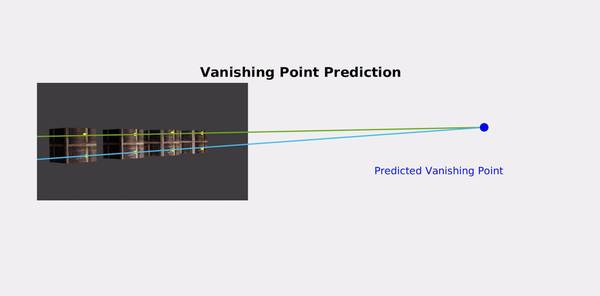}
    \caption{\small  Example 4, Vanishing Point Detection}
    \end{subfigure}
    \hfill
    \begin{subfigure}[t]{0.18\linewidth}
    \includegraphics[width = \textwidth,valign=t]{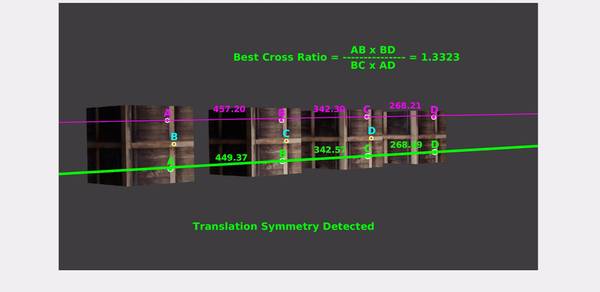}
    \caption{\small  Example 4, Translation Symmetry Detection}
    \end{subfigure}
    \hfill
    \begin{subfigure}[t]{0.18\linewidth}
    \includegraphics[width = \textwidth,valign=t]{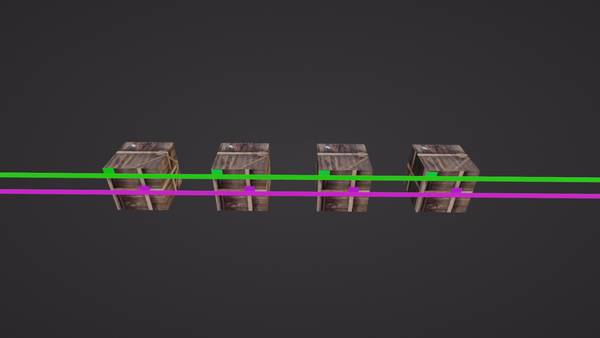}
    \caption{\small  Example 4, 3D Scene Visualization}
    \end{subfigure}
    
    \caption{\small Demo video representative frames}
    \label{fig:demo_frames}
\end{figure}

Fig.~\ref{fig:demo_frames} shows the demo video representative frames of four showcases, including results of RESCU-I, RESCU-II, vanishing point detection, translation symmetry detection, enhanced image caption, image rectification, and 3D scene visualization based on RP discovery.

\end{document}


\newcommand{\COMMENT}[1]{}
\definecolor{aqua}{rgb}{0.0, 1.0, 1.0}

\pagestyle{headings}
\mainmatter

\def\ACCV22SubNumber{756}  

\title{Supplementary Material:\\From Recurrence in a Single Image to \\3D Scene Understanding} 
\titlerunning{ACCV-22 submission ID \ACCV22SubNumber}
\authorrunning{ACCV-22 submission ID \ACCV22SubNumber}

\author{Anonymous ACCV 2022 submission}
\institute{Paper ID \ACCV22SubNumber}

\maketitle
\section{Methodology}
\subsection{RESCU Stage-I}

\subsubsection{Affinity Score Definition}
\label{sec:affinity_definition}
\begin{figure}
    \vspace{-5pt}
    \begin{subfigure}[t]{0.49\linewidth}
    \centering
        \includegraphics[width=\textwidth]{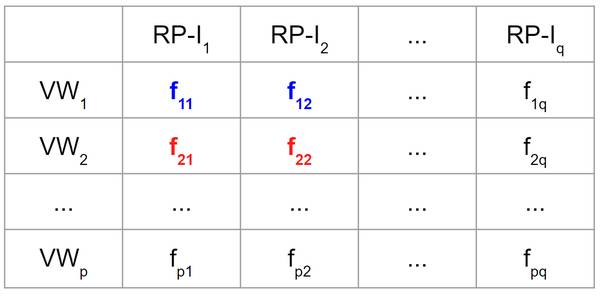}
         \vspace{-15pt}
         \caption{\scriptsize{an RP Matrix} \label{fig:rp_matrix}}
    \end{subfigure}
    \hfill    
    \begin{subfigure}[t]{0.49\linewidth}
    \centering
        \includegraphics[width=\textwidth]{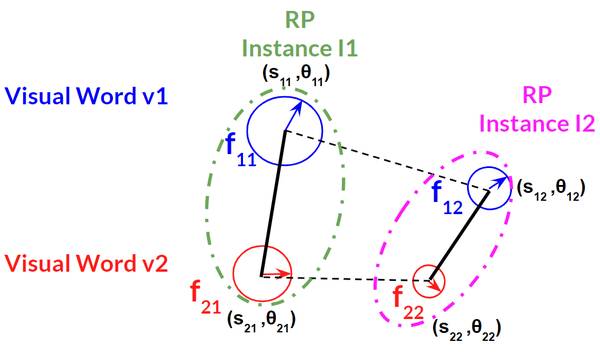}
         \vspace{-15pt}
         \caption{\scriptsize{ Unit Recurring Pattern (URP)} \label{fig:basic_rp_structure}}
    \end{subfigure}
    \vspace{-5pt}
    \caption{ \small Same as Fig.~3 of the main paper.
    \textbf{(a)} an RP matrix, rows are features belonging to the same visual word, and columns are features belonging to the same RP instance.
    \textbf{(b)} a Unit Recurring Pattern (URP) is formed from the RP matrix by choosing two visual words (\textit{rows}) and two RP instances (\textit{columns}) to select four visual primitives, e.g.~local SIFT features, $(f_{11},f_{12},f_{21},f_{22})$, each of which has a scale $s$ and angle $\theta$. 
    }
    \vspace{-5pt}
\end{figure}

As mentioned in Sec.~3.1 of the paper, affinity score $u$ of a Unit Recurring Patter (URP) $\mathbf{\Omega}_{2,2}$ is similarly defined as in \cite{liu2013grasp}.
A URP is formed2x2 sub-matrix $\mathbf{\Omega}_{2,2}$ , formed by choosing 2 rows $m_1, m_2$ and 2 columns $n_1, n_2$, as Fig.~\ref{fig:basic_rp_structure} (also Fig.~3b of the main paper) shows.

The affinity score $u$ is defined as follows:
\begin{align}
    u(m_1,m_2,n_1,n_2|\mathbf{\Omega}) = u(f_{11},f_{12},f_{21},f_{22}) = \exp (-\frac{\Delta_s^2}{2\sigma_s} - \frac{\Delta_{\theta}^2}{2\sigma_{\theta}})
\end{align}
where $\Delta_{s}$ and $\Delta_{\theta}$ measure normalized scale and angular differences and $\sigma_{s}$ and $\sigma_{\theta}$ are deformation tolerance parameters both set to 0.2. 

Here we provide the details of $\Delta_{s}$ and $\Delta_{\theta}$:

Follow \cite{liu2013grasp}, we first define URP instance size ratio $r = \frac{d(f_{11} - f_{21})}{d(f_{12} - f_{22})}$, where $d(\cdot)$ is the Euclidean distance between two features of an URP instance.

We then define $D_s(f_i, f_j)$ as the normalized scale difference by RPI size ratio $r$ of two SIFT features $f_i$ and $f_j$, as follows:
\begin{align}
    D_s(f_i, f_j) &= \frac{s_i - r\cdot s_j}{s_i + r\cdot s_j}
\end{align}
where $s_k$ is the scale of feature $f_k$.

Similarly, we define $D_\theta(f_i, f_j)$ as the normalized angle difference by RPI size ratio $r$ of two SIFT features $f_i$ and $f_j$, as follows:
\begin{align}
    D_\theta(f_i, f_j) &= \frac{{\theta}_1 - r\cdot {\theta}_2}{{\theta}_1 + r\cdot {\theta}_2}
\end{align}

Finally, we define $\Delta_{s}$ as the largest normalized scale difference:
\begin{align}
    \Delta_{s} = \max(D_s(f_{11}, f_{12}), D_s(f_{21}, f_{22}))
\end{align}

We define $\Delta_{\theta}$ as the largest normalized angle difference:
\begin{align}
    \Delta_{\theta} = \max(D_{\theta}(f_{11}, f_{12}), D_{\theta}(f_{21}, f_{22}))
\end{align}

\subsubsection{Unit RP-based Search}
In each detection iteration, our approach maintains an RP matrix for each initial as Figure \ref{fig:rp_matrix} shows. The algorithm expend the URP (as a $2\times2$ initial matrix) with two movement directions: (1) add/remove a column for more/less instances, (2) add/remove a row for more/less visual words.

Without losing generalizability, consider the current RP as $\Omega_{p,q}$ with $p$ visual words, $q$ instances as Fig~\ref{fig:rp_matrix} shows.

To add a column for one more instance is to add a series of features $\{f_{1,q+1}, \allowbreak f_{2,q+1}, \cdot f_{p,q+1}\}$ into column $(q+1)$, to form RP $\Omega_{p,q+1}$.
To find the candidate feature $f_{i,q+1}$ to be put to visual word $i$, we define the visual word affinity sum-up score of $f_{i,q+1}$ respect to $\Omega_{p,q+1}$, as follows: 
\begin{align}
    \mathbf{V}(f_{i,q+1}|\Omega_{p,q+1}) &= \sum_{j\in [1,q]} V(f_{i,q+1}, f_{i,j})\\
    V(f_{i,q+1}, f_{i,j}) &= \sum_{\substack{k\in [1,p],\\k\neq i}} u(f_{i,q+1}, f_{i,j}, f_{k,q+1}, f_{k,j})\\
\end{align}
hence,
\begin{align}
    \mathbf{V}(f_{i,q+1}|\Omega_{p,q+1}) = \sum_{j\in [1,q]} \sum_{\substack{k\in [1,p],\\k\neq i}} u(f_{i,q+1}, f_{i,j}, f_{k,q+1}, f_{k,j})
\end{align}
$\mathbf{V}(f_{i,q+1}|\Omega_{p,q+1})$ measures the overall affinity of URPs containing $f_{i,q+1}$ in $\Omega_{p,q+1}$. A candidate feature $f_{i,q+1}$ that maximize $\mathbf{V}(f_{i,q+1} |\Omega_{p,q+1})$ will be added to row $i$, column $q+1$ of RP $\Omega_{p,q+1}$.

\subsection{RESCU Stage-II}
\begin{figure}[tb!]
    \centering
    \begin{subfigure}[t]{0.22\linewidth}
    \caption{Original Image}
    \includegraphics[width = \textwidth,valign=t]{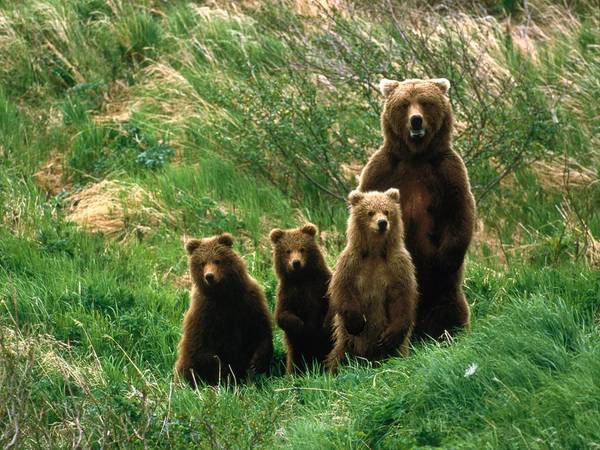}
    \end{subfigure}
    \hfill
    \begin{subfigure}[t]{0.22\linewidth}
    \caption{RESCU-I}
    \includegraphics[width = \textwidth,valign=t]{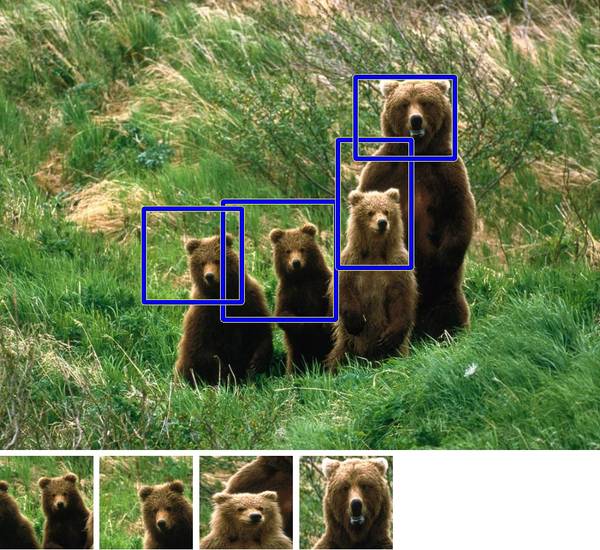}
    \end{subfigure}
    \hfill
    \begin{subfigure}[t]{0.22\linewidth}
    \caption{Positive}
    \includegraphics[width = \textwidth,valign=t]{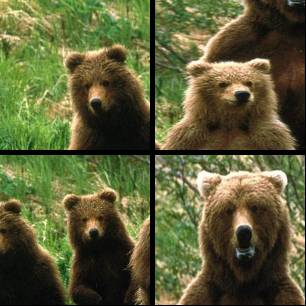}
    \end{subfigure}
    \hfill
    \begin{subfigure}[t]{0.22\linewidth}
    \caption{Negative}
    \includegraphics[width = \textwidth,valign=t]{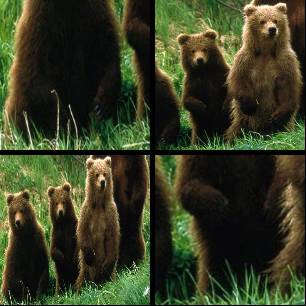}
    \end{subfigure}
    \vfill
    \begin{subfigure}[t]{0.22\linewidth}
    \includegraphics[width = \textwidth,valign=t]{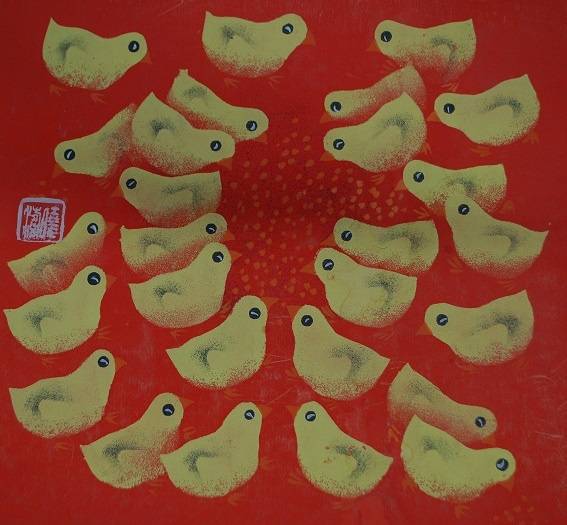}
    \end{subfigure}
    \hfill
    \begin{subfigure}[t]{0.22\linewidth}
    \includegraphics[width = \textwidth,valign=t]{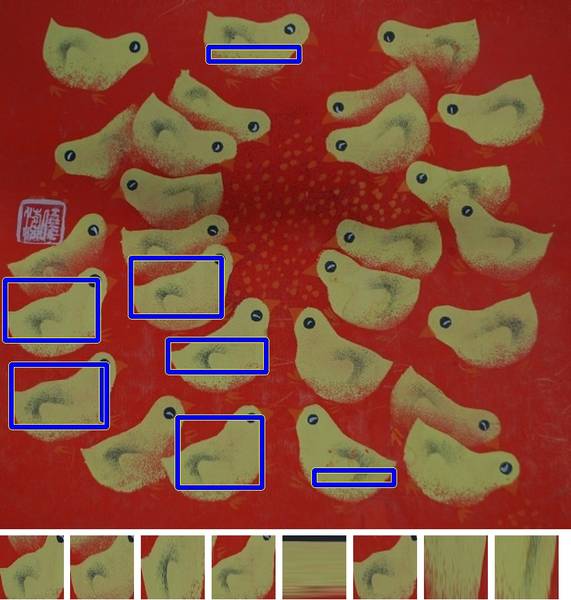}
    \end{subfigure}
    \hfill
    \begin{subfigure}[t]{0.22\linewidth}
    \includegraphics[width = \textwidth,valign=t]{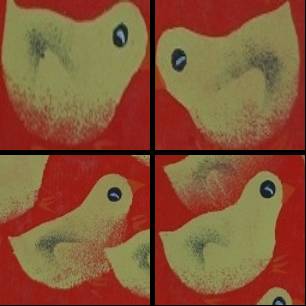}
    \end{subfigure}
    \hfill
    \begin{subfigure}[t]{0.22\linewidth}
    \includegraphics[width = \textwidth,valign=t]{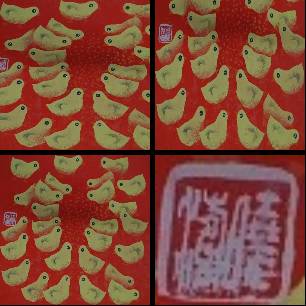}
    \end{subfigure}
    
    \caption{RESCU-II positive \& negative samples of some images. 
    \textbf{(a)} Original Image.
    \textbf{(b)} RESCU-I output.
    \textbf{(c)} Positive samples generated by RESCU-II.
    \textbf{(d)} Negative samples generated by RESCU-II.
    }
    \label{fig:pos_neg}
\end{figure}

This (Sec.~3.2 of the paper) is a self-supervised learning stage, where positive samples are composed of RP instances detected by stage-I and proposed regions with feature closer to the cluster centers of RP instances.
In practice, we select the top $k==N\times P$ nearest proposals to the cluster centers of RP instances, where $N$ is the number of cluster centers and $P$ is an empirical parameter, set to $40$. 
By this procedure, we maintain $<10\%$ of the total proposals generated by the off-the-shelf \cite{zitnick2014edge}. 
Fig.~\ref{fig:pos_neg} shows some example of generated positive \& negative samples.

\begin{figure}[tb!]
    \centering
    \includegraphics[width=\textwidth]{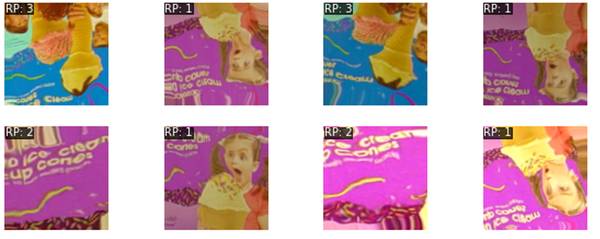}
    \caption{RESCU-II local Deformation Showcase. Each sample is applied with various deformation factors.}
    \label{fig:local_deformation}
\end{figure}

We apply data augmentations to enlarge positive samples for a better training. 
Besides the general global transformations applied for augmentation, we also apply local appearance \& geometric deformation based on corresponding features of RP instances, as Fig.~\ref{fig:local_deformation} shows.

The training parameters in our experiments are set as follows:
\begin{enumerate}
    \item Loss function: Cross-Entropy Loss
    \item Batch Size: 32
    \item Optimizer: Adam
    \item Learning Rate: 1$e^{-4}$
    \item Decay Rate: ($beta\_1$) = 0.5
    \item Maximum Number of Epochs: 20
    \item Validation Split: 20\%
    \item Early stopping if validation loss does not reduce by 0.0005 in 3 iterations.
\end{enumerate}

\clearpage
\section{Experiments \& Results}
\subsection{Study of Quantitative Validation}
As mentioned in Sec.~4.2 of the paper, here we provide more details about the quantitative validation.
For simplicity, we denote \textbf{RESCU-I} to be RESCU with only stage-I, and \textbf{RESCU-II} to be RESCU with both stage-I and stage-II.
Tab.~\ref{tab:rp_eval} shows the evaluation on the full \textbf{RP-1K} dataset as well as broken down into its various viewpoint and category subsets. 
Some RP detection examples from each subset category are shown in Fig.~\ref{fig:enhanced_ex}. All examples shown have \textbf{RESCU-II} performs better or as good as \textbf{RESCU-I}, and better than the baseline method \cite{liu2013grasp}.

The results show that our RESCU performs the best on Animal/Human and Painting subsets. 
Performance on the
Front view and Man-made rigid subsets, comprising the largest part of \textbf{RP-1K} (808 of 1024 images and 813 of 1024 images), is mostly consistent with the whole dataset. 
The results also reveal that our RESCU does not perform as well on the Projective view and Man-made deformable subsets. Images from these subsets are usually more challenging for detecting RPs due to severe geometric distortion, non-uniform lighting, and blurring. 

To study the effect of different values of IOD threshold $h$, Figure \ref{fig:eval_curves} shows the performance metrics plotted with respect to increasing values of IOD threshold $h$. We observe that the trends of both RP level and RP-Instance level metrics  monotonically decrease with increasing $h$. Moreover, the RP recall rate and Instance precision \& recall rates are mostly flat when $h$ changes. This indicates that most metrics are not losing correctness when altering $h$ value. These results further support the conclusion that our \textbf{URPD} method performs quantitatively better than the \textbf{baseline}.

\begin{table}[tb!]
\caption{\small RP Discovery Evaluation on \textbf{RP-1K}.
The values in \textit{italic} are not statistically significant with range in the same column section.
The values in \textbf{bold} are the best mean/std with range in the same column section.
}
\label{tab:rp_eval}
\centering

\resizebox{\linewidth}{!}
{
\begin{tabular}{l|cc|cc} 
\toprule
 & \multicolumn{2}{c|}{RP Level} & \multicolumn{2}{c}{RP Instance Level} \\
Method &     Precision &        Recall &   Precision &      Recall \\
\midrule
\multicolumn{5}{c}{\textbf{The Whole RP-1K}}\\
\hline
\textbf{Baseline\cite{liu2013grasp}} &           0.40 $\pm$\textbf{ 0.37} &           0.32 $\pm$ 0.35 &           0.67 $\pm$ 0.41 &           0.47 $\pm$ 0.38 \\
\textbf{RESCU-I} & 0.44 $\pm$ \textbf{0.32} & \textit{0.50 $\pm$ 0.38} &  0.68 $\pm$ 0.33 &   0.62 $\pm$ 0.34\\
\textbf{RESCU-II} &  \textbf{0.45} $\pm$ \textbf{0.32} &  \textbf{0.52} $\pm$ 0.38 &  \textbf{0.71} $\pm$ \textbf{0.32} &  \textbf{0.64} $\pm$ \textbf{0.33} \\

\hline
\midrule
\multicolumn{5}{c}{\textbf{Front View Subset}}\\
\hline
\textbf{Baseline\cite{liu2013grasp}} &           0.45 $\pm$ 0.36 &           0.34 $\pm$ 0.34 &  \textbf{0.73} $\pm$ 0.37 &           0.51 $\pm$ 0.36 \\
\textbf{RESCU-I  } &   0.47 $\pm$ \textbf{0.30} & 0.51 $\pm$ \textbf{0.35} &  0.70 $\pm$ 0.31 &  \textbf{ 0.65} $\pm$ \textbf{0.32} \\
\textbf{RESCU-II } &  \textbf{0.48} $\pm$ 0.31 &  \textbf{0.52} $\pm$ \textbf{0.35} &           \textbf{0.73} $\pm$ \textbf{0.30} & \textbf{0.65} $\pm$ \textbf{0.32} \\

\hline
\midrule
\multicolumn{5}{c}{\textbf{Projective View Subset}}\\
\hline
\textbf{Baseline\cite{liu2013grasp}} &           0.23 $\pm$ 0.36 &           0.22 $\pm$ \textbf{0.37} &           0.36 $\pm$ 0.45 &           0.27 $\pm$ 0.38 \\
\textbf{RESCU-I  } &           0.32 $\pm$ \textbf{0.34} &           \textbf{0.47} $\pm$ 0.45 &           0.58 $\pm$ 0.40 &           0.44 $\pm$ 0.37 \\
\textbf{RESCU-II } &  \textbf{0.33} $\pm$ 0.35 &  \textbf{0.47} $\pm$ 0.45 &  \textbf{0.63 $\pm$ 0.38} &  \textbf{0.49 $\pm$ 0.37} \\

\hline
\midrule
\multicolumn{5}{c}{\textbf{Man-Made Rigid Subset}}\\
\hline
\textbf{Baseline\cite{liu2013grasp}} &           0.42 $\pm$ 0.37 &           0.32 $\pm$ \textbf{0.35} &           0.69 $\pm$ 0.40 &           0.49 $\pm$ 0.37 \\
\textbf{RESCU-I  } &           0.44 $\pm$ \textbf{0.32} &          \textbf{ 0.50} $\pm$ 0.37 &           0.69 $\pm$ 0.33 &           0.62 $\pm$ \textbf{0.34} \\
\textbf{RESCU-II } &  \textbf{0.45} $\pm$ \textbf{0.32} &  \textbf{0.51} $\pm$ 0.37 &  \textbf{0.72} $\pm$ \textbf{0.32} &  \textbf{0.63} $\pm$ \textbf{0.34} \\

\hline
\midrule
\multicolumn{5}{c}{\textbf{Man-Made Deformable Subset}}\\
\hline
\textbf{Baseline\cite{liu2013grasp}} &           0.33 $\pm$ 0.35 &           0.21 $\pm$ \textbf{0.26} &           0.58 $\pm$ 0.43 &           0.43 $\pm$ 0.37 \\
\textbf{RESCU-I  } &           0.38 $\pm$ \textbf{0.28} &           0.41 $\pm$ 0.33 &           0.63 $\pm$ \textbf{0.31} &           \textbf{0.63} $\pm$ \textbf{0.30} \\
\textbf{RESCU-II } &  \textbf{0.40} $\pm$ 0.29 &  \textbf{0.44} $\pm$ 0.33 &  \textbf{0.66} $\pm$ \textbf{0.31} &  \textbf{0.63} $\pm$ \textbf{0.30} \\

\hline
\midrule
\multicolumn{5}{c}{\textbf{Painting Subset}}\\
\hline
\textbf{Baseline\cite{liu2013grasp}} &           0.45 $\pm$ 0.40 &           0.61 $\pm$ 0.48 &           0.64 $\pm$ 0.46 &           0.34 $\pm$ 0.34 \\
\textbf{RESCU-I  } &           0.73 $\pm$ \textbf{0.30} &  \textit{\textbf{0.89} $\pm$ \textbf{0.29}} &           \textbf{0.82} $\pm$ \textbf{0.24} &  \textbf{0.60} $\pm$ \textbf{0.31} \\
\textbf{RESCU-II } &  \textbf{0.74} $\pm$ \textbf{0.30} &           \textit{\textbf{0.89} $\pm$ \textbf{0.29}} &  \textbf{0.82} $\pm$ 0.25 &           \textbf{0.60} $\pm$ \textbf{0.31} \\

\hline
\midrule
\multicolumn{5}{c}{\textbf{Animal/Human Subset}}\\
\hline
\textbf{Baseline\cite{liu2013grasp}} &           0.45 $\pm$ 0.47 &           0.52 $\pm$ 0.50 &           0.39 $\pm$ 0.48 &           0.25 $\pm$ 0.35 \\
\textbf{RESCU-I  } &           0.57 $\pm$ \textbf{0.37} &           0.78 $\pm$ 0.42 &           0.76 $\pm$ 0.36 &           0.59 $\pm$ 0.35 \\
\textbf{RESCU-II } &  \textbf{0.62} $\pm$ \textbf{0.37} &  \textbf{0.81} $\pm$ \textbf{0.39} &  \textbf{0.83} $\pm$ \textbf{0.30} &  \textbf{0.67} $\pm$ \textbf{0.32} \\

\bottomrule
\end{tabular}
}

\end{table}

\begin{figure}[tb!]
\centering
\foreach \id/\subset in {1/All, 2/Front View, 3/Projective View, 4/Man-Made Rigid, 5/Man-Made Deformable, 6/Painting, 7/Animal Human}
{ %
    \foreach \x/\metric in {1/RP_Prec, 2/RP_Rec, 3/Inst_Prec, 4/Inst_Rec}
    { %
    \begin{subfigure}[t]{0.23\linewidth}
        \if \id 1
            \if \x 1 \caption{\scriptsize{RP Precision}} \fi
            \if \x 2 \caption{\scriptsize{RP Recall}} \fi
            \if \x 3 \caption{\scriptsize{Instance Precision}} \fi
            \if \x 4 \caption{\scriptsize{Instance Recall}} \fi
        \fi
        \includegraphics[width = \textwidth,valign=t]{Shimian's Figures/Evaluation Curves/\subset/\metric.jpg} 
    \end{subfigure} 
    \hfill
    }
    \vspace{2pt}
    
} \vspace{-5pt}
\caption{IOD Curves showing each evaluation metric at different IOD thresholds $h$.
\textcolor{blue}{Blue}: the \textbf{baseline} method\cite{liu2013grasp}. \textcolor{red}{Red}: ours \textbf{RESCU-I}. \textcolor{green}{Green}: ours \textbf{RESCU-II}.
\textbf{Top Row}: Full \textbf{RP-1K} dataset.
\textbf{Second Row to Bottom Row:} Results broken out by subset categories: Front View, Projective View, Man-Made Rigid, Man-Made Deformable, Painting, Animal/Human subsets.
\textbf{Row 3, 6 (b)}: the baseline \cite{liu2013grasp} cannot detect the same RP as RESCU does.
}
\label{fig:eval_curves}
\end{figure}

\begin{figure}[tb!]
\centering
\foreach \id/\picname in {1/new295, 2/new764, 3/new997, 4/new498, 5/new014, 6/new017}
{ %
    \foreach \x/\method in {1/GT,2/Baseline,3/URPD-I,4/URPD-II}
    { %
    \begin{subfigure}[t]{0.23\linewidth}
        \if \id 1
            \if \x 1 \caption{Ground Truth} \fi
            \if \x 2 \caption{Baseline \cite{liu2013grasp}} \fi
            \if \x 3 \caption{RESCU-I} \fi
            \if \x 4 \caption{RESCU-II} \fi
        \fi
        \includegraphics[width = \textwidth,valign=t]{Shimian's Figures/Select Results/Subset Collections/RP1K_\picname_\method.jpg} 
    \end{subfigure} 
    \hfill
    }
    \vspace{2pt}
    
} \vspace{-5pt}
\caption{Examples of RP detection results from each camera view/RP category.
\textbf{Top to Down:} Results broken out by subset categories: Front View, Projective View, Man-Made Rigid, Man-Made Deformable, Painting, Animal/Human subsets.}
\label{fig:enhanced_ex}
\end{figure}

\subsection{Ablation Study of RESCU Stage-I}
As mentioned in Sec.~3.1  of the paper, here we study the impact of each adaptive parameter.
\begin{itemize}
    \item $P_d$ sets the maximum feature distance, which controls the granularity of RP detection.
    \item $P_s$ sets the maximum size difference among RP instances.
    \item $P_{\theta}$ sets the maximum orientation difference among RP instances.
\end{itemize}

To study the impact of these adaptive parameters, we design experiments to separately optimize each parameter (and each pair of two parameters), with fixed other parameters. Tab.~\ref{tab:para_settings} shows the fixed value and optimization values of each parameter.

\begin{table}[tb!]

\caption{\small A Summary of Adaptive Parameters used in RESCU Stage-I.}
\label{tab:para_settings}
\centering

\resizebox{\linewidth}{!}
{
\begin{tabular}{l|c|c|c} 
\toprule
Adaptive Parameter & Description &  Fixed Value & Optimized Values\\
\midrule

$P_d$ & the maximum feature distance & 0.2 & $[0.1, 0.15, 0.2]$\\
$P_s$ & the maximum size difference among RP instances & 0.5 & $[0.1, 0.2, 0.3, 0.4, 0.5]$\\
$P_\theta$ & the maximum orientation difference among RP instances & 30 & $[30, 90, 180]$\\
\bottomrule
\end{tabular}
}
\vspace{-15pt}
\end{table}

\begin{figure}[tb!]
    \centering
    \includegraphics[width=0.5\linewidth]{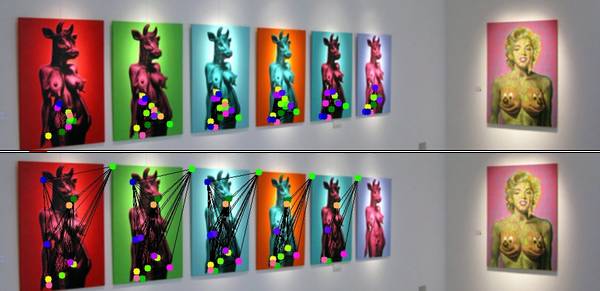}
    \caption{An example of using different $P_d$ in RESCU. 
    \textbf{Up}: with $P_d=0.05$ to detect smaller RP instance, \textbf{Down}: with $P_d=0.25$ to detect larger RP instance.}
    \label{fig:p_d_ex}
\end{figure}

Tab.~\ref{tab:stage_1_ablation} shows the ablation study of adaptive parameters. From the study we can see that $P_d$ parameter influences the performance most.
Fig.~\ref{fig:p_d_ex} shows an example of how changing $P_d$ can control the size of an RP-instance.

\begin{table}[tb!]

\caption{\small Ablation Study of RESCU-I IOD Threshold $h=0.5$.
$P_d$ (maximum RPI feature distance),
$P_s$ (maximum RPI size difference),
$P_{\theta}$ (maximum RPI orientation difference),
See Tab.~\ref{tab:para_settings} for parameter details.
The values in \textit{italic} are not statistically significant with range in the same column.
The values in \textbf{bold} are the best mean/std with range in the same column.
}
\label{tab:stage_1_ablation}
\centering

\begin{tabularx}{\textwidth}{|c|XXX|cc|cc|} 
\toprule
 &  \multicolumn{3}{c|}{Parameter} & \multicolumn{2}{c|}{RP Level} & \multicolumn{2}{c|}{RP Instance Level} \\
Method &  $P_d$ & $P_s$ & $P_\theta$ &   Precision &        Recall &   Precision &      Recall \\
\midrule

\textbf{Baseline\cite{liu2013grasp}} &  & & &         0.40 $\pm$ 0.37 &           0.32 $\pm$ \textbf{0.35} &           0.67 $\pm$ 0.41 &           0.47 $\pm$ 0.38 \\

\textbf{RESCU-I} &  & & &  0.37 $\pm$ \textbf{0.31} &  0.44 $\pm$ 0.38 &  0.70 $\pm$ 0.30 &  0.64 $\pm$ \textbf{0.32} \\

\textbf{RESCU-I} &  \checkmark & & & \textit{\textbf{0.48} $\pm$ 0.33} &  \textit{\textbf{0.53} $\pm$ 0.38} &  0.73 $\pm$ 0.29 & \textit{\textbf{0.67} $\pm$ \textbf{0.32}} \\
\textbf{RESCU-I} & & \checkmark & &  0.43 $\pm$ 0.33 &  0.46 $\pm$ 0.38 &  0.72 $\pm$ 0.30 &  0.65 $\pm$ \textbf{0.32} \\
\textbf{RESCU-I} & & & \checkmark &  0.41 $\pm$ 0.33 &  0.44 $\pm$ 0.38 &  0.67 $\pm$ 0.30 &  0.63 $\pm$ \textbf{0.32} \\

\textbf{RESCU-I} & \checkmark & \checkmark &  & \textit{0.47 $\pm$ 0.33} &  \textit{\textbf{0.53} $\pm$ 0.38} & \textbf{ 0.75} $\pm$ \textbf{0.28} &  \textit{\textbf{0.67} $\pm$ \textbf{0.32}} \\
\textbf{RESCU-I} & \checkmark &  & \checkmark &  \textit{\textbf{0.48} $\pm$ 0.33} &  \textit{\textbf{0.53} $\pm$ 0.38} &  0.72 $\pm$ 0.29 & \textit{ 0.66 $\pm$ \textbf{0.32}} \\
\textbf{RESCU-I} &  & \checkmark & \checkmark & 0.43 $\pm$ 0.33 &  0.45 $\pm$ 0.38 &  0.69 $\pm$ 0.30 &  0.64 $\pm$ \textbf{0.32} \\

\textbf{RESCU-I} & \checkmark & \checkmark & \checkmark &  \textit{0.47 $\pm$ 0.34} &  \textit{0.52 $\pm$ 0.38} &  0.72 $\pm$ 0.29 &  \textit{0.66 $\pm$ \textbf{0.32} }\\
\bottomrule
\end{tabularx}
\vspace{-15pt}
\end{table}


\clearpage
\subsection{Vanishing Point Prediction from A Single View}
\subsubsection{Line Fitting and RANSAC in Vanishing Point Detection}
\begin{figure*}[th]
    \centering
    \includegraphics[width=0.85\linewidth]{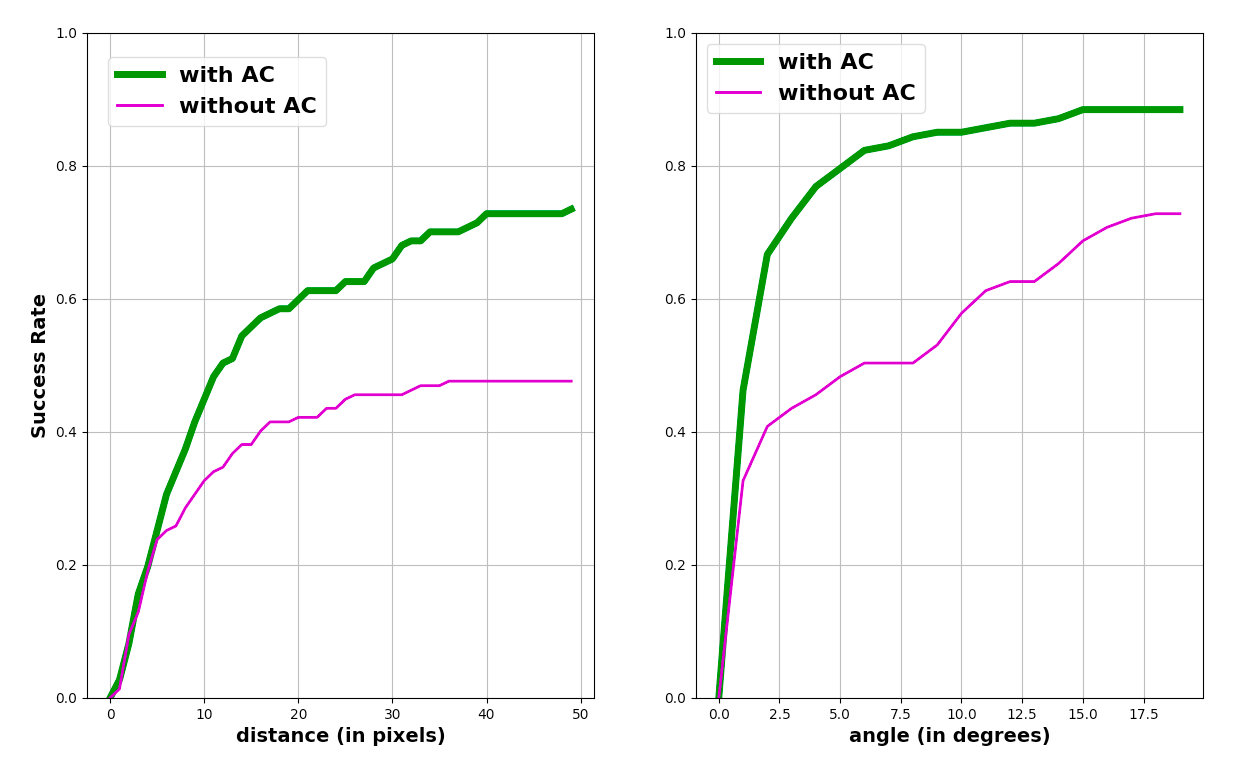}
    \caption{Success rate of vanishing point detection (VPD) with respect to distance (left) and angle (right) difference between predicted and groundtruth VPs. With the introduction of the angular constraint (AC) we see a significant improvement in the performance. The performance of the proposed approach is statistically different with {\it p-value} = 1.13e-06 with the angular constraint.}
    \label{fig:angular_constraint}
\end{figure*}

The intersection of near-parallel lines often results in faulty estimation of the vanishing point. To overcome this ill-conditioning, we introduce an angular constraint (AC) into our algorithm by comparing the angle between lines before selecting them to initialize the RANSAC algorithm. If the angle is smaller (near-parallel lines) than a threshold, we do not consider their contribution towards vanishing point calculation. Using this angular constraint produces a better estimate of the vanishing point. Fig.~\ref{fig:angular_constraint} shows the success rate of our method with and without the Angular Constraint.

\subsubsection{Vanishing Point Detection Evaluation}
\begin{figure*}\vspace{-1pt}
\centering
\begin{subfigure}[t]{0.2375\linewidth}

\caption{\tiny{$A^{\circ}<1^{\circ}$ \& $D<10p$}} 
    \includegraphics[height = \linewidth, width = \linewidth]{Skanda's Figures/VP/GenericOutputs/Best/59.png}
    \includegraphics[height = \linewidth, width = \linewidth]{Skanda's Figures/VP/GenericOutputs/Best/82.png}
    \includegraphics[height = \linewidth, width = \linewidth]{Skanda's Figures/VP/GenericOutputs/Best/141.png}
    \includegraphics[height = \linewidth, width = \linewidth]{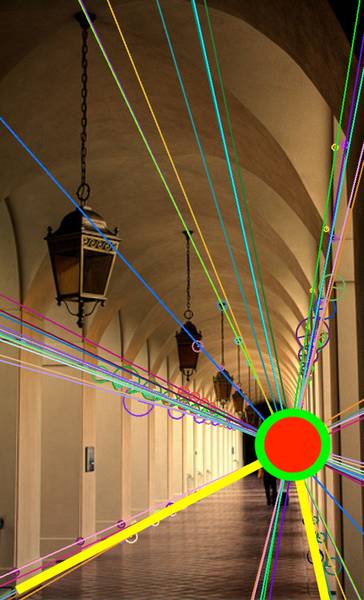}
    \includegraphics[height = \linewidth, width = \linewidth]{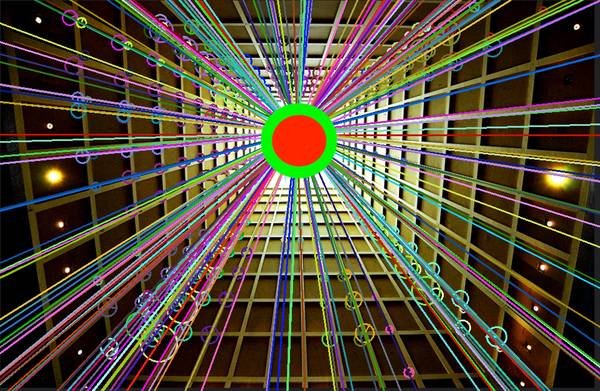}
\end{subfigure}\hfill
\begin{subfigure}[t]{0.2375\linewidth}
    \caption{\tiny{$A^{\circ}<5^{\circ}$ \& $D < 20p$}}
    \includegraphics[height = \linewidth, width = \linewidth]{Skanda's Figures/VP/GenericOutputs/Better/148.png}
    \includegraphics[height = \linewidth, width = \linewidth]{Skanda's Figures/VP/GenericOutputs/Better/252.png}
    \includegraphics[height = \linewidth, width = \linewidth]{Skanda's Figures/VP/GenericOutputs/Better/0638.png}
    \includegraphics[height = \linewidth, width = \linewidth]{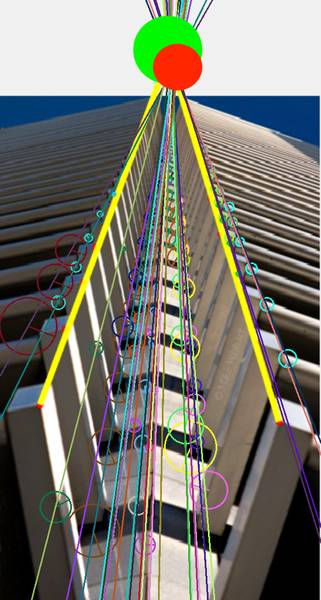}
    \includegraphics[height = \linewidth, width = \linewidth]{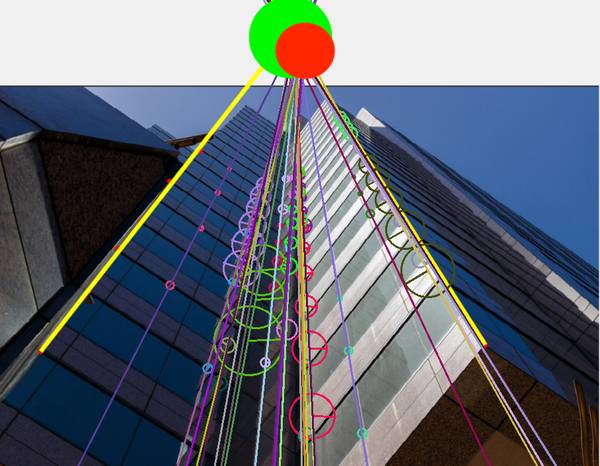}
\end{subfigure}\hfill
\begin{subfigure}[t]{0.2375\linewidth}
    \caption{\tiny{$A^{\circ}>5^{\circ}$ or $D<20p$}}
    \includegraphics[height = \linewidth, width = \linewidth]{Skanda's Figures/VP/GenericOutputs/Worse/1032.png}
    \includegraphics[height = \linewidth, width = \linewidth]{Skanda's Figures/VP/GenericOutputs/Worse/1308.png}
    \includegraphics[height = \linewidth, width = \linewidth]{Skanda's Figures/VP/GenericOutputs/Worse/1678.png}
    \includegraphics[height = \linewidth, width = \linewidth]{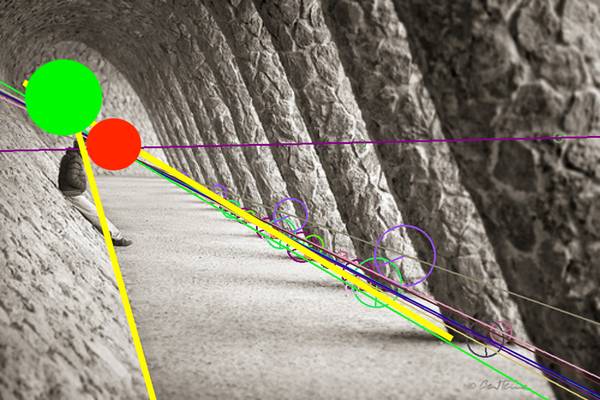}
    \includegraphics[height = \linewidth, width = \linewidth]{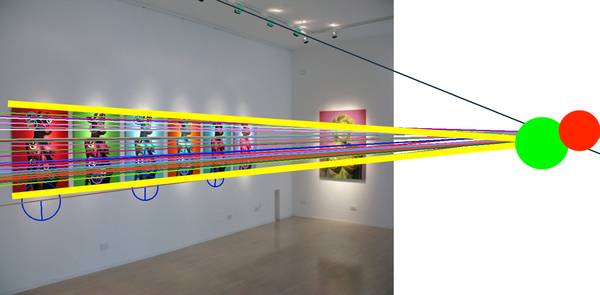}
\end{subfigure}\hfill 
\begin{subfigure}[t]{0.2375\linewidth}
    \caption{\tiny{$A^{\circ}>20^{\circ}$ or $D > 30p$}}
    \includegraphics[height = \linewidth, width = \linewidth]{Skanda's Figures/VP/GenericOutputs/Worst/43.png}
    \includegraphics[height = \textwidth, width = \linewidth]{Skanda's Figures/VP/GenericOutputs/Worst/100.png}
    \includegraphics[height = \linewidth, width = \linewidth]{Skanda's Figures/VP/GenericOutputs/Worst/165.png}
    \includegraphics[height = \linewidth, width = \linewidth]{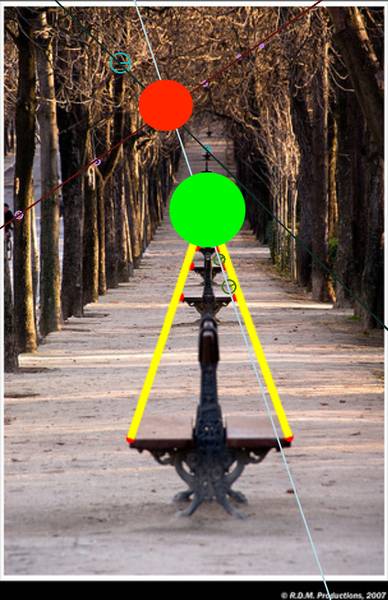}
    \includegraphics[height = \linewidth, width = \linewidth]{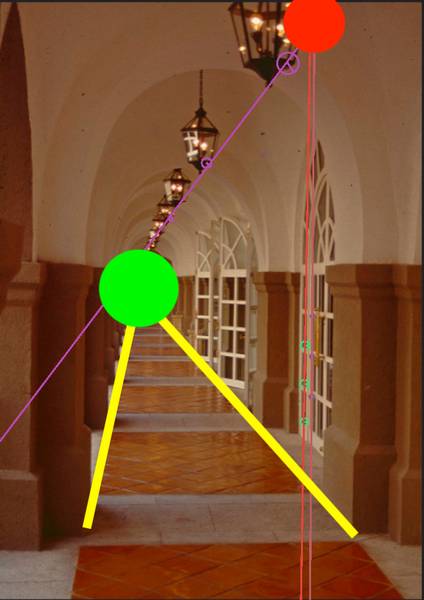}
\end{subfigure}\vspace{-8pt}
\caption{Example vanishing point prediction by our method. $A^{\circ}$ denotes the vector angle between predicted and ground truth VP vectors in degrees. $D$ represents the distance between predicted and ground truth VP image locations in pixels ($p$). \textcolor{red}{Red} dot: our prediction. \colorbox{yellow}{Yellow} lines and \textcolor{green}{Green} dot: two supporting lines intersected at the ground truth vanishing point.  
}
\label{fig:vp_best_worst}
\end{figure*}

As promised in Sec.~5.1 of the paper, here we provide the complete results for Vanishing Point (VP) prediction. Fig. \ref{fig:vp_best_worst} shows prediction results from our method separated in columns based on the prediction quality compared with the ground truth. We use angle and distance with respect to the ground truth to assess the prediction quality.

\begin{figure*}
\centering
\begin{subfigure}[t]{0.198\linewidth}
    \caption{GT}
    \includegraphics[height = \textwidth, width = \textwidth]{Skanda's Figures/VP/AllOuts/GT/1004.png}
    \includegraphics[height = \textwidth, width = \textwidth]{Skanda's Figures/VP/AllOuts/GT/0252.png}
    \includegraphics[height = \textwidth, width = \textwidth]{Skanda's Figures/VP/AllOuts/GT/1008.png}
    \includegraphics[height = \textwidth, width = \textwidth]{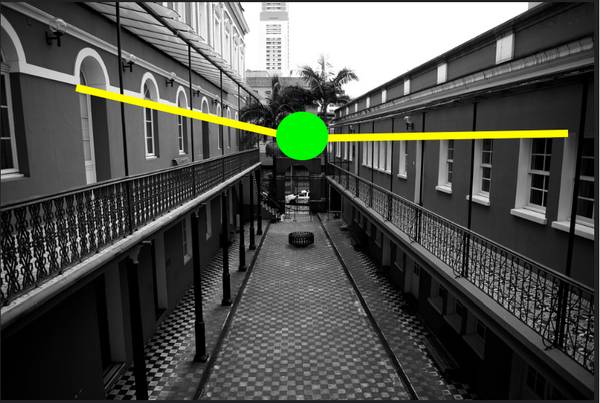}
    \includegraphics[height = \textwidth, width = \textwidth]{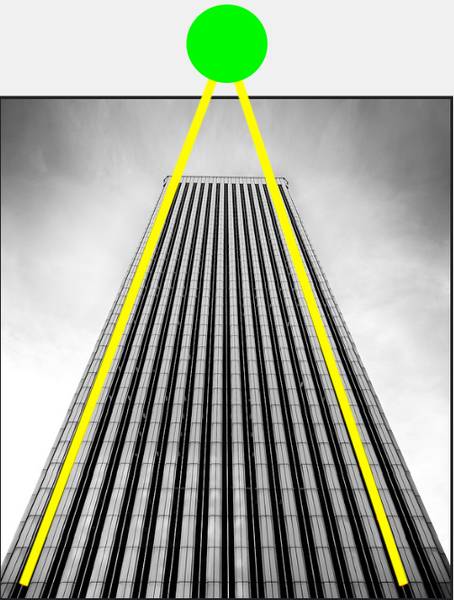}
\end{subfigure}\hfill
\begin{subfigure}[t]{0.198\linewidth}
    \caption{Our's}
    \includegraphics[height = \textwidth, width = \textwidth]{Skanda's Figures/VP/AllOuts/Ours/1004.png}
    \includegraphics[height = \textwidth, width = \textwidth]{Skanda's Figures/VP/AllOuts/Ours/0252.png}
    \includegraphics[height = \textwidth, width = \textwidth]{Skanda's Figures/VP/AllOuts/Ours/1008.png}
    \includegraphics[height = \textwidth, width = \textwidth]{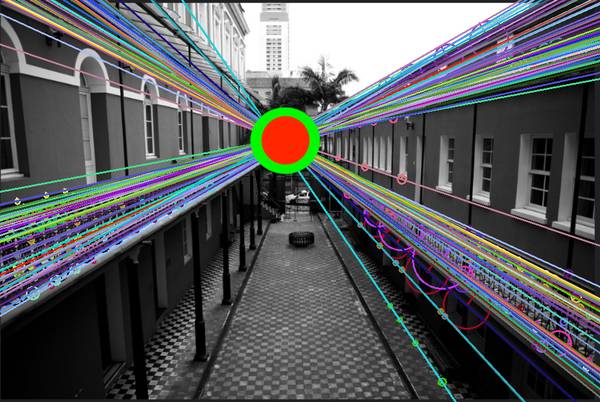}
    \includegraphics[height = \textwidth, width = \textwidth]{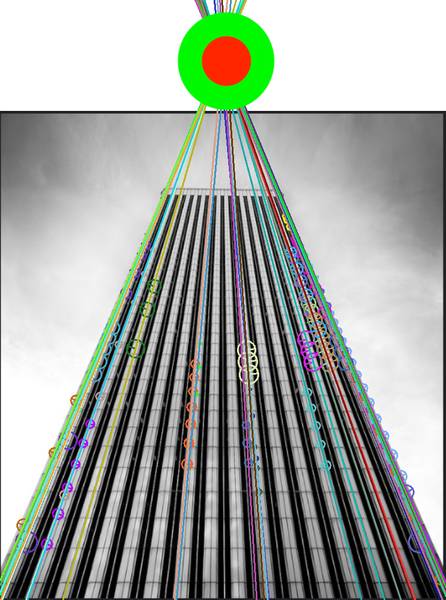}
\end{subfigure}\hfill
\begin{subfigure}[t]{0.198\linewidth}
    \caption{NeurVPS\label{fig:zhou_vp}}
    \includegraphics[height = \textwidth, width = \textwidth]{Skanda's Figures/VP/AllOuts/yima/1004.png}
    \includegraphics[height = \textwidth, width = \textwidth]{Skanda's Figures/VP/AllOuts/yima/0252.png}
    \includegraphics[height = \textwidth, width = \textwidth]{Skanda's Figures/VP/AllOuts/yima/1008.png}
    \includegraphics[height = \textwidth, width = \textwidth]{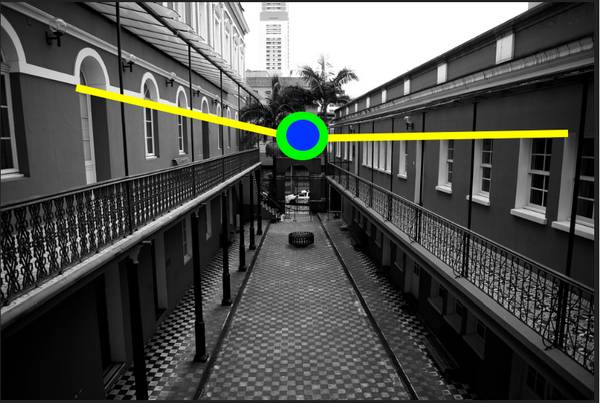}
    \includegraphics[height = \textwidth, width = \textwidth]{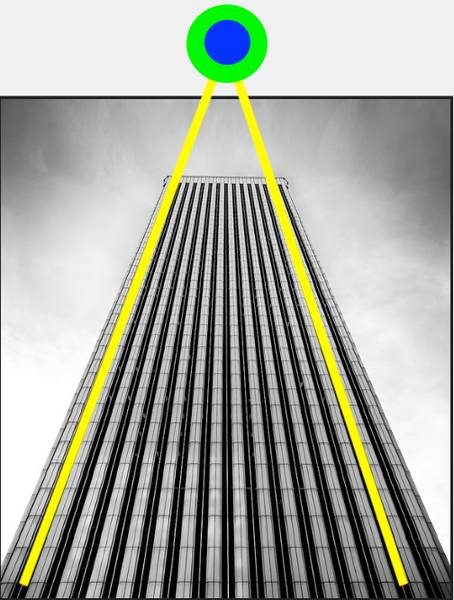}
\end{subfigure}\hfill
\begin{subfigure}[t]{0.198\linewidth}
    \caption{Zhou's\label{fig:zhou_vp_cont2}}
    \includegraphics[height = \textwidth, width = \textwidth]{Skanda's Figures/VP/AllOuts/Zhous/1004.png}
    \includegraphics[height = \textwidth, width = \textwidth]{Skanda's Figures/VP/AllOuts/Zhous/0252.png}
    \includegraphics[height = \textwidth, width = \textwidth]{Skanda's Figures/VP/AllOuts/Zhous/1008.png}
    \includegraphics[height = \textwidth, width = \textwidth]{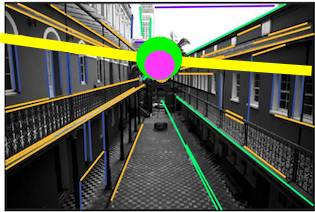}
    \includegraphics[height = \textwidth, width = \textwidth]{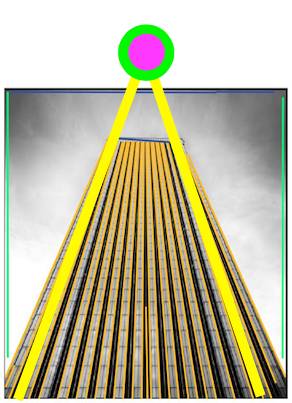}
\end{subfigure}\hfill
\caption{
Sample results for vanishing point (VP) detection using our method and a comparison with \cite{zhou2019neurvps} and \cite{zhou2017detecting}. a) Input image with ground truth, 
b) VP prediction using our method, 
c) VP prediction using NeurVPS\cite{zhou2019neurvps}, 
d) VP prediction using Zhou et.al.'s method \cite{zhou2017detecting}. \textcolor{red}{Red} dot: our prediction. \colorbox{yellow}{Yellow} lines and \textcolor{green}{Green} dot: two supporting lines intersected at the ground truth vanishing point. \textcolor{blue}{Blue} is the output of NeurVPS\cite{zhou2019neurvps} and \textcolor{magenta}{Magenta} is the output of Zhou et.al.'s method\cite{zhou2017detecting}.
}
\label{fig:vp_ex_1}
\end{figure*}

\begin{figure*}
\centering
\begin{subfigure}[t]{0.198\linewidth}
    \caption{GT}
    \includegraphics[height = \textwidth, width = \textwidth]{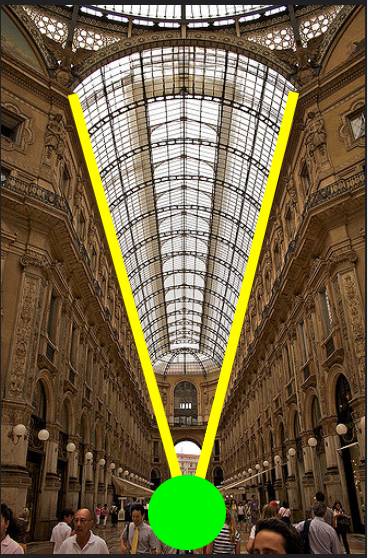}
    \includegraphics[height = \textwidth, width = \textwidth]{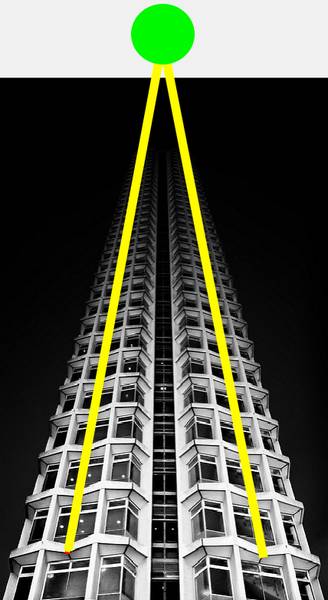}
    \includegraphics[height = \textwidth, width = \textwidth]{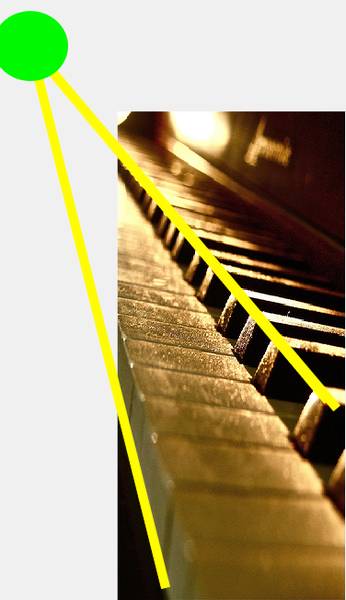}
    \includegraphics[height = \textwidth, width = \textwidth]{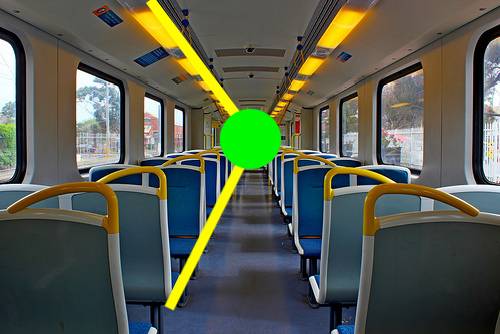}
    \includegraphics[height = \textwidth, width = \textwidth]{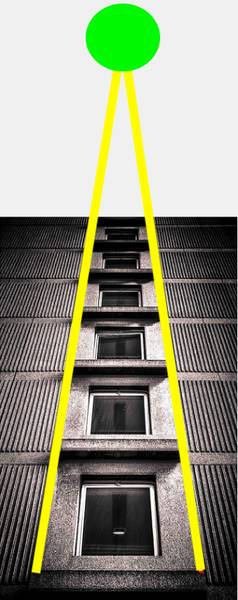}
\end{subfigure}\hfill
\begin{subfigure}[t]{0.198\linewidth}
    \caption{Our's}
    \includegraphics[height = \textwidth, width = \textwidth]{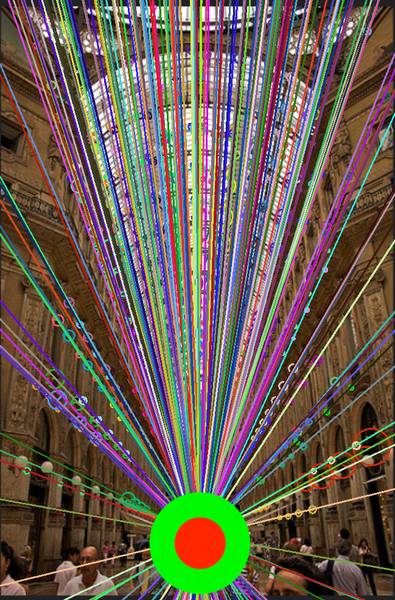}
    \includegraphics[height = \textwidth, width = \textwidth]{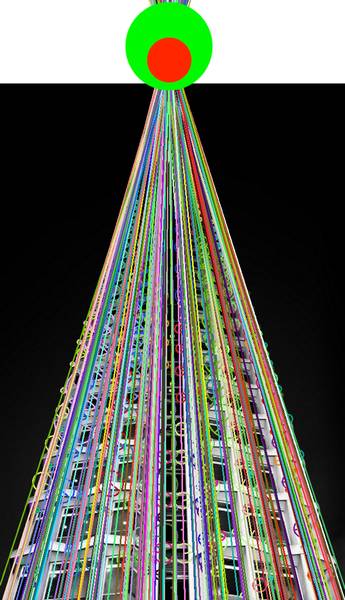}
    \includegraphics[height = \textwidth, width = \textwidth]{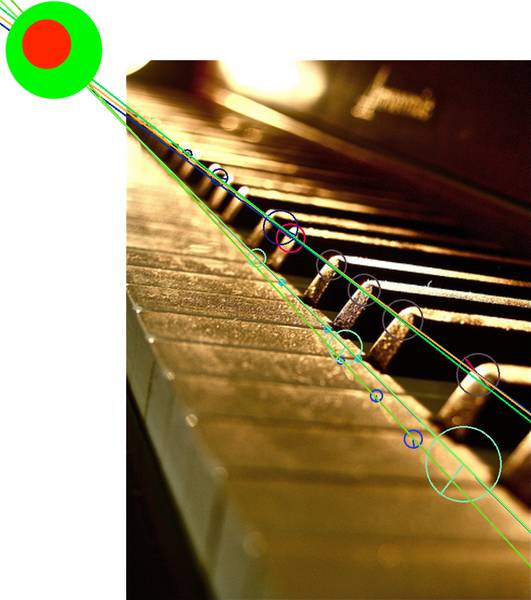}
    \includegraphics[height = \textwidth, width = \textwidth]{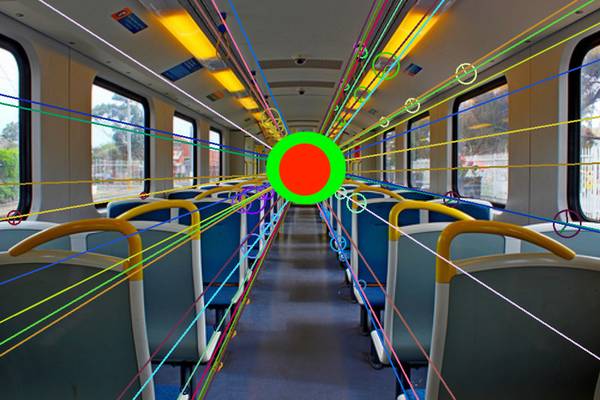}
    \includegraphics[height = \textwidth, width = \textwidth]{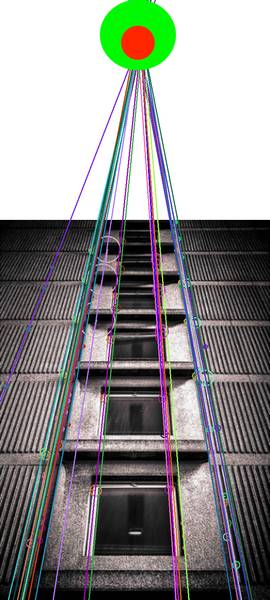}
\end{subfigure}\hfill
\begin{subfigure}[t]{0.198\linewidth}
    \caption{NeurVPS\label{fig:zhou_vp}}
    \includegraphics[height = \textwidth, width = \textwidth]{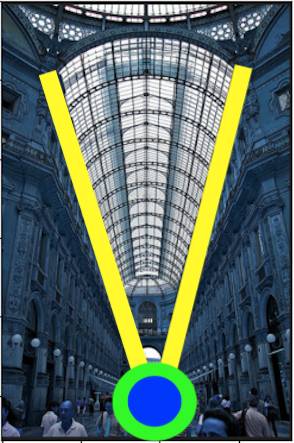}
    \includegraphics[height = \textwidth, width = \textwidth]{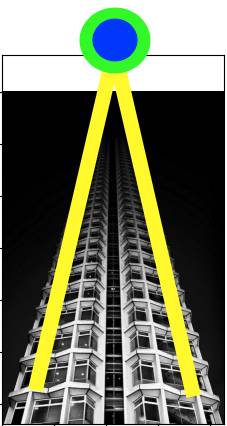}
    \includegraphics[height = \textwidth, width = \textwidth]{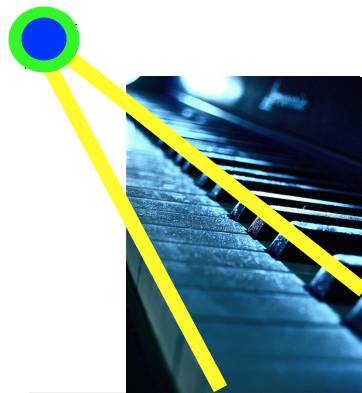}
    \includegraphics[height = \textwidth, width = \textwidth]{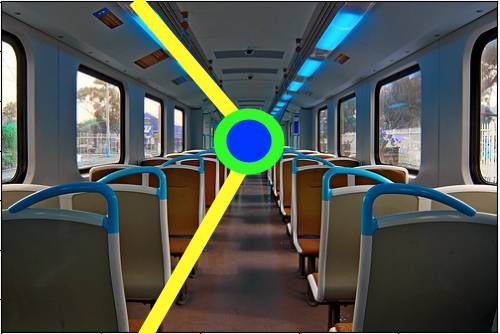}
    \includegraphics[height = \textwidth, width = \textwidth]{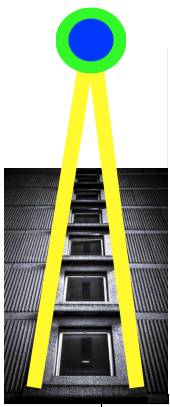}
\end{subfigure}\hfill
\begin{subfigure}[t]{0.198\linewidth}
    \caption{Zhou's\label{fig:zhou_vp_cont2}}
    \includegraphics[height = \textwidth, width = \textwidth]{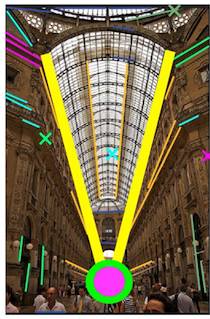}
    \includegraphics[height = \textwidth, width = \textwidth]{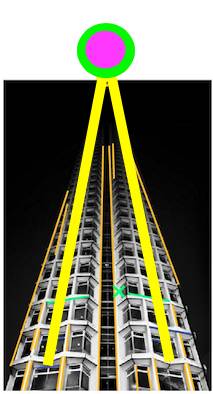}
    \includegraphics[height = \textwidth, width = \textwidth]{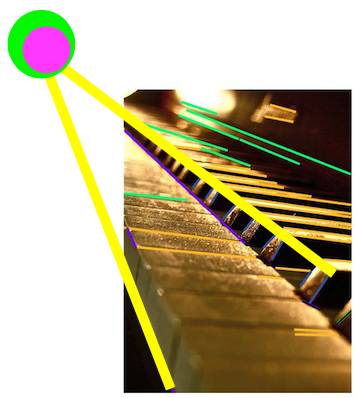}
    \includegraphics[height = \textwidth, width = \textwidth]{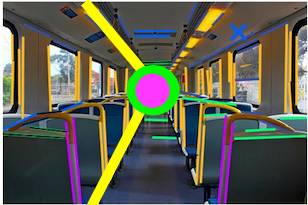}
    \includegraphics[height = \textwidth, width = \textwidth]{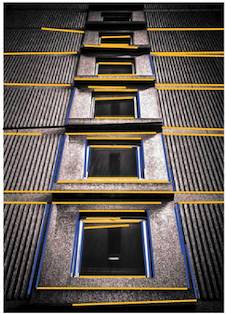}
\end{subfigure}\hfill
\caption{
Sample results for vanishing point (VP) detection using our method and a comparison with \cite{zhou2019neurvps} and \cite{zhou2017detecting}. a) Input image with ground truth, 
b) VP prediction using our method, 
c) VP prediction using NeurVPS\cite{zhou2019neurvps}, 
d) VP prediction using Zhou et.al.'s method \cite{zhou2019neurvps}. \textcolor{red}{Red} dot: our prediction. \colorbox{yellow}{Yellow} lines and \textcolor{green}{Green} dot: two supporting lines intersected at the ground truth vanishing point. \textcolor{blue}{Blue} is the output of NeurVPS\cite{zhou2019neurvps} and \textcolor{magenta}{Magenta} is the output of Zhou et.al.'s method\cite{zhou2017detecting}.
}
\label{fig:vp_ex_2}
\end{figure*}

We evaluate our unsupervised recurring pattern based VPD method on a projective view subset of 147 images with labeled VP ground truth and compare it with  a state-of-the-art supervised deep learning vanishing point detector \textbf{NeurVPS} \cite{zhou2019neurvps} and another baseline method \cite{zhou2017detecting}. Fig.~\ref{fig:vp_ex_1} and Fig.~\ref{fig:vp_ex_2} show the output of the three methods along with the ground truth. 

\clearpage
\subsection{Translation Symmetry Detection}

In Sec.~5.2 of the paper we stated that the cross ratio of feature points on four RP instances related by translation symmetry would take a value close to 4/3.  To prove this, consider four co-linear 3D points, in sequential order $A$, $B$, $C$, $D$, and their projections $A'$, $B'$, $C'$, $D'$ on the image plane (Fig.~{\color{red}8b} in the main paper).
By virtue of translation symmetry, $AB=BC=CD= d$ for some 3D translation vector length $||t||=d$.
The cross ratio, defined as \cite{mundy1992projective}:
\begin{equation} \label{eq:cross_ratio}
    cr (A, B, C, D) = \frac{AC*BD}{BC*AD}=
    cr (A', B', C', D')
\end{equation}
is an invariant value shared by four co-linear 3D points and their 2D image projections regardless of camera position and orientation. For the case of 3D translation symmetry, this computed invariant value will be
$$  \frac{AC*BD}{BC*AD}  = 
\frac{(d+d)(d+d)}{d(d+d+d)} = \frac{4}{3}. $$

\begin{figure}[b!]
    \centering
    \begin{subfigure}{0.48\linewidth}
    \includegraphics[width = \textwidth]{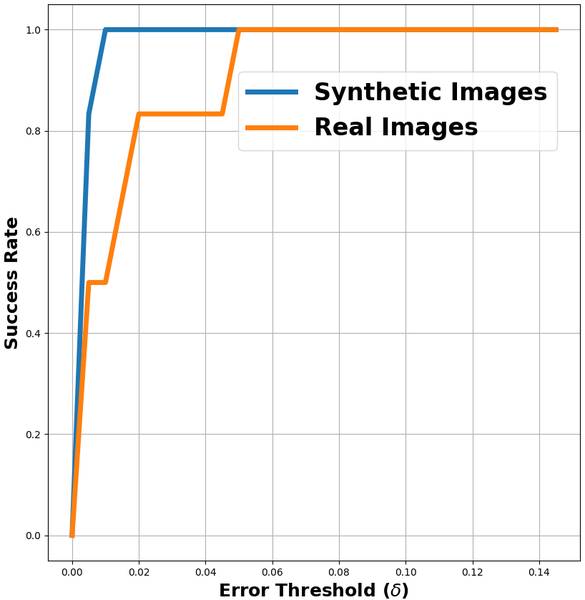}
    \caption{TS\_GT Dataset (12 images)}
    \end{subfigure}
    \hfill
    \begin{subfigure}{0.48\linewidth}
    \includegraphics[width = \textwidth]{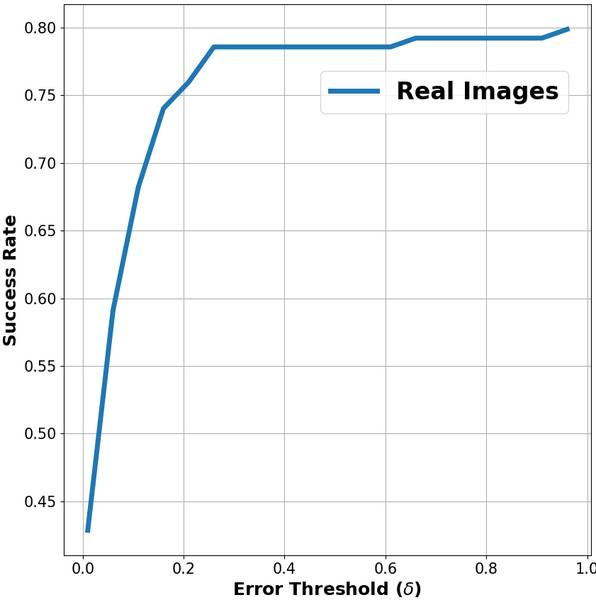}
    \caption{VPD Dataset (147 images)}
    \end{subfigure}
    \caption{The above figure represents the success rate of our translation symmetry detection for varying thresholds. (a) represents the success rate on a translation symmetry ground truth (TS\_GT) dataset consisting of 6 synthetic images and 6 real images for which the translation symmetry is known to exist (example images are shown in Fig.~\ref{fig:ts_examples}). (b) represents the success rate on the VPD dataset with 147 images.}
    \label{fig:ts_roc}
\end{figure}

\begin{figure}[b!]\centering
\vspace{-10pt}
    \setlength\tabcolsep{0.25pt}
    \begin{tabular}{cccc}
    \includegraphics[width = 0.25\textwidth]{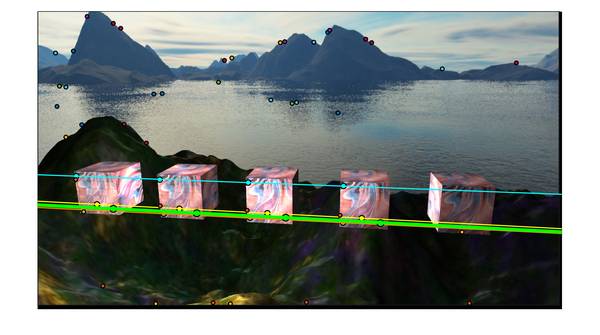} &
    \includegraphics[width = 0.25\textwidth]{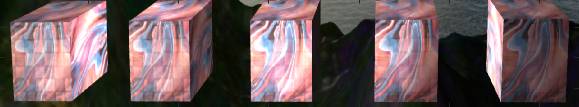} &
    \includegraphics[width = 0.25\textwidth]{Skanda's Figures/ts_supl/org/1004.png} &
    \includegraphics[width = 0.25\textwidth]{Skanda's Figures/ts_supl/rect/1004_rect.png}  \\
    
    \includegraphics[width = 0.25\textwidth]{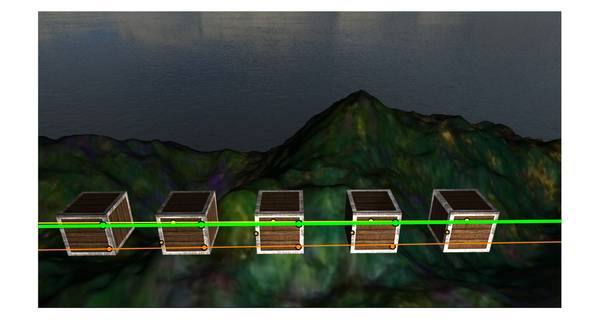} &
    \includegraphics[width = 0.25\textwidth]{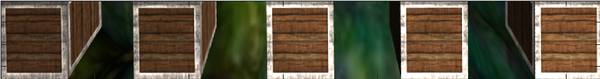} &
    \includegraphics[width = 0.25\textwidth]{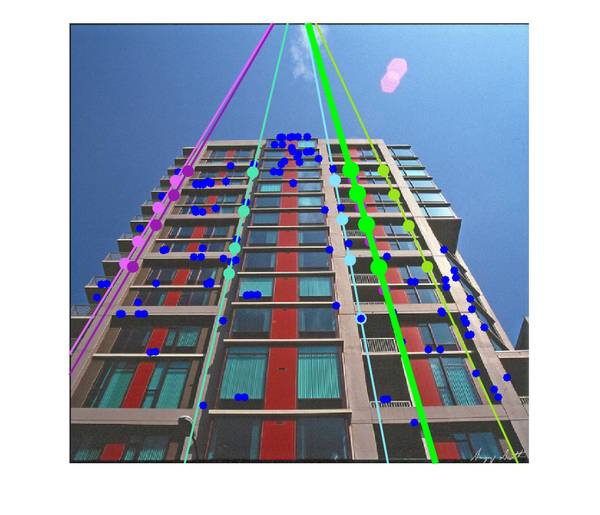} &
    \includegraphics[width = 0.25\textwidth]{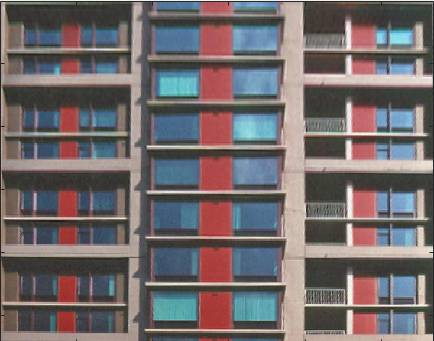} \\ 

    \includegraphics[width = 0.25\textwidth]{Skanda's Figures/ts_supl/org/12__1.3333.png} &
    \includegraphics[width = 0.25\textwidth]{Skanda's Figures/ts_supl/rect/12_rect.png} &
    \includegraphics[width = 0.25\textwidth]{Skanda's Figures/ts_supl/org/1017.png} &
    \includegraphics[width = 0.25\textwidth]{Skanda's Figures/ts_supl/rect/1017_rect.png}  \\
    
    \includegraphics[width = 0.25\textwidth]{Skanda's Figures/ts_supl/org/19.png} &
    \includegraphics[width = 0.25\textwidth]{Skanda's Figures/ts_supl/rect/19_rect.png} &
    \includegraphics[width = 0.25\textwidth]{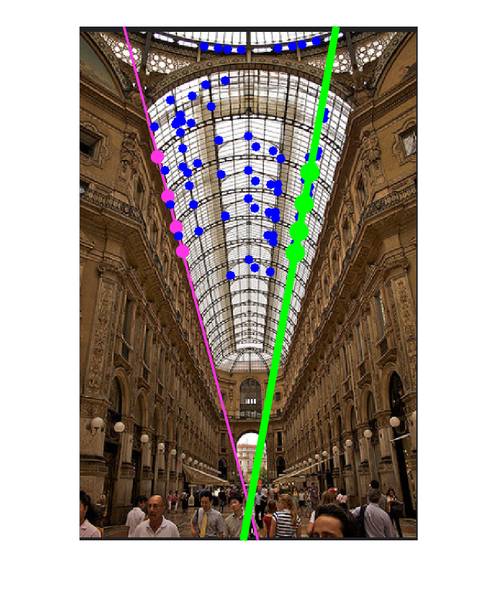} &
    \includegraphics[width = 0.25\textwidth]{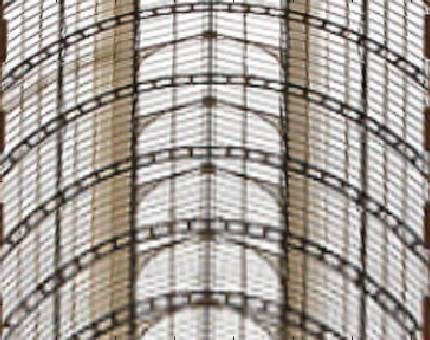} \\ 
    
    \includegraphics[width = 0.25\textwidth]{Skanda's Figures/ts_supl/org/ts_014__1.3114.png} &
    \includegraphics[width = 0.25\textwidth]{Skanda's Figures/ts_supl/rect/ts_014_rect.png} &
    \includegraphics[width = 0.25\textwidth]{Skanda's Figures/ts_supl/org/1044.png} &
    \includegraphics[width = 0.25\textwidth]{Skanda's Figures/ts_supl/rect/1044_rect.png}  \\
    
    (a) TS\_GT & (b) Rectified & (c) VPD  & (D) Rectified\\
    
    \end{tabular}
    \vspace{-5pt}
    \caption{Sample results for translation symmetry detection. (a) represents the original images of the translation symmetry ground truth (TS\_GT) dataset for which translation symmetry was known to exist. (b) represents rectified outputs of TS\_GT images. (c) represents translation symmetry detected in sample images from VPD dataset, and (d) represents rectified outputs of VPD images.} 
    \vspace{-5pt}
    \label{fig:ts_examples}
\end{figure}

Fig.~\ref{fig:ts_roc} evaluates the success rate of translation symmetry for varying thresholds on two different dataset. First, a small translation symmetry ground truth (TS\_GT) dataset was created consisting of 12 images (6 synthetic, 6 real images) shown in Fig.~\ref{fig:ts_examples}, for which translation symmetry is known to exist. Second, success rate of translation symmetry was calculated on the VPD dataset. We define success rate for translation symmetry ($SR_{ts}$) as 
$$ SR_{ts} = \dfrac{\# \text{ of images with translation symmetry}}{\# \text{ of images in the dataset}}.$$ 
The success rate was calculated for varying thresholds in the range $t = (0, 0.15)$ for TS\_GT dataset and $t =(0, 1)$ for VPD dataset. It can be seen that success rate reaches a maximum for TS\_GT images at the very low threshold of $t = 0.06$ 

Fig.~\ref{fig:ts_examples} shows different examples in which translation symmetry was detected from both the aforementioned datasets. Column (a) represents images from TS\_GT dataset and column (c) represents images from VPD datasets. Columns to their right represent the rectified outputs of the RPs from their corresponding images. Since URPD is completely unsupervised, we make no assumptions regarding any image. Consequently, we do not have any information regarding the camera parameters with which the image was captured. We maintain this assumption even for the synthetically generated images. Thus the rectified outputs are affine views of the original RPs \cite{Schaffalitzky00a} and therefore, aspect ratios of the rectified outputs do not always match the aspect ratio of the RPs in the real 3D scene (example: first synthetic image of Fig.~\ref{fig:ts_examples} (row 1, columns 1 and 2)). This is also the case in Fig.~{\color{red}9} of the main paper. It should also be noted that all images in our VPD dataset have translation symmetry of the RPs. But the ground truth for the same is unknown.

\clearpage
\subsection{RP Instance Counting}
\begin{figure}[b!]
    \centering
    \begin{subfigure}{0.45\linewidth}
    \includegraphics[width = \textwidth]{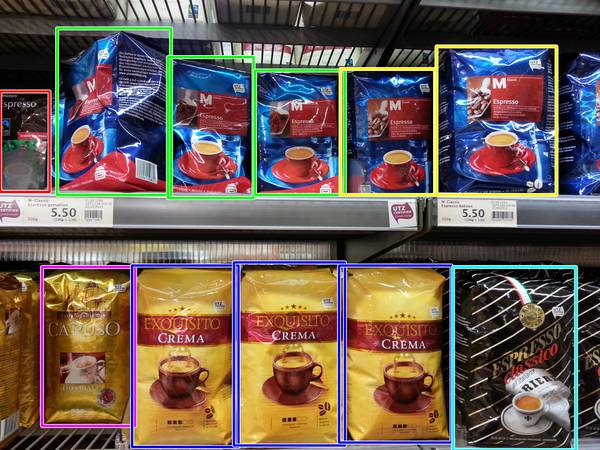}
    \caption{Original Product Counting Annotation from \cite{george2014recognizing}. Different color indicates product belonging to different product types.}
    \end{subfigure}
    \hfill
    \begin{subfigure}{0.45\linewidth}
    \includegraphics[width = \textwidth]{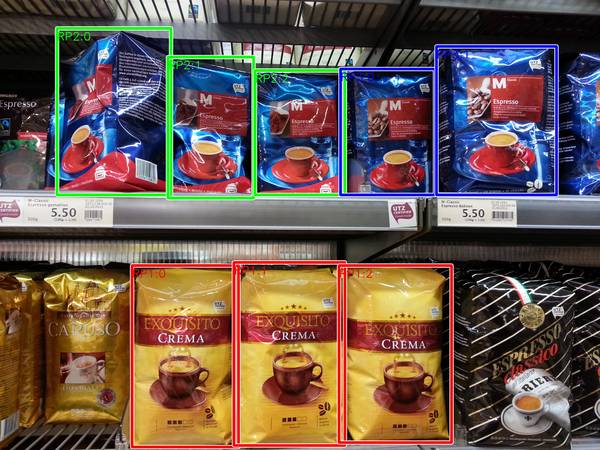}
    \caption{Our revised RP annotation. Different color indicates different RPs.}
    \end{subfigure}
    \caption{A comparison of \textbf{Grozi-3.2K} annotation}
    \label{fig:grozi_label_comp}
\end{figure}

As mentioned in Sec.~5.3 of the paper, we exclude the annotation of certain product types in \textbf{Grozi-3.2K} \cite{george2014recognizing} with no RPs. Fig.~\ref{fig:grozi_label_comp} shows the comparison of annotation difference.

\begin{figure}[b!]
    \centering
    \begin{subfigure}[t]{0.22\linewidth}
    \includegraphics[width = \textwidth,valign=t]{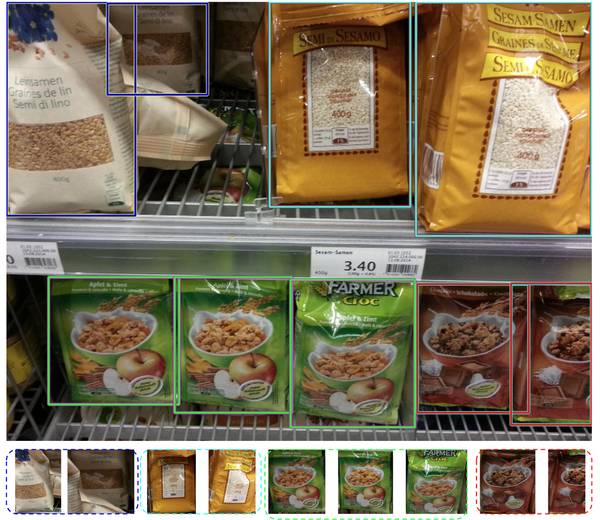}
    \caption{Ground Truth}
    \end{subfigure}
    \hfill
    \begin{subfigure}[t]{0.22\linewidth}
    \includegraphics[width = \textwidth,valign=t]{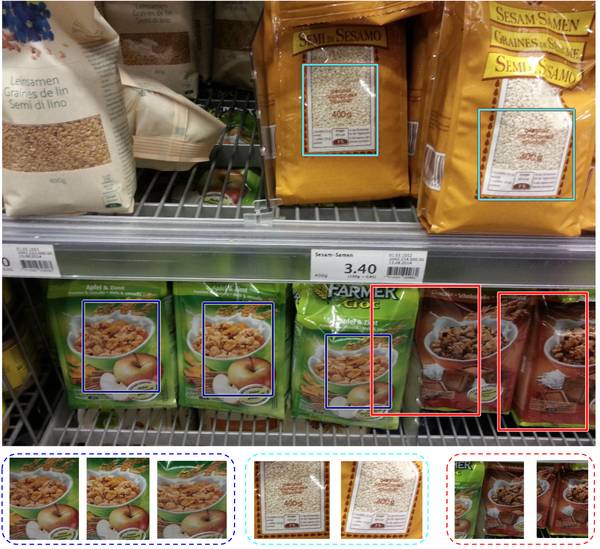}
    \caption{RESCU}
    \end{subfigure}
    \hfill
    \begin{subfigure}[t]{0.22\linewidth}
    \includegraphics[width = \textwidth,valign=t]{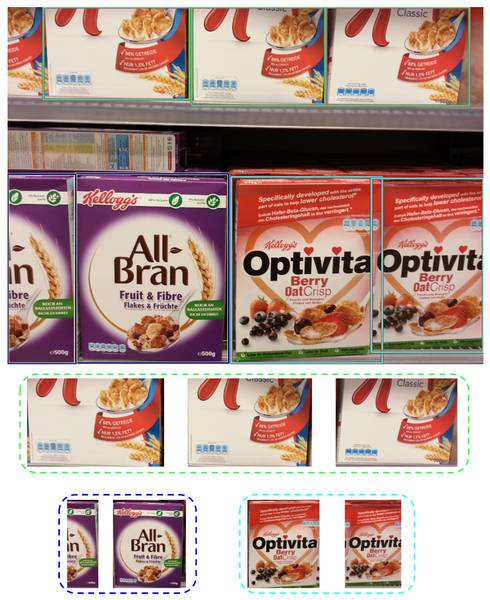}
    \caption{Ground Truth}
    \end{subfigure}
    \hfill
    \begin{subfigure}[t]{0.22\linewidth}
    \includegraphics[width = \textwidth,valign=t]{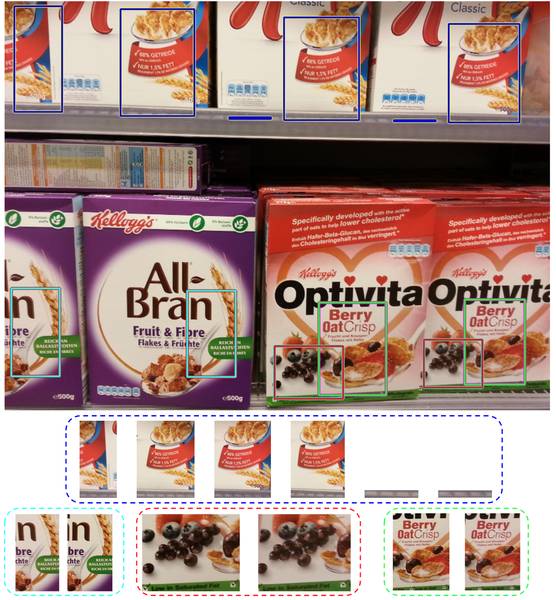}
    \caption{RESCU}
    \end{subfigure}
    \caption{Examples of our RESCU on \textbf{Grozi-3.2K}. 
    \textbf{(a, b):} Multiple RP detections with correct RPI counting result on man-made deformable products. The top-left RP is not detected due to rapid occlusion and lighting condition variation.
    \textbf{(c, d):} Multiple RP detections with \textit{mostly} correct RPI counting on man-made rigid products. The top RP is wrongly detected with two more RP instances, which due to the feature similarity on the shelf bar, and RESCU tolerance to partial occlusion.
    }
    \label{fig:grozi_ex}
\end{figure}

Fig.~\ref{fig:grozi_ex} shows some examples of our RESCU on \textbf{Grozi-3.2K}, including both success and failure RP counting cases. \cite{geng2018fine} has no available code, so Fig.~\ref{fig:grozi_ex} doesn't include the comparison with their method.

\clearpage
\subsection{Enhanced Image Caption from Detected RPs}
As mentioned in Sec.~5.4 of the paper, we enhance image captions with information obtained from detected RPs, VPs, TS, and counting. To achieve this, we employ OFA, a unified multimodal pre-trained network \cite{wang2022unifying}. OFA is a Transformer based, modal-agnostic pre-trained network that has achieved state-of-the-art results in multiple multimodal benchmarks including image captioning and visual grouding. We use the \textit{Large} size OFA model and weights pre-trained on MS COCO Caption dataset \cite{chen2015microsoft} for generating captions.

\begin{figure*}
    \centering
    \includegraphics[width=\linewidth]{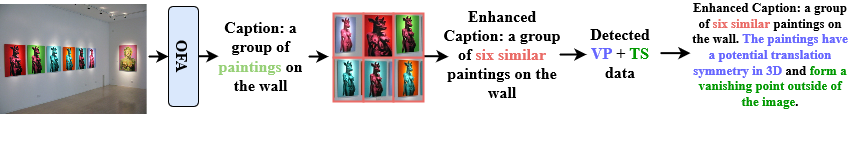}
    \caption{Proposed image caption enhancement pipeline. We obtain image captions using OFA \cite{wang2022unifying}. We parse the sentence for collective nouns or a subject noun and use the previously discovered dominant RP count for that corresponding image to add the detected noun's count. We further add discovered VP and TS information to the final caption. }
    \label{fig:cap_ex}
\end{figure*}

\begin{figure}[h]
\centering
\begin{subfigure}[t]{0.23\linewidth}
    \centering
    \includegraphics[width=1\linewidth]{Keaton's Figures/captions/animate/babies.jpg}
    \caption*{\sout{A~group~of} \textcolor{red}{six similar} babies sitting on the couch.}
\end{subfigure}
\unskip\ \vrule\
\begin{subfigure}[t]{0.23\linewidth}
    \centering
    \includegraphics[width=1\linewidth]{Keaton's Figures/captions/animate/llamas.jpg}
    \caption*{\sout{A~heard~of} \textcolor{red}{six~similar} llamas standing on top of a hill.}
\end{subfigure}
\unskip\ \vrule\
\begin{subfigure}[t]{0.23\linewidth}
    \centering
    \includegraphics[width=1\linewidth]{Keaton's Figures/captions/animate/ducks.jpg}
    \caption*{\sout{A~group~of} \textcolor{red}{four~similar} ducks swimming in the water.}
\end{subfigure}
\unskip\ \vrule\
\begin{subfigure}[t]{0.23\linewidth}
    \centering
    \includegraphics[width=1\linewidth]{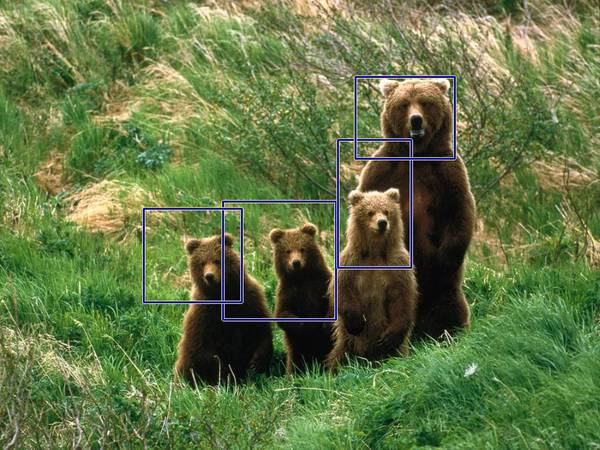}
    \caption*{\sout{A~group~of} \textcolor{red}{four~similar} brown bears standing in the grass.}
\end{subfigure}
\\
\vfill
\par\noindent\rule{\textwidth}{0.4pt}
\begin{subfigure}[t]{0.23\linewidth}
    \centering
    \includegraphics[width=1\linewidth]{Keaton's Figures/captions/objects/dolls.jpg}
    \caption*{\sout{A~group~of} \textcolor{red}{four~similar} Russian nesting dolls on a white background.}
\end{subfigure}
\unskip\ \vrule\
\begin{subfigure}[t]{0.23\linewidth}
    \centering
    \includegraphics[width=1\linewidth]{Keaton's Figures/captions/objects/statue.jpg}
    \caption*{An old picture of \textcolor{red}{six~similar} stone statues on a wall.}
\end{subfigure}
\unskip\ \vrule\
\begin{subfigure}[t]{0.23\linewidth}
    \centering
    \includegraphics[width=1\linewidth]{Keaton's Figures/captions/grocery/rice.jpg}
    \caption*{\sout{A~row~of} \textcolor{red}{six~similar} mister rice boxes on a supermarket shelf.}
\end{subfigure}
\unskip\ \vrule\
\begin{subfigure}[t]{0.23\linewidth}
    \centering
    \includegraphics[width = 1\textwidth]{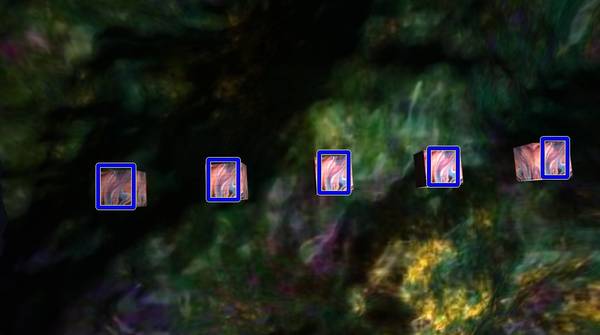}
    \caption*{\sout{A~group~of} \textcolor{red}{five~similar} squares with different colors on a tree.}
\end{subfigure}
\\
\vfill
\par\noindent\rule{\textwidth}{0.4pt}
\begin{subfigure}[t]{0.23\linewidth}
    \centering
    \includegraphics[width=1\linewidth]{Keaton's Figures/captions/objects/bottles.jpg}
    \caption*{\sout{A~group~of} \textcolor{red}{five~similar} bottles with different colored liquid in them on a table. \textcolor{blue}{The~bottles~have~a~potential~translation~symmetry~in~3D} and \textcolor{green}{form~a~vanishing~point~outside~of~the~image}.}
\end{subfigure}
\unskip\ \vrule\
\begin{subfigure}[t]{0.23\linewidth}
    \centering
    \includegraphics[width=1\linewidth]{Keaton's Figures/captions/simul/people_sky.jpg}
    \caption*{\sout{A~group~of} \textcolor{red}{five~similar} men jumping in the sky. \textcolor{blue}{The~men~have~a~potential~translation~symmetry~in~3D} and \textcolor{green}{form~a~vanishing~point~outside~of~the~image}.}
\end{subfigure}
\unskip\ \vrule\
\begin{subfigure}[t]{0.23\linewidth}
    \centering
    \includegraphics[width=1\linewidth]{Keaton's Figures/captions/simul/box_rows.jpg}
    \caption*{\sout{A~series~of} \textcolor{red}{five~similar} canvases in front of a lake and mountains. \textcolor{blue}{The canvases~have~a~potential~translation~symmetry~in~3D} and  \textcolor{green}{form~a~vanishing~point~outside~of~the~image}.}
\end{subfigure}
\unskip\ \vrule\
\begin{subfigure}[t]{0.23\linewidth}
    \centering
    \includegraphics[width=1\linewidth]{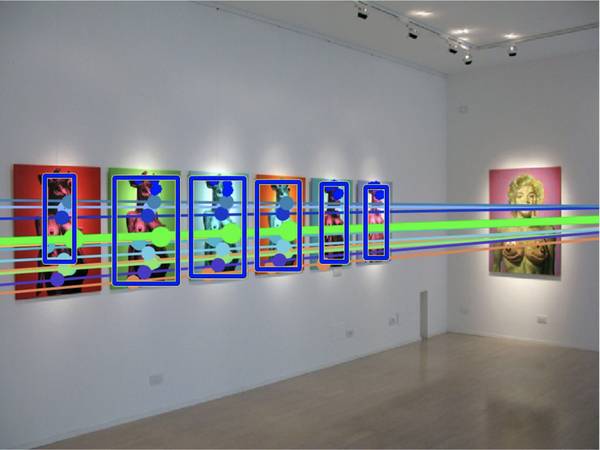}
    \caption*{\sout{A~group~of} \textcolor{red}{six~similar} paintings on the wall. \textcolor{blue}{The~paintings~have~a~potential~translation~symmetry~in~3D} and  \textcolor{green}{form~a~vanishing~point~outside~of~the~image}.}
\end{subfigure}

\caption{Example image-caption pairs we enhance using detected RP and VP information. The captions below each image contain the original text generated by OFA \cite{wang2022unifying}. We add the dominant RP's \textcolor{red}{count} in red following collective nouns. If detected, additional \textcolor{blue}{translation symmetry} and \textcolor{green}{VP} information is added to the caption.}
\label{fig:captions}
\end{figure}

We begin by generating a caption for the image using OFA. We then parse the image's generated caption for instances of collective nouns such as ``group''. If no collective nouns are found, we search the string for noun instances proceeded by the word ``of''. Finally, we follow a simple grammar to replace collective nouns and words dependent on them with the dominant RP's count discovered by by RESCU. If no collective nouns are detected, we place the RP's count after ``of'' and before the detected noun. In addition to enhancing the image caption with RP counts, we add further context by noting potential translation symmetry and whether a detected vanishing point lies inside or outside the image borders. The proposed caption enhancing pipeline is shown in Figure \ref{fig:cap_ex}. Figure \ref{fig:captions} shows example image-caption pairs augmented by our detected RP and VP information. 

\begin{figure*}
    \centering
    \includegraphics[width=\linewidth]{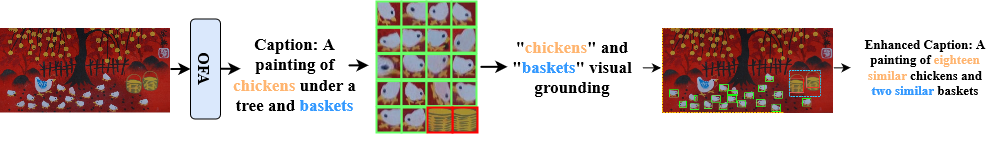}
    \caption{Caption enhancement using two detected RPs. We utilize OFA's visual grounding for the prompts ``\textcolor{orange}{chickens}'' and ``\textcolor{cyan}{baskets}'' to determine which noun subject the RP is most likely to corresponds with.}
    \label{fig:two_rp}
\end{figure*}

We further utilize OFA to handle multiple potential RPs and their corresponding nouns in a caption. If the image's caption contains multiple subject nouns, we use OFA's visual grounding to ground each detected subject noun in the caption. This gives OFA's bounding box for the potential region of the text input. We then check the detected RPs overlap with OFA's proposed region. We assign the corresponding RP to the region and its corresponding noun if the RP's total area has at least 0.90 overlap with OFA's proposed region. Figure \ref{fig:two_rp} demonstrates an example image-caption pair enhanced with multiple detected RPs and OFA's proposed regions for two detected nouns in the image's caption.

\clearpage


\bibliographystyle{splncs}
\bibliography{egbib}